\pdfoutput=1

\documentclass[11pt]{article}

\usepackage{ACL2023}

\usepackage{times}
\usepackage{latexsym}
\usepackage{tikz}
\usetikzlibrary{arrows.meta, positioning}

\usepackage[T1]{fontenc}

\usepackage[utf8]{inputenc}

\usepackage{microtype}

\usepackage{inconsolata}

\title{A Breadth-First Catalog of Text Processing, Speech Processing and Multimodal Research in South Asian Languages}

\author{Pranav Gupta \\
  Lowe's\\
  1000 Lowes Boulevard \\
  Mooresville, NC, USA - 28117 \\
  \texttt{pranav.gupta@lowes.com} \\} 

\begin{document}
\maketitle

\begin{abstract}
We review the recent literature (January 2022-October 2024) in South Asian languages on text-based language processing, multimodal models, and speech processing, and provide a spotlight analysis focused on 21 low-resource South Asian languages, namely Saraiki, Assamese, Balochi, Bhojpuri, Bodo, Burmese, Chhattisgarhi, Dhivehi, Gujarati, Kannada, Kashmiri, Konkani, Khasi, Malayalam, Meitei, Nepali, Odia, Pashto, Rajasthani, Sindhi, and Telugu. We identify trends, challenges, and future research directions, using a step-wise approach that incorporates relevance classification and clustering based on large language models (LLMs). Our goal is to provide a breadth-first overview of the recent developments in South Asian language technologies to NLP researchers interested in working with South Asian languages.
\end{abstract}

\section{Introduction}
South Asian languages, with their diverse script systems, phonological structures, sociolinguistic contexts, and around 2 billion speakers, present unique challenges for both natural language processing (NLP) and speech processing tasks \citep{subbarao2008typological, bhatt-etal-2021-universality, joshi2020state}. The region is home to hundreds of languages, many of which are low-resource, complicating the development of robust models for diverse tasks such as machine translation, speech recognition, and large language models (LLMs). In recent years, with the advent of large language models and similar breakthroughs in vision and speech, there has been hope that the benefits reaped by high-resource language communities will manifest to the same extent in low- and medium-resource languages in South Asia. The Asia Pacific artificial intelligence market, which includes several South Asian countries and valued at \$50.41 million in 2023, is projected to grow at $45.7\%$ from 2024 to 2030 \citep{asia-pacific-ai-market-2024}. 

Several commercial organizations have already allocated significant resources in this direction. The main ones are Google, Microsoft, and Meta, who each have dedicated projects aimed at improving the linguistic diversities of their models and training data, such as Project Vaani \citep{vaani}, Microsoft Cognitive Services \citep{microsoft-indian-languages}, and No Language Left Behind \citep{no-language-left-behind}. Other major South Asian companies have also started building highly capable language models that are built with a ``local-first'' approach, resulting in highly focused models for South Asian applications \citep{tata-nvidia-llm, techmahindra-indus, ola-krutrim, reliance-nvidia-llm, zoho-llm}. Many South Asian startups are also focused on creating and applying large language models to South Asia-specific applications \citep{lankalawchatbot2024, naamche2024, traversaal2024, corover-bharatgpt}. Government organizations are also leading efforts in this direction, such as the Ministry of Culture and Information Technology in Nepal \citep{mocit2024}, the Ministry of Electronics and Information Technology in India \citep{meity2023}, the National Technology Fund in Pakistan \citep{ignite-urdu}, the Computer Council in Bangladesh \citep{bangladesh-edge}, the Information and Communication Technology Agency in Srilanka \citep{icta-local-languages}, and the National Center for Information Technology in Maldives \citep{ncit-maldives}, in addition to non-profit AI initiatives from organizations in south Asia and elsewhere, for example, \citet{Gupta-Banerjee-2024} and \citet{ardila2020commonvoicemassivelymultilingualspeech}.

Given the pace of innovation in language technologies in South Asia and other demographics, survey and review articles that summarize the latest developments can help foster collaboration, inform perspectives, and educate new researchers in the field. This paper aims to provide a systematic review of the existing literature, identify trends, and propose future research directions in text-based models, multimodal models, and speech-based models. 

The closest article we found that is similar to our review is \citet{kj2024decoding}, which focuses on 84 relevant publications, prioritizing depth over breadth. In this paper, we choose to prioritize breadth over depth, relying on machine learning-based literature survey techniques such as classification, clustering and topic modeling. We also cover multiple themes within the same article, namely language models, multimodal approaches, audio models. Other articles in this field focus on specific topics such as performance \citep{hasan2024large}, data \citep{parida2024building}, and bias \citep{gupta-etal-2024-sociodemographic}. 


\section{Methodology}

\begin{figure}[!ht]
    \centering
    \begin{tikzpicture}[
        node distance=0.8cm and 0.6cm,
        box/.style = {rectangle, draw, rounded corners, align=center, fill=blue!20, text width=6cm, minimum height=1.5em},
        arrow/.style = {-{Stealth}}
    ]

    \node[box] (step1) {Google Scholar queries};
    \node[box, below=of step1] (step2) {Randomized sampling and labeling as relevant/irrelevant};
    \node[box, below=of step2] (step3) {GPT-4o predicts relevance for each paper title};
    \node[box, below=of step3] (step4) {O1 model groups papers into several themes};

    \draw[arrow] (step1) -- (step2);
    \draw[arrow] (step2) -- (step3);
    \draw[arrow] (step3) -- (step4);

    \end{tikzpicture}
    \caption{\label{fig:scrape-methodology}Flowchart for the Research Paper Classification and Theme Generation Process}
\end{figure}
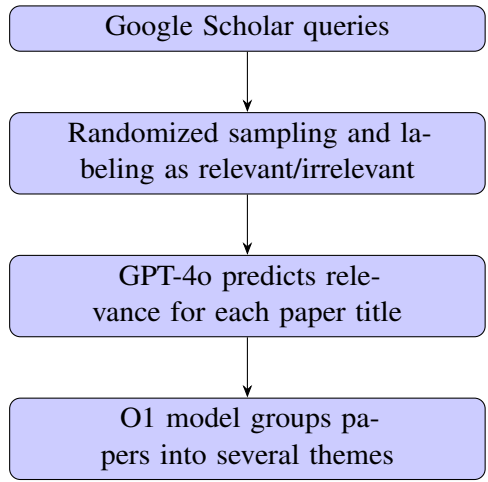

We adopt a multi-stage approach to curate and analyze articles, according to the schematic mentioned in Figure~\ref{fig:scrape-methodology}:
\begin{itemize}
    \item Google Scholar queries: We used Google Scholar for our initial discovery of candidate papers, given its versatility and ability to efficiently index literature from multiple sources. We used \textit{Publish or Perish} software \citep{harzing2013} to compile the metadata of the articles efficiently. In order to generate more diversity in our searches, we started with a base prompt, generated 5 queries from GPT-4o \citep{openai2024gpt4}, and then manually reviewed those queries to make any necessary edits before using them to search for articles on Google Scholar.
    \item We sampled 20 paper titles out of our search results and labeled them as relevant or irrelevant.
    \item GPT-4o used these labels as in-context training data to predict relevance labels for all articles, using a batch size of 100 articles. Articles predicted to be irrelevant and/or published before 2022 were excluded from further analysis.
    \item We used OpenAI's O1 model, given its advanced reasoning capabilities \cite{zhong2024evaluationopenaio1opportunities}, to group the papers predicted to be relevant into various topics. We also tried other approaches such as BERTopic \citep{grootendorst2022bertopic} for topic modeling, but the resulting topics were not as coherent as the topics predicted by the O1 model. 
\end{itemize}
Table~\ref{tab:summary-table} summarizes the results of our approach for each of the 3 fields we investigated. 
\begin{table}[!ht]
\small
\centering
\begin{tabular}{p{2.8cm}p{2cm}p{2.3cm}}
\hline
\textbf{Category} & \textbf{Initial Papers} & \textbf{Relevant Papers Detected (since 2022)}\\
\hline
Language Models & 1519 & 369\\
Multimodal Models & 1709 & 84\\
Speech Processing & 558 & 52\\
\hline
\end{tabular}
\caption{Summary of relevant papers detected}
\label{tab:summary-table}
\end{table}

\section{\label{sec:text}Findings and Trends: Language Models}
We found the following major themes in textual language processing:

\subsection{Machine translation and cross-lingual transfer learning}
In this topic we found papers on analyzing machine translation evaluations in low resource languages \citep{singh2024}, lexical enhancing of pre-trained language models for low-resource machine translation \citep{lexicalenhancing}, translating code-switched languages \citep{huzaifah-etal-2024-evaluating}, and Romanization-based LLM finetuning \citep{j-etal-2024-romansetu}. 

\subsection{Hate speech, offensive content, sentiment analysis, and other applications}
In this topic, we found several task-specific papers, such as sentiment analysis \cite{llm-2023-Tonmoy, llm-2024-rani-wildre, llm-2023-maqsood}, sarcasm detection \citep{llm-2023-Bhaumik, llm-2024-Hassan}, hope speech detection \citep{llm-2023-Nath}, hate speech detection \citep{llm-2022-Roy, llm-2023-Bansod}, emotion detection \citep{llm-2023-Vedula}, fake news detection \citep{llm-2022-Rauf, balaji-etal-2023-nlp}, depression detection \citep{Hoque-Salma-2023}, cyberbullying classification \cite{llm-2023-Hoque}, humor detection \citep{humor}, and misogyny detection \cite{mimic-misogyny}.

\subsection{Bias and fairness}
Bias and fairness studies have typically focused on Western cultural contexts and stereotypes, which do not take into account south Asia specific biases such as caste and religion. In this topic, the most comprehensive dataset we found was BheD \citet{indian-bhed}, where even GPT 3.5, a leading LLM, was found to have a much higher propensity to stereotype along the dimensions of caste and religion (79.52 $\%$ and 70.49 $\%$ respectively). Other studies have included aspects such as inclusivity \cite{khanuja-etal-2023-evaluating}, gender and religious bias \cite{sadhu2024}, and gender bias that emerges while translating from English to South Asian languages with gender-neutral pronouns, for example, Bengali \citep{pronoun-bias}.

\subsection{Adapting LLMs for South Asian Languages}
In this topic we found papers on aspects such as prompt engineering \cite{hasan-zero}, efficient foundation models \cite{niyogi2024}, and translation-assisted chain-of-thought prompting \cite{taco}. 

\subsection{Datasets and Benchmarks}
In this topic, we found several datasets spanning various domains such as legal documents \citet{kapoor-etal-2022-hldc}, abstractive summarization of medical queries \citet{khan-etal-2023-banglachq}, multilingual question-answering \citep{singh2024}, machine translation metric meta-evaluation \citep{sai-b-etal-2023-indicmt}, named entity recognition \cite{10103464, murthy-etal-2022-hiner}, classical literature-based monolingual corpora \citep{bhattacharyya-etal-2023-vacaspati}, language-specific benchmark suites \citep{joshi:2022:WILDRE6}, treebanks \cite{abirami-etal-2024-aalamaram}, LLM generation \citep{indicgenbench}, and parallel corpora \citep{ramesh-etal-2022-samanantar}. We also found task-specific datasets such as fake review detection \citep{fakereview}.

\section{\label{sec:multimodal}Findings and Trends: Multimodal Models}
We found the following major themes in multimodal modeling:
\subsection{Multimodal Machine Translation}
Multimodal machine translation aims to improve translation quality by using multimodal input such as static images, such as the ViTA model \citep{gupta-etal-2021-vita}. Representative papers from this theme include \citet{multimodal-2023-LS-Meetei} (Hindi-English), \citet{multimodal-2023-SR-Laskar} (English-Assamese), and \citet{multimodal-2022-SR-Laskar} (English-Bengali), and \citet{multimodal-2022-N-Sethi} (Sanskrit-Hindi). 

\subsection{Image Captioning}
On this topic, we mostly found papers on image captioning in languages such as Hindi \citep{multimodal-2022-SK-Mishra, multimodal-2023-SK-Mishra}, Assamese \citep{multimodal-2023-P-Choudhury, multimodal-2022-R-Das}, Urdu \citep{multimodal-2023-MK-Afzal}, Tamil \citep{multimodal-2022-VHV-Kumar}, Telugu \citep{multimodal-2023-R-Reddy}, and Bengali \citep{multimodal-2023-B-Das}. 

\subsection{Multimodal Hate Speech and Offensive Content}
On this topic, we found papers such as multimodal troll meme detection \citep{multimodal-2022-Hasan}, multimodal misogyny identification \citep{multimodal-2024-A-Singh}, and multimodal hate speech detection \citep{multimodal-2024-A-Singh}. 

\subsection{Multimodal Sentiment Analysis and Emotion Recognition}
In this topic, we found papers on multimodal emotion recognition \citep{Taheri2023-mt}, along with multimodal sentiment analysis of various types of content such as news \citep{Das2023-ej} and memes \citep{Elahi2023-ni}. 

\subsection{Multimodal Datasets}
Similar to the Visual Genome \citep{Krishna2017-qy}, we found Visual Genome datasets in South Asian languages such as Bengali \citep{Sen2022-tg} and Mizo \citep{miso-visual-genome}. In addition to well-known multilingual multimodal evaluation datasets such as \citet{thapliyal2022crossmodal3600massivelymultilingualmultimodal} and \citet{nielsen2022mumin}, we also found other interesting datasets such as multilingual audio/video facial expressions \citep{Singh2023-sy} and multimodal face emotion recognition on code-mixed memes in Tamil \citep{Kannan2023-py}. 

\section{\label{sec:speech}Findings and Trends: Speech Processing}
We found the following major themes in speech processing:
\subsection{Code Mixing}
We found papers on several code mixing-related topics such as normalizing code-mixed text to generate speech \citep{speech-2022-Manghat}, automatic speech recognition (ASR) in noisy code-mixed speech \citep{speech-2023-Verma}, adapters for code-mixing \citep{speech-2023-Kulkarni}, collaging monolingual corpora for code-mixed speech synthesis \citep{speech-2024-Hussein}, and benchmarking code-mixed speech synthesis \citep{speech-2023-Hamed}.

\subsection{Abusive speech, language identification, and speech-to-speech translation}
In this section, we found papers on specific tasks such as abusive speech detection \citep{speech-2024-Spiesberger}, language identification \citep{speech-2022-Kulkarni}, and speech-to-speech translation for English $\rightarrow$ Kannada \citep{speech-2023-Malage}, Punjabi $\rightarrow$ English \citep{speech-2024-Kaur}, and English $\rightarrow \{$ Hindi, Telugu, Gujarati, Marathi, Punjabi$\}$ \citep{speech-2024-Spiesberger}.  

\subsection{Speech modeling and speech recognition}
The most prominent of the recent models we found in this topic are the AI4Bharat wav2vec models for South Asian language speech recognition \citep{speech-2022-Javed}. We also found papers on aspects such as end-to-end speech recognition \citep{speech-2022-Changrampadi}, speech recognition focused on specific phonemes \citep{speech-2022-Gupta}, postprocessing for error correction \citep{speech-2022-Kumar}, senone prediction and senone mapping \citep{speech-2022-Madhavaraj}, and joint speech-text representation learning \citep{speech-2023-chen}. 

\section{Speech processing with low resources}
In this topic, we found papers on approaches relevant to low-resource speech recognition and synthesis, such as data augmentation \citep{speech-2024-Majhi}, novel loss functions for low-resource speech recognition \citep{speech-2023-Chen-meta}, text-to-speech pseudolabels \citep{speech-2022-Gupta}, mining audio and text pairs from public data \citep{speech-2023-Bhogale}, deep prefix tuning for dialect adaptation \citep{speech-2023-Alumäe}, and merging Hidden Markov Models with HiFiGAN \cite{su2020hifigan} to improve voice quality in text-to-speech systems \citep{speech-2023-Srivastava}.

\subsection{Datasets and benchmarks}
In this topic, we found papers on metrics \citep{shah2022worderrorrategood}, south Asia-specific versions \citep{speech-2023-Javed} of popular speech processing benchmarks such as \citet{Yang2021-sd}, and language-specific datasets such as \citet{speech-2023-Mirishkar} (Telugu) and \citet{speech-2022-Alam} (Bengali).




\section{\label{sec:lowres} Findings and Trends: Low-Resource Language Spotlight}
For this section, we searched the literature on the following low-resource languages. The languages were chosen because they have an official status at least in the regions in which they are widely spoken. We used literal string-based matching rather than topic modeling and LLM-based relevance, because the precision and recall of the GPT-4o relevance classifier was significantly low (precision 0.33, recall 0.7) when compared to the relevance models for textual, multimodal and speech processing on the labeled dataset used for in-context learning in the GPT-4o prompt. We observed that Nepali, Assamese, and Telugu were the most popular languages, with more than 50 papers each since 2022.

\subsection{Nepali (63 papers)}
Spoken primarily in Nepal and parts of India, Nepali belongs to the Indo-Aryan language family and is spoken by over 28 million people. We found the most papers in this language, with several major themes being covered, such as text summarization \citep{lowres-2022-S-Awale, lowres-2023-C-Pokhrel}, keyword generation \citep{lowres-2024-S-Shrestha}, sentiment analysis \citep{lowres-2023-N-Pahari, lowres-2024-F-Zimmer, lowres-2024-P-Gurung, lowres-2024-S-Pudasaini}, and named entity recognition/parts of speech tagging \citep{lowres-2024-B-Subedi, lowres-2023-N-Niraula, lowres-2024-A-Pradhan}. There are several papers on language modeling and evaluation in Nepali \citep{lowres-2024-B-Subedi, lowres-2024-S-Majhi, lowres-2023-S-Pudasaini, lowres-2024-N-Luitel}.
There have also been papers on multimodal applications such as visual question answering \citep{lowres-2024-A-Gyanwali}, 
\citep{lowres-2023-K-Parajuli}, video captioning \citep{hyoju2023nepali, lowres-2023-R-Budhathoki}, optical character recognition \citep{lowres-2024-P-Paudel}, and the translation of American Sign Language to Nepali \citep{lowres-2024-B-Paneru}. On the speech side, there has been significant work in speech recognition \citep{lowres-2022-A-Acharya, lowres-2023-B-Joshi-end2end, lowres-2023-B-Joshi-asr-attention, lowres-2022-B-Joshi-novel, lowres-2024-RR-Ghimire, lowres-2023-GR-Raj-active, lowres-2023-GR-Raj-aware} and text-to-speech synthesis \citep{lowres-2024-A-Rai, lowres-2023-S-Khadka}. 

We also observe an emerging trend in terms of studies on AI ethics \citep{lowres-2024-B-Jha} and AI applied to various domains such as social good \citep{lowres-2024-A-Kumar-safety, lowres-2022-U-Kapri}, studying social media trends \citep{lowres-2024-A-Dhakal}, along with health care applications such as Alzheimer's \citep{lowres-2022-S-Adhikari} and prenatal health \citep{lowres-2023-S-Poudel}. This shows how NLP research in Nepali has extended to cutting-edge topics beyond traditional topics such as machine translation.

\subsection{Assamese (58 papers)}
Assamese, an Indo-Aryan language, is spoken predominantly in the Indian state of Assam and parts of neighboring Indian states. It has around 14 million speakers. Research themes have included chatbots \citep{lowres-2024-S-Sarma-blstm}, transliteration \citep{lowres-2024-H-Baruah-social, lowres-2024-H-Baruah-roman}, named entity recognition \citep{lowres-2022-D-Pathak}, parts of speech tagging \citep{lowres-2023-D-Pathak-ensemble, lowres-2023-K-Talukdar, lowres-2024-S-Talukdar, lowres-2024-K-Talukdar, lowres-2024-R-Phukan-pos}, machine translation \citep{lowres-2022-SR-Laskar, lowres-2023-SR-Laskar, lowres-2024-K-Kashyap}, sentiment analysis \citep{Das2023-ej}, text summarization \citep{lowres-2024-PJ-Goutom, lowres-2022-N-Baruah, lowres-2023-PJ-Goutom}, data augmentation \citep{lowres-2023-C-Lalrempuii}, tokenizers \citep{lowres-2024-S-Tamang-tokenizer}, and text classification \citep{lowres-2023-C-Talukdar, lowres-2023-SP-Medhi}. 

Outside the text modality, we encountered research in image captioning \citep{lowres-2022-P-Nath}, speech-based vocabulary identification \citep{lowres-2022-D-Dutta}, Assamese sign language recognition \citep{lowres-2023-J-Bora}, emotion recognition from speech in Assamese \citep{lowres-2023-N-Choudhury}, and speech-based gender detection \citep{lowres-2022-K-Dutta}. We also came across speech signal and phonetic analysis studies, such as vowel speech signal identification \citep{lowres-2023-A-Sarmah}, word stress and its effects on Assamese and Mizo languages \citep{lowres-2023-J-Gogoi}, and Assamese-Bengali cognate detection \citep{lowres-2023-A-Nath}. There was some work on cultural and societal empowerment too, for example, linguistic preservation \citep{lowres-2023-N-Borah}, and machine learning-based detection of Assamese cultural objects such as Assamese Dhol (musical instrument) and Mohor sing (musical instrument) \citep{lowres-2023-MP-Lahkar}.

A striking feature in Assamese NLP literature when compared to other low-resource South Asian languages was the relatively larger number of papers on hate speech. Hate speech is a growing concern in South Asian communities \citep{liebowitz2021digitalization, minorityrights2022southasia}, and existing literature has covered offensive language \citep{lowres-2023-JK-Mim} and hate speech \citep{lowres-2023-K-Ghosh-hate, lowres-2023-K-Ghosh-hate2, lowres-2023-N-Baruah-hate} detection in Assamese and other neighboring languages such as Bengali and Bodo.

\subsection{Telugu (57 papers)}
Telugu, a Dravidian language, is spoken primarily in the Indian states of Andhra Pradesh and Telangana. It has over 85 million speakers. In Telugu, we found work on morphological analysis \citep{lowres-2023-P-Dasari}, sentiment analysis of news articles, social media and other textual data \citep{lowres-2024-VB-Viswanadh, lowres-2024-GJ-Naidu, lowres-2023-UR-Rayala, lowres-2022-DAV-Padmaja-Tallu}, text summarization \citep{lowres-2023-K-Bhuvaneshwari, lowres-2022-A-Lakshmi, lowres-2023-A-Babu-GL}, named entity recognition \citep{lowres-2022-SK-Gorla, duggenpudi-etal-2022-teluguner}, hate speech detection \citep{lowres-2024-C-Sai, lowres-2024-T-Achamaleh}, abusive comment detection \citep{lowres-2023-R-Priyadharshini}, LLM evaluation \citep{lowres-2024-KS-Kishore}, event extraction \citep{lowres-2022-SS-Burramsetty}, domain adaptation \citep{lowres-2022-A-Hema}, reinforcement learning from human feedback \citep{lowres-2023-J-Srinivas}, language identification \citep{lowres-2022-M-Jaswanth}, 
text classification \citep{lowres-2023-M-Marreddy, lowres-2022-G-Santhoshi}, graph-based topic modeling \citep{lowres-2024-SSG-Namburu}, word sense disambiguation \citep{lowres-2022-N-Koppula}, machine translation \citep{lowres-2023-B-Vamsi}, and question-answering \citep{lowres-2024-VR-KUMAR, lowres-2023-P-Ravva}. 

Multimodal research included handwriting recognition \citep{lowres-2022-B-Sankara-Babu, lowres-2024-B-Revathi, lowres-2022-B-MEENA, lowres-2024-BG-Reedy}, whereas speech processing literature consisted of speech recognition \citep{lowres-2022-A-Yadavalli, lowres-2023-GS-Mirishkar}, spoken digits modeling \citep{lowres-2022-P-Bhagath-corpora, lowres-2023-P-Bhagath}, speech intelligibility \citep{lowres-2022-SC-Venkateswarlu}, and text-to-speech modeling \citep{lowres-2023-CA-Kumar}. Code mixing is important in Telugu as it is in the majority of South Asian languages, and we observed that papers on other themes in Telugu NLP considered this aspect in their work. For a survey on Telugu-English code-mixed text, we refer the reader to \citet{lowres-2024-S-Maddu}. Applications included tourist assistance \citep{lowres-2023-S-Kolar}, farmer assistance \citep{lowres-2023-J-Srinivas, lowres-2024-GJ-Naidu}, and IoT devices \citep{lowres-2023-P-Bhagath-iot}. 

\subsection{Kannada (49 papers)}
Kannada, another Dravidian language, is spoken primarily in the Indian state of Karnataka. It has over 44 million speakers. In Kannada, we found papers on hope speech detection \citep{lowres-2022-A-Hande}, code-mixing \citep{lowres-2022-F-Balouchzahi, lowres-2022-V-Vajrobol, lowres-2022-HL-Shashirekha}, machine translation \citep{lowres-2022-SK-Sheshadri, lowres-2023-S-Shetty, lowres-2023-SR-Kashi, lowres-2024-HS-Patil}, paraphrase generation \citep{lowres-2023-HM-Anagha}, part-of-speech tagging \citep{lowres-2023-HR-Mamatha, lowres-2022-S-Shetty}, preprocessing \citep{lowres-2024-CB-Lavanya}, authorship attribution \citep{lowres-2022-CP-Chandrika-authorship-attribution}, and syntactical parsing \citep{lowres-2022-M-Rajani-Shree}. 

On the speech side, we found papers on code-mixed speech synthesis \citep{lowres-2024-SK-Suresh}, speech recognition in noisy environments \citep{lowres-2024-GT-Yadava, lowres-2023-YG-Thimmaraja}, visual speech recognition \citep{lowres-2024-R-Shashidhar, lowres-2023-S-Rudregowda}, and speech-to-speech translation \citep{lowres-2023-RV-Malage}. Other themes included image caption generation \citep{lowres-2023-K-Chethas}, subtitle generation \citep{lowres-2023-Santosh}, and handwritten character recognition \citep{lowres-2023-C-Hebbi, lowres-2022-S-Tejas}. We also found a paper analyzing Kannada social media trends \citep{lowres-2023-A-Dey}. According to our survey, within the low-resource languauges, Kannada seemed to be the most advanced when it came to speech-related research, especially in multimodal speech recognition, subtitle generation, and speech-to-speech translation, which also happen cutting-edge topics in high-resource languages. 

We also found papers describing AI applications such as virtual medical assistance \citep{lowres-2023-BGM-Bai}, legal document summarization \citep{lowres-2024-S-Megha}, and oral community knowledge management \citep{lowres-2023-M-Aparna}.

\subsection{Malayalam (28 papers)}
Malayalam, a Dravidian language, is spoken primarily in the Indian state of Kerala. It has over 35 million speakers. In Malayalam, we found papers on text classification \citep{lowres-2022-J-Krishnan}, typographical error correction \citep{lowres-2024-DJ-Ratnam}, named entity recognition \citep{lowres-2023-MK-Harikrishnan}, summarization \citep{lowres-2023-S-K.-Nambiar, lowres-2023-DS-Pankaj}, and question-answering \citep{lowres-2023-R-Rahmath-K}. Interesting finds were papers on machine translation between Sanskrit and Malayalam \citep{lowres-2022-R-Chingamtotattil, lowres-2023-C-Rahul}, along with a language-specific benchmark dataset, MbAbI, for Malayalam text understanding and reasoning \citep{lowres-2023-K-Reji-Rahmath}, which consists of 20 natural language understanding/reasoning tasks, which are similar to the English language Babi benchmark \citep{weston2015aicompletequestionansweringset}. 

In speech processing, we found papers on speech recognition \citep{lowres-2023-K-Manohar-subword, lowres-2023-K-Manohar-multilingual, lowres-2023-K-Manohar-challenges}, grapheme to phoneme conversion \citep{lowres-2022-R-Priyamvada} , and represenation learning \citep{lowres-2023-RK-Thandil}. We also found papers on handwritten character recognition from contemporary contributors \citep{lowres-2024-D-Sudarsan} and palm leaf manuscripts \citep{lowres-2023-KB-Baiju}.

\subsection{Gujarati (21 papers)}
Gujarati, an Indo-Aryan language, is spoken primarily in the Indian state of Gujarat. It has over 55 million speakers. In Gujarati, we found papers in topics such as hate speech and offensive content detection \citep{lowres-2023-T-Ranasinghe, lowres-2023-MK-Sathya}, sentiment analysis on Twitter \citep{lowres-2023-M-Gokani}, reviews \citep{lowres-2022-P-Shah-reviews}, and movies \citep{lowres-2022-P-Shah-film}, text summarization \citep{lowres-2023-R-Kevat-review, lowres-2024-R-Kevat-pagerank, lowres-2022-H-Mehta, lowres-2022-N-Desai}, machine translation \citep{lowres-2022-M-Patel-dsetgens, lowres-2022-M-Patel-gujagra, lowres-2024-D-Ganatra}, and typographical error correction \citep{lowres-2024-BY-Panchal}. 

Similar to other languages, we also found papers on handwritten character recognition, for example, \citep{lowres-2023-DR-Kothadiya}, which uses vision transformers to recognize handwritten characters in Gujarati. Such studies are important, given the lack of access to efficient digital typing tools \citep{hossain-typing-2024} and the prevalence of handwritten documentation. In speech processing, we found papers on spoken digits \citep{lowres-2023-P-Pandit}, speech-to-text models to assist disabled people \citep{lowres-2023-N-Aasofwala}. For a review of speech recognition in Gujarati, we refer the reader to \citet{lowres-2024-M-Dua}. We also found a paper on optical character recognition of various Gujarati language fonts \citep{lowres-2022-K-Joshi}. In terms of applications, we found a paper that implements a model to recommend recipes to cardiac patients \citep{lowres-2023-N-Mehta}.

\subsection{Odia (19 papers)}
Odia, an Indo-Aryan language, is spoken primarily in the Indian state of Odisha. It has over 33 million speakers. We found papers on Hindi-Odia machine translation \citep{lowres-2023-RC-Balabantaray}, named entity recognition \citep{lowres-2023-A-Anandika}, parts of speech tagging \citep{lowres-2023-T-Dalai, lowres-2024-T-Dalai}, hate speech detection \citep{lowres-2024-P-Som}, language modeling \citep{lowres-2023-G-Dey, lowres-2023-GS-Kohli, lowres-2023-P-Agarwal, lowres-2023-S-Parida}, and document summarization \citep{lowres-2022-M-Nayak, lowres-2022-S-Pattnaik}. We alsio found work on handwritten digit \citep{lowres-2023-A-Das-odia} and character \citep{lowres-2022-R-Panda} recognition. 

In speech processing, we found papers on voiced digit recognition \citep{lowres-2022-P-Mohanty-dl, lowres-2022-P-Mohanty-cnn}, automated speech recognition Voiced Odia digit recognition \citep{lowres-2024-MK-Majhi}, and speech emotion recognition \citep{lowres-2022-M-Swain}. \citet{lowres-2023-N-Mishra} gives a detailed review of automated speech recognition in Odia.

\subsection{Sindhi (16 papers)}
Sindhi, an Indo-Aryan language, is spoken primarily in the Sindh province of Pakistan and parts of India. It has over 25 million speakers. In Sindhi, we found papers on corpus analysis \citep{lowres-2023-N-Talpur, lowres-2023-IN-Sodhar}, chunking \citep{lowres-2024-P-Arora}, coreference resolution \citep{lowres-2023-SB-Farooqui}, part of speech tagging \citep{lowres-2024-AA-Memon, lowres-2023-B-Nathani}, grammar \citep{lowres-2023-E-Trumpp}, sentiment analysis \citep{lowres-2022-F-Barakzai, lowres-2023-MB-Alvi, lowres-2023-IN-Sodhar-senti}, morphological analysis \citep{lowres-2023-IN-Sodhar-morpho}, fake news detection \citep{lowres-2023-R-Roshan}. We found a paper on speech reognition \citep{lowres-2022-NK-Bux} but no papers on other multimodal studies. For a survey of NLP resources in Sindhi, we refer the reader to \citet{lowres-2022-A-Rajan}. 

\subsection{Bodo (15 papers)}
Bodo, a Sino-Tibetan language, is spoken primarily in the Indian state of Assam. It has around 3 million speakers. In Bodo, we found papers on corpus creation \citep{lowres-2022-S-Narzary}, hate speech and offensive content detection \citep{lowres-2023-JK-Mim, lowres-2023-K-Ghosh-hate, lowres-2023-T-Ranasinghe}, named entity recognition \citep{lowres-2024-S-Narzary-ner}, part-of-speech tagging \citep{lowres-2024-D-Pathak-bodo, lowres-2023-B-Basumatary}, machine translation \citep{lowres-2023-S-Kalita, lowres-2024-S-Narzary-smt}, word sense disambiguation \citep{lowres-2024-S-Basumatary, lowres-2022-S-Basumatary}, text summarization \citep{lowres-2024-A-Das-bodosum}, and autoregressive text generation \citep{lowres-2023-A-Das-ab-nextword, lowres-2023-A-Das-bodo-nextword}. In speech processing, we found a paper on tonal identification in the Bodo language \citep{lowres-2022-M-Narzary}.

\subsection{Pashto (8 papers)}
Pashto, an Indo-Iranian language, is spoken mainly in Afghanistan and Pakistan. It has around 50 million speakers. In Pashto, we found papers on handwriting recognition \citep{lowres-2022-I-Hussain, lowres-2023-F-Khaliq}, 
sign language recognition \citep{lowres-2022-AF-Shokoori}, offensive language detection \citep{lowres-2023-I-Haq-hate}, poetry generation \citep{lowres-2024-I-Ullah}, speech recognition \citep{lowres-2024-I-Ahmed}, and alphabet analysis \citep{lowres-2023-PN-Nasrat}. \citet{lowres-2023-I-Haq-toolkit} provides a toolkit for Pashto NLP.

\subsection{Burmese (8 papers)}
Burmese, a Sino-Tibetan language, is spoken primarily in Myanmar. It has over 32 million speakers. In Burmese, we found papers on machine translation between Burmese and other languages such as Wa and Rohingya \citep{lowres-2023-F-Yune, lowres-2022-TM-Oo, lowres-2023-TM-Oo}, multilingualism studies \citep{lowres-2022-J-Li}, and medical automatic speech recognition \citep{lowres-2024-HM-Htun}. We also found a clinical microbiology dataset \citep{lowres-2024-M-Si-Thu} in Burmese.

\subsection{Konkani (5 papers)}
Konkani, an Indo-Aryan language, is spoken primarily in the Indian states of Goa, Maharashtra, Karnataka and Kerala. It has around 3 million speakers. In Konkani, we found papers on text summarization \citep{lowres-2023-C-More, lowres-2022-J-D'Silva, lowres-2023-J-D'Silva} and machine translation \citep{lowres-2023-BS-Kamath}. For a survey on NLP resources in Konkani, we refer the reader to \citet{lowres-2022-A-Rajan}. 

\subsection{Khasi (4 papers)}
Khasi, a Sino-Tibetan language, is spoken primarily in the Indian state of Meghalaya. It has around 1.5 million speakers. In Khasi we found papers on machine translation \citep{lowres-2024-AV-Hujon}, speech recognition \citep{lowres-2022-F-Rynjah, lowres-2024-S-Deepajothi}, and word embeddings \citep{lowres-2022-NDJ-Thabah}. 

\subsection{Bhojpuri (4 papers)}
Bhojpuri, an Indo-Aryan language, is spoken mainly in India and Nepal. It has around 55 million speakers. In Bhojpuri, we found papers on named entity recognition \citep{lowres-2023-R-Mundotiya}, translating Hindi synsets to Bhojpuri \citep{lowres-2023-I-Ali}, speech corpora \citep{lowres-2022-R-Kumar}, and part of speech tagging and chunking \citep{lowres-2022-RK-Mundotiya}.

\subsection{Kashmiri (3 papers)}
Kashmiri, an Indo-Aryan language, is spoken primarily in the Indian state of Jammu and Kashmir and in Pakistani-administered Kashmir. It has around 7 million speakers. In Kashmiri, we were only able to find dataset-related papers \citep{lowres-2024-SMU-Qumar-emerging, lowres-2022-NA-Lone, lowres-2024-SMU-Qumar-parallel}.

\subsection{Meitei (2 papers)}
Meitei, a Sino-Tibetan language, is spoken primarily in the Indian state of Manipur. It has around 2.5 million speakers. In Meitei, we found 1 paper on handwriting recognition \citep{lowres-2024-D-Hijam} and 1 paper on speech recognition \citep{lowres-2023-H-Pangsatabam}. 

\subsection{Chattisgarhi, Dhivehi, Sairaki, Rajasthani, and Balochi (1 paper each)}
Chhattisgarhi, an Indo-Aryan language, is spoken primarily in the Indian state of Chhattisgarh. It has around 16 million speakers. We found only 1 paper in Chhattisgarhi, which was a text-to-speech model trained on 20 hours of data from 2 speakers \citep{lowres-2023-A-Singh-chhattis}.

Dhivehi, an Indo-Aryan language, is the official language of the Maldives. It has around 350,000 speakers. We found a paper that improved automatic speech recognition \citep{ahmed-2023-improving} in Dhivehi using sub-word modeling, language model decoding, and automatic spelling correction.

Saraiki, an Indo-Aryan language, is spoken primarily in Pakistan. It has around 30 million speakers. We found a paper on a part-of-speech tag set in Saraiki \citep{lowres-2023-MH-Malik}. 

Rajasthani, an Indo-Aryan language, is spoken primarily in the Indian state of Rajasthan. It has around 40 million speakers. We found a paper on the dialect recognition of the Bagri dialect of the Rajasthani language using an optimized feature swarm convolutional neural network (CNN) \citep{lowres-2024-P-Kukana}.

Balochi, an Iranian language, is spoken primarily in Pakistan and Iran. It has around 10 million speakers. We found a paper on parts-of-speech tagging in Balochi using conditional random fields \citep{lowres-2024-S-Ullah}. 

We were unable to find any papers for Dari, Rohingya, Kurukh, and Santali, although they were a part of our search. This might be because of the fact that Dari is very similar to Persian and the Rohingya language has a different name that did not show up in our search results. We tried searching again in Google Scholar by replacing ``Rohingya'' with ``Rakhine,'' but it did not yield any relevant search results. For Kurukh and Santali, a deeper search focused on these languages might yield some relevant results.
\section{Conclusion}
In this paper, we reviewed the recent research and development in South Asian language processing, prioritizing breadth of coverage over its depth. We emphasized key research trends at a field-specific level (text, speech, multimodal) in Sections \ref{sec:text}, \ref{sec:multimodal}, and \ref{sec:speech}, and at a language-specific level in Section \ref{sec:lowres} with our low resource language spotlight survey, which covered 21 languages that are typically seen in the ``long tail'' of South Asian language processing and language processing in general. Such languages are masked by other higher-resource languages in papers and surveys on multilingual language processing. Finally, we hope that this survey inspires further investigations into the models, tasks, and benchmarks described here.

\section*{Limitations}
Our survey shows that research and development in South Asian languages is thriving, and in many South Asian languages, researchers are trying to or have successfully replicated the successes of language, speech, and multimodal models currently enjoyed by high resource languages. However, in many areas, there is a clear gap among high-resource and low-resource languages. In particular, we wish to present the following observations:
\begin{itemize}
    \item We see that South Asian language publications are restricted to smaller venues, making it difficult for researchers to discover, share and collaborate on models, datasets, and tasks. We also see that in lower-resourced languages, there is a lack of awareness in the research community, given the lack of diversity in the number of distinct authors in publications related to language processing in those languages. We discovered many benchmarks that are typically not included in multilingual LLM evaluations, and it would be helpful for the community to combine these lesser-known benchmarks into an aggregated benchmark that can be hosted on popular platforms such as Huggingface.
    \item There has been some focus on sign languages and South Asian demographics with disabilities. However, such sign languages tend to be highly localized in South Asian communities and we need to perform more extensive surveys of the entire spectrum of sign languages in south Asia, as we build AI-based assistance tools for them.
    \item Most South Asian languages covered in our survey have some literature on code-mixing/switching, but we notice a lack of comparative studies on code-mixing in various South Asian languages. 
    \item Meta-analysis using cutting-edge language models such as O1 and GPT-4o is promising and provides a strong baseline in the absence of sufficient labeled data, however, problems such as lack of cohesion within predicted topics and low precision/recall persist. Future efforts must focus on improving the precision and recall metrics of such meta-analyses, along with improving the coherence of LLM-generated topics. 
    \item There is a need for a consistent ethics framework that is based in the unique cultural contexts of South Asian communities. For example, while depression detection and cyberbullying detection are important applications of language models, we must ensure that appropriate care is taken before these models are deployed in a real-world setting. 
\end{itemize}

\bibliography{anthology,custom}

\begin{thebibliography}{372}
\expandafter\ifx\csname natexlab\endcsname\relax\def\natexlab#1{#1}\fi

\bibitem[{Aasofwala et~al.(2023)Aasofwala, Verma, and Patel}]{lowres-2023-N-Aasofwala}
Nasrin Aasofwala, Shanti Verma, and Kalyani Patel. 2023.
\newblock \href {https://doi.org/10.1109/ICCCNT56998.2023.10308284} {Nlp based model to convert english speech to gujarati text for deaf and dumb people}.
\newblock In \emph{2023 14th International Conference on Computing Communication and Networking Technologies (ICCCNT)}, pages 1--6.

\bibitem[{Abirami et~al.(2024)Abirami, Leong, Rengarajan, Anitha, Suganya, Singh, Sarveswaran, Tjhi, and Shah}]{abirami-etal-2024-aalamaram}
A~M Abirami, Wei~Qi Leong, Hamsawardhini Rengarajan, D~Anitha, R~Suganya, Himanshu Singh, Kengatharaiyer Sarveswaran, William~Chandra Tjhi, and Rajiv~Ratn Shah. 2024.
\newblock \href {https://aclanthology.org/2024.wildre-1.11} {Aalamaram: A large-scale linguistically annotated treebank for the {T}amil language}.
\newblock In \emph{Proceedings of the 7th Workshop on Indian Language Data: Resources and Evaluation}, pages 73--83, Torino, Italia. ELRA and ICCL.

\bibitem[{Achamaleh et~al.(2024)Achamaleh, Kawo, Batyrshini, and Sidorov}]{lowres-2024-T-Achamaleh}
Tewodros Achamaleh, Lemlem Kawo, Ildar Batyrshini, and Grigori Sidorov. 2024.
\newblock \href {https://aclanthology.org/2024.dravidianlangtech-1.15} {Tewodros@{D}ravidian{L}ang{T}ech 2024: Hate speech recognition in {T}elugu codemixed text}.
\newblock In \emph{Proceedings of the Fourth Workshop on Speech, Vision, and Language Technologies for Dravidian Languages}, pages 96--100, St. Julian's, Malta. Association for Computational Linguistics.

\bibitem[{Acharya(2022)}]{lowres-2022-A-Acharya}
A~Acharya. 2022.
\newblock \href {https://aayushacharya.com.np/docs/be_thesis.pdf} {Automatic speech recognition and classification of nepali speech}.
\newblock \emph{Tribhuvan University}.

\bibitem[{Adhikari et~al.(2022)Adhikari, Thapa, Naseem, Singh, Huo, Bharathy, and Prasad}]{lowres-2022-S-Adhikari}
Surabhi Adhikari, Surendrabikram Thapa, Usman Naseem, Priyanka Singh, Huan Huo, Gnana Bharathy, and Mukesh Prasad. 2022.
\newblock \href {https://doi.org/https://doi.org/10.1016/j.ijhcs.2021.102761} {Exploiting linguistic information from nepali transcripts for early detection of alzheimer's disease using natural language processing and machine learning techniques}.
\newblock \emph{International Journal of Human-Computer Studies}, 160:102761.

\bibitem[{Afzal et~al.(2023)Afzal, Shardlow, Tuarob, Zaman, Sarwar, Ali, Aljohani, Lytras, Nawaz, and Hassan}]{multimodal-2023-MK-Afzal}
Muhammad~Kashif Afzal, Matthew Shardlow, Suppawong Tuarob, Farooq Zaman, Raheem Sarwar, Mohsen Ali, Naif~Radi Aljohani, Miltiades~D Lytras, Raheel Nawaz, and Saeed-Ul Hassan. 2023.
\newblock Generative image captioning in urdu using deep learning.
\newblock \emph{J. Ambient Intell. Humaniz. Comput.}, 14(6):7719--7731.

\bibitem[{Agarwal et~al.(2023)Agarwal, Asif, Parida, and Sekhar}]{lowres-2023-P-Agarwal}
P~Agarwal, A~Asif, S~Parida, and S~Sekhar. 2023.
\newblock \href {https://ieeexplore.ieee.org/abstract/document/10291329/?casa_token=m0fbBdYapXsAAAAA:D3o-YRViTOGG5o0t9R-4rd3q1wBrnijdTHoEmzlQ_ze6oCoPQY52MaTmUK-kzhaZMsK93FeA} {Generative chatbot adaptation for odia language: a critical evaluation}.
\newblock \emph{… on Circuits, Power …}.

\bibitem[{Ahmed(2023)}]{ahmed-2023-improving}
Arushad Ahmed. 2023.
\newblock \href {https://aclanthology.org/2023.icnlsp-1.27} {Improving {D}hivehi automatic speech recognition ({ASR}) with sub-word modelling, language model decoding and automatic spelling correction}.
\newblock In \emph{Proceedings of the 6th International Conference on Natural Language and Speech Processing (ICNLSP 2023)}, pages 256--265, Online. Association for Computational Linguistics.

\bibitem[{Ahmed et~al.(2024)Ahmed, Irfan, Iqbal, and Khalil}]{lowres-2024-I-Ahmed}
I~Ahmed, MA~Irfan, A~Iqbal, and A~Khalil. 2024.
\newblock \href {https://link.springer.com/article/10.1007/s11042-023-17684-w} {Efficient feature extraction and classification for the development of pashto speech recognition system}.
\newblock \emph{Multimedia Tools and …}.

\bibitem[{Alam et~al.(2022)Alam, Sushmit, Abdullah, Nakkhatra, Ansary, Hossen, Mehnaz, Reasat, and Humayun}]{speech-2022-Alam}
Samiul Alam, Asif Sushmit, Zaowad Abdullah, Shahrin Nakkhatra, MD.~Nazmuddoha Ansary, Syed~Mobassir Hossen, Sazia~Morshed Mehnaz, Tahsin Reasat, and Ahmed~Imtiaz Humayun. 2022.
\newblock \href {http://arxiv.org/abs/2206.14053} {Bengali common voice speech dataset for automatic speech recognition}.

\bibitem[{Ali and Gatla(2023)}]{lowres-2023-I-Ali}
I~Ali and P~Gatla. 2023.
\newblock \href {https://aclanthology.org/2023.ranlp-1.7/} {Bhojpuri wordnet: Problems in translating hindi synsets into bhojpuri}.
\newblock \emph{Proceedings of the 14th International Conference on …}.

\bibitem[{Alumäe et~al.(2023)Alumäe, Kong, and Robnikov}]{speech-2023-Alumäe}
Tanel Alumäe, Jiaming Kong, and Daniil Robnikov. 2023.
\newblock \href {https://doi.org/10.1109/ASRU57964.2023.10389668} {Dialect adaptation and data augmentation for low-resource asr: Taltech systems for the madasr 2023 challenge}.
\newblock In \emph{2023 IEEE Automatic Speech Recognition and Understanding Workshop (ASRU)}, pages 1--7.

\bibitem[{Alvi et~al.(2023)Alvi, Mahoto, Reshan, and Unar}]{lowres-2023-MB-Alvi}
MB~Alvi, NA~Mahoto, MSA Reshan, and M~Unar. 2023.
\newblock \href {https://journals.sagepub.com/doi/abs/10.1177/21582440231197452} {Count me too: Sentiment analysis of roman sindhi script}.
\newblock \emph{SAGE …}.

\bibitem[{Anagha et~al.(2023)Anagha, Sairam, and Mahesh}]{lowres-2023-HM-Anagha}
HM~Anagha, K~Sairam, and J~Mahesh. 2023.
\newblock \href {https://link.springer.com/chapter/10.1007/978-981-99-9531-8_23} {Paraphrase generation and deep learning models for paraphrase detection in a low-resourced language: Kannada}.
\newblock \emph{… Conference on Advances …}.

\bibitem[{Anandika and Chakravarty(2023)}]{lowres-2023-A-Anandika}
A~Anandika and S~Chakravarty. 2023.
\newblock \href {https://www.inderscienceonline.com/doi/abs/10.1504/IJRIS.2023.128379} {Named entity recognition in odia language: a rule-based approach}.
\newblock \emph{International Journal of …}.

\bibitem[{Aparna et~al.(2023)Aparna, Srivatsa, and Madhavan}]{lowres-2023-M-Aparna}
M~Aparna, S~Srivatsa, and G~Sai Madhavan. 2023.
\newblock \href {https://link.springer.com/chapter/10.1007/978-3-031-58502-9_1} {Ai-based assistance for management of oral community knowledge in low-resource and colloquial kannada language}.
\newblock \emph{… Conference on Big Data …}.

\bibitem[{Ardila et~al.(2020)Ardila, Branson, Davis, Henretty, Kohler, Meyer, Morais, Saunders, Tyers, and Weber}]{ardila2020commonvoicemassivelymultilingualspeech}
Rosana Ardila, Megan Branson, Kelly Davis, Michael Henretty, Michael Kohler, Josh Meyer, Reuben Morais, Lindsay Saunders, Francis~M. Tyers, and Gregor Weber. 2020.
\newblock \href {http://arxiv.org/abs/1912.06670} {Common voice: A massively-multilingual speech corpus}.

\bibitem[{Arora et~al.(2024)Arora, Nathani, Joshi, and Katyayan}]{lowres-2024-P-Arora}
P~Arora, B~Nathani, N~Joshi, and P~Katyayan. 2024.
\newblock \href {https://link.springer.com/chapter/10.1007/978-3-031-71484-9_10} {Development of rule-based chunker for sindhi}.
\newblock \emph{… of Artificial Intelligence and …}.

\bibitem[{Awale et~al.(2022)Awale, Prasai, Rijal, and Basnet}]{lowres-2022-S-Awale}
S~Awale, S~Prasai, B~Rijal, and SB~Basnet. 2022.
\newblock \href {https://www.academia.edu/download/92471977/34797.pdf} {Preprocessing of nepali news corpus for downstream tasks}.
\newblock \emph{Nepalese Linguistics}.

\bibitem[{Babu et~al.(2022)Babu, Nalajala, and Sarada}]{lowres-2022-B-Sankara-Babu}
B~Sankara Babu, S~Nalajala, and K~Sarada. 2022.
\newblock \href {https://link.springer.com/chapter/10.1007/978-3-030-76653-5_12} {Machine learning based online handwritten telugu letters recognition for different domains}.
\newblock \emph{… of Artificial Intelligence …}.

\bibitem[{Bai et~al.(2023)Bai, Prerana, and Benki}]{lowres-2023-BGM-Bai}
BGM Bai, P~Prerana, and AA~Benki. 2023.
\newblock \href {https://ieeexplore.ieee.org/abstract/document/10199524/?casa_token=7LrZ1ArfD9QAAAAA:-K35aiEbz6t2UbU-FyGKoHiDjjlV-onyXVJnPiNcCL07sxcBTyxKoOGX0sEO0KRxxcsTqqE8} {Virtual medical assistant in english and kannada languages}.
\newblock \emph{… Conference on Applied …}.

\bibitem[{Baiju(2023)}]{lowres-2023-KB-Baiju}
KB~Baiju. 2023.
\newblock \href {http://scholar.uoc.ac.in/handle/20.500.12818/1461} {Pattern primitive based malayalam handwritten character recognition studies for real-time applications}.
\newblock \emph{University of Calicut}.

\bibitem[{Balabantaray et~al.(2023)Balabantaray, Paul, and Raj}]{lowres-2023-RC-Balabantaray}
RC~Balabantaray, NR~Paul, and M~Raj. 2023.
\newblock \href {https://ceur-ws.org/Vol-3681/T9-2.pdf} {Hindi-odia machine translation system.}
\newblock \emph{FIRE (Working Notes)}.

\bibitem[{Balaji et~al.(2023)Balaji, T, and B}]{balaji-etal-2023-nlp}
Varsha Balaji, Shahul~Hameed T, and Bharathi B. 2023.
\newblock \href {https://aclanthology.org/2023.dravidianlangtech-1.17} {{NLP}{\_}{SSN}{\_}{CSE}@{D}ravidian{L}ang{T}ech: Fake news detection in {D}ravidian languages using transformer models}.
\newblock In \emph{Proceedings of the Third Workshop on Speech and Language Technologies for Dravidian Languages}, pages 133--139, Varna, Bulgaria. INCOMA Ltd., Shoumen, Bulgaria.

\bibitem[{Balouchzahi et~al.(2022)Balouchzahi, Butt, Hegde, and Ashraf}]{lowres-2022-F-Balouchzahi}
F~Balouchzahi, S~Butt, A~Hegde, and N~Ashraf. 2022.
\newblock \href {https://aclanthology.org/2022.icon-wlli.8/} {Overview of coli-kanglish: Word level language identification in code-mixed kannada-english texts at icon 2022}.
\newblock \emph{… Code-mixed Kannada …}.

\bibitem[{Bansod(2023)}]{llm-2023-Bansod}
PP~Bansod. 2023.
\newblock \href {https://scholarworks.sjsu.edu/etd_projects/1265/} {Hate speech detection in hindi}.
\newblock \emph{San Jose State University}.

\bibitem[{Barakzai et~al.(2022)Barakzai, Bhatti, and Saddar}]{lowres-2022-F-Barakzai}
F~Barakzai, S~Bhatti, and S~Saddar. 2022.
\newblock \href {https://ieeexplore.ieee.org/abstract/document/10000519/?casa_token=-iDWKNh7Ys0AAAAA:3ZcE-Rz5umeaXg9vx1-7g4Zou4dtOG0v_LqpkKVnj43ltilr9VyqUf6TFanRVAHO5hnqB-B-} {Sentiment analysis of sindhi news articles using deep learning}.
\newblock \emph{2022 IEEE 17th International …}.

\bibitem[{Baruah et~al.(2024{\natexlab{a}})Baruah, Singh, and Sarmah}]{lowres-2024-H-Baruah-roman}
H~Baruah, SR~Singh, and P~Sarmah. 2024{\natexlab{a}}.
\newblock \href {https://aclanthology.org/2024.lrec-main.143/} {Assamesebacktranslit: Back transliteration of romanized assamese social media text}.
\newblock \emph{Proceedings of the 2024 Joint …}.

\bibitem[{Baruah et~al.(2024{\natexlab{b}})Baruah, Singh, and Sarmah}]{lowres-2024-H-Baruah-social}
H~Baruah, SR~Singh, and P~Sarmah. 2024{\natexlab{b}}.
\newblock \href {https://dl.acm.org/doi/abs/10.1145/3639565?casa_token=kdjjKpLarocAAAAA:j3zY56weO_ZGUOdlYxRsPi3GNe2_H7PZCcCd4xKY_smLoPbh1eLZOq4n5t_gR-7w5nxI100M78mu} {Transliteration characteristics in romanized assamese language social media text and machine transliteration}.
\newblock \emph{ACM Transactions on Asian and Low Resource Languages}.

\bibitem[{Baruah et~al.(2023)Baruah, Gogoi, and Neog}]{lowres-2023-N-Baruah-hate}
N~Baruah, A~Gogoi, and M~Neog. 2023.
\newblock \href {https://link.springer.com/chapter/10.1007/978-981-99-3485-0_52} {Detection of hate speech in assamese text}.
\newblock \emph{International Conference on Communication …}.

\bibitem[{Baruah et~al.(2022)Baruah, Sarma, Borkotokey, and Borah}]{lowres-2022-N-Baruah}
N~Baruah, SK~Sarma, S~Borkotokey, and R~Borah. 2022.
\newblock \href {https://link.springer.com/chapter/10.1007/978-981-19-3015-7_1} {A graph-based extractive assamese text summarization}.
\newblock \emph{… Methods and Data …}.

\bibitem[{Basumatary et~al.(2023)Basumatary, Rahman, and Sarma}]{lowres-2023-B-Basumatary}
B~Basumatary, M~Rahman, and SK~Sarma. 2023.
\newblock \href {https://ieeexplore.ieee.org/abstract/document/10308365/?casa_token=2TlyzwAOMQ8AAAAA:4StQscvyS6TDv4o7MBnTZViR_5MKmGyBebcNH0OFprBov7pCe5bSfzH64f0jPqJoBgLYu07n} {Deep learning based bodo parts of speech tagger}.
\newblock \emph{2023 14th …}.

\bibitem[{Basumatary and Barman(2024)}]{lowres-2024-S-Basumatary}
S~Basumatary and M~Barman. 2024.
\newblock \href {https://journal.uob.edu.bh/handle/123456789/5822} {Word sense disambiguation task for bodo language using attention based deep cnn architecture}.
\newblock \emph{… of Computing and …}.

\bibitem[{Basumatary et~al.(2022)Basumatary, Brahma, and Barman}]{lowres-2022-S-Basumatary}
S~Basumatary, K~Brahma, and AK~Barman. 2022.
\newblock \href {https://link.springer.com/chapter/10.1007/978-981-99-6866-4_37} {An approach to bodo word sense disambiguation (wsd) using word2vec}.
\newblock \emph{… Conference on Modeling …}.

\bibitem[{Bhagath et~al.(2023{\natexlab{a}})Bhagath, Lasya, and Dhyeya}]{lowres-2023-P-Bhagath-iot}
P~Bhagath, V~Lasya, and P~Dhyeya. 2023{\natexlab{a}}.
\newblock \href {https://ieeexplore.ieee.org/abstract/document/10482996/?casa_token=3G9OZ02RDjgAAAAA:TIJcVY2hRLxFOkcBZ9kNo0Ehnf5RQd30ubqshxnc5qAKN-LEOQoHXbWGgmgd7DXlVjhzynHW} {Telugu vakyalu: Spoken telugu sentences for iot applications}.
\newblock \emph{2023 26th Conference of}.

\bibitem[{Bhagath et~al.(2022)Bhagath, Pullagura, and Das}]{lowres-2022-P-Bhagath-corpora}
P~Bhagath, M~Pullagura, and PK~Das. 2022.
\newblock \href {https://ieeexplore.ieee.org/abstract/document/9854065/?casa_token=jv1vSSM3OKQAAAAA:6ASefIx0wFYTXpYDCSAIRjqxofHsIRKjFUOLzrAc3hFM3fjsX5vEsedTK2iQkZEs7WvDIvAm} {Telugu ankelu: A telugu spoken digits corpora for mobile speech recognition}.
\newblock \emph{2022 12th …}.

\bibitem[{Bhagath et~al.(2023{\natexlab{b}})Bhagath, Rao, and Ram}]{lowres-2023-P-Bhagath}
P~Bhagath, AUM Rao, and BS~Ram. 2023{\natexlab{b}}.
\newblock \href {https://ieeexplore.ieee.org/abstract/document/10178989/?casa_token=f1koanIux5EAAAAA:qdCY70sT1YxiOOVUAflMxODBKzDQacVe2OSVx9I_Mj90YXMX4GDUXshnD8X_bGxstBiiLa22} {Telugu spoken digits modeling using convolutional neural networks}.
\newblock \emph{2023 IEEE 13th …}.

\bibitem[{Bhatt et~al.(2021)Bhatt, Goyal, Dandapat, Choudhury, and Sitaram}]{bhatt-etal-2021-universality}
Shaily Bhatt, Poonam Goyal, Sandipan Dandapat, Monojit Choudhury, and Sunayana Sitaram. 2021.
\newblock \href {https://aclanthology.org/2021.icon-main.15} {On the universality of deep contextual language models}.
\newblock In \emph{Proceedings of the 18th International Conference on Natural Language Processing (ICON)}, pages 106--119, National Institute of Technology Silchar, Silchar, India. NLP Association of India (NLPAI).

\bibitem[{Bhattacharyya et~al.(2023)Bhattacharyya, Mondal, Maji, and Bhattacharya}]{bhattacharyya-etal-2023-vacaspati}
Pramit Bhattacharyya, Joydeep Mondal, Subhadip Maji, and Arnab Bhattacharya. 2023.
\newblock \href {https://doi.org/10.18653/v1/2023.ijcnlp-main.72} {{VACASPATI}: A diverse corpus of {B}angla literature}.
\newblock In \emph{Proceedings of the 13th International Joint Conference on Natural Language Processing and the 3rd Conference of the Asia-Pacific Chapter of the Association for Computational Linguistics (Volume 1: Long Papers)}, pages 1118--1130, Nusa Dua, Bali. Association for Computational Linguistics.

\bibitem[{Bhaumik(2023)}]{llm-2023-Bhaumik}
A~K Bhaumik. 2023.
\newblock Sarcasm detection in dravidian code-mixed text using transformer-based models.
\newblock In \emph{Forum for Information Retrieval Evaluation (FIRE)}, Goa, India.

\bibitem[{Bhogale et~al.(2023)Bhogale, Raman, Javed, Doddapaneni, Kunchukuttan, Kumar, and Khapra}]{speech-2023-Bhogale}
Kaushal Bhogale, Abhigyan Raman, Tahir Javed, Sumanth Doddapaneni, Anoop Kunchukuttan, Pratyush Kumar, and Mitesh~M. Khapra. 2023.
\newblock \href {https://doi.org/10.1109/ICASSP49357.2023.10096933} {Effectiveness of mining audio and text pairs from public data for improving asr systems for low-resource languages}.
\newblock In \emph{ICASSP 2023 - 2023 IEEE International Conference on Acoustics, Speech and Signal Processing (ICASSP)}, pages 1--5.

\bibitem[{Bhuvaneshwari and JyothiRani(2023)}]{lowres-2023-K-Bhuvaneshwari}
K~Bhuvaneshwari and SA~JyothiRani. 2023.
\newblock \href {https://link.springer.com/chapter/10.1007/978-3-031-68617-7_22} {Summarization of telugu text discourses}.
\newblock \emph{… on Artificial Intelligence …}.

\bibitem[{Bora et~al.(2023)Bora, Dehingia, Boruah, and Chetia}]{lowres-2023-J-Bora}
J~Bora, S~Dehingia, A~Boruah, and AA~Chetia. 2023.
\newblock \href {https://www.sciencedirect.com/science/article/pii/S1877050923001175} {Real-time assamese sign language recognition using mediapipe and deep learning}.
\newblock \emph{Procedia Computer …}.

\bibitem[{Borah et~al.(2023)Borah, Baruah, Ramakrishna, and Kumar}]{lowres-2023-N-Borah}
N~Borah, U~Baruah, MT~Ramakrishna, and VV~Kumar. 2023.
\newblock \href {https://ieeexplore.ieee.org/abstract/document/10206436/} {Efficient assamese word recognition for societal empowerment: A comparative feature-based analysis}.
\newblock \emph{IEEE …}.

\bibitem[{Budhathoki and Timilsina(2023)}]{lowres-2023-R-Budhathoki}
R~Budhathoki and S~Timilsina. 2023.
\newblock \href {https://www.researchgate.net/profile/Suresh-Timilsina/publication/376283001_Image_Captioning_in_Nepali_Using_CNN_and_Transformer_Decoder/links/6572753acbd2c535ea04ad5c/Image-Captioning-in-Nepali-Using-CNN-and-Transformer-Decoder.pdf} {Image captioning in nepali using cnn and transformer decoder}.
\newblock \emph{Journal of Engineering and Sciences}.

\bibitem[{Burramsetty and Gonugunta(2022)}]{lowres-2022-SS-Burramsetty}
SS~Burramsetty and NP~Gonugunta. 2022.
\newblock \href {https://ieeexplore.ieee.org/abstract/document/9835861/?casa_token=Y0w16PkDVCkAAAAA:FWsgKx9_HuF2UPHBgoNqapRHjzuSpaWAH0d4CnXPsHjFavkqD9f7bR8YiZ0eoRF-t2e-ZTrb} {Event extraction from telugu-english code mixed social media text}.
\newblock \emph{2022 7th International …}.

\bibitem[{Bux et~al.(2022)Bux, Khan, and Bakhsh}]{lowres-2022-NK-Bux}
NK~Bux, A~Khan, and K~Bakhsh. 2022.
\newblock \href {https://www.academia.edu/download/103544413/10.pdf} {Speech to text by using the sindhi language}.
\newblock \emph{… Journal of Artificial Intelligence and …}.

\bibitem[{Chandrika and Kallimani(2022)}]{lowres-2022-CP-Chandrika-authorship-attribution}
CP~Chandrika and JS~Kallimani. 2022.
\newblock \href {https://www.researchgate.net/profile/Chandrika-Prasad/publication/364173203_Authorship_Attribution_on_Kannada_Text_using_Bi-Directional_LSTM_Technique/links/639aae8411e9f00cda47067c/Authorship-Attribution-on-Kannada-Text-using-Bi-Directional-LSTM-Technique.pdf} {Authorship attribution on kannada text using bi-directional lstm technique}.
\newblock \emph{International Journal of Advanced …}.

\bibitem[{Changrampadi et~al.(2022)Changrampadi, Shahina, Narayanan, and Khan}]{speech-2022-Changrampadi}
Mohamed~Hashim Changrampadi, A.~Shahina, M.~Badri Narayanan, and A.~Nayeemulla Khan. 2022.
\newblock \href {https://doi.org/10.32604/iasc.2022.022021} {End-to-end speech recognition of tamil language}.
\newblock \emph{Intelligent Automation \& Soft Computing}, 32(2):1309--1323.

\bibitem[{Chen et~al.(2023{\natexlab{a}})Chen, Zhang, Zhang, Qu, and Yang}]{speech-2023-Chen-meta}
Yaqi Chen, Wenlin Zhang, Hao Zhang, Dan Qu, and Xu-Kui Yang. 2023{\natexlab{a}}.
\newblock \href {https://doi.org/10.1145/3626187} {Task-based meta focal loss for multilingual low-resource speech recognition}.
\newblock \emph{ACM Trans. Asian Low-Resour. Lang. Inf. Process.}, 22(11).

\bibitem[{Chen et~al.(2023{\natexlab{b}})Chen, Bapna, Rosenberg, Zhang, Ramabhadran, Moreno, and Chen}]{speech-2023-chen}
Zhehuai Chen, Ankur Bapna, Andrew Rosenberg, Yu~Zhang, Bhuvana Ramabhadran, Pedro Moreno, and Nanxin Chen. 2023{\natexlab{b}}.
\newblock \href {https://doi.org/10.1109/SLT54892.2023.10022791} {Maestro-u: Leveraging joint speech-text representation learning for zero supervised speech asr}.
\newblock In \emph{2022 IEEE Spoken Language Technology Workshop (SLT)}, pages 68--75.

\bibitem[{Chethas et~al.(2023)Chethas, Ankita, and Apoorva}]{lowres-2023-K-Chethas}
K~Chethas, V~Ankita, and BS~Apoorva. 2023.
\newblock \href {https://ieeexplore.ieee.org/abstract/document/10170312/?casa_token=kXp7WUPrHogAAAAA:Db4-3ZuBBdvCnl7CWHoL6bTY1Ro_-ELpljDb4Ax0pzRY4zlB0ouZ-KJtIifZyYrT4HXOrK60} {Image caption generation in kannada using deep learning frameworks}.
\newblock \emph{… on Advances in …}.

\bibitem[{Chingamtotattil and Gopikakumari(2022)}]{lowres-2022-R-Chingamtotattil}
R~Chingamtotattil and R~Gopikakumari. 2022.
\newblock \href {https://www.academia.edu/download/94767512/52_28117_v28i3_Dec22.pdf} {Neural machine translation for sanskrit to malayalam using morphology and evolutionary word sense disambiguation}.
\newblock \emph{Indonesian Journal of Electrical …}.

\bibitem[{Choudhury and Sharma(2023)}]{lowres-2023-N-Choudhury}
N~Choudhury and U~Sharma. 2023.
\newblock \href {https://search.ebscohost.com/login.aspx?direct=true&profile=ehost&scope=site&authtype=crawler&jrnl=07650019&AN=173376353&h=7effNlhRBE%2B3y9jCSq68fQg7RxMLOC6Rvawbt7XDHXsEhhsUupwdjJ7Tt68SUhFFc3jt8tOyTzDLn84t7Ltfzg%3D%3D&crl=c} {Enhanced emotion recognition from spoken assamese dialect: A machine learning approach with language-independent features.}
\newblock \emph{Traitement du Signal}.

\bibitem[{Choudhury et~al.(2023)Choudhury, Guha, and Nandi}]{multimodal-2023-P-Choudhury}
Pankaj Choudhury, Prithwijit Guha, and Sukumar Nandi. 2023.
\newblock \href {https://doi.org/10.1109/IALP61005.2023.10337310} {Relevance of language-specific training on image caption synthesis for low resource assamese language}.
\newblock In \emph{2023 International Conference on Asian Language Processing (IALP)}, pages 13--18.

\bibitem[{CoRover.ai(2024)}]{corover-bharatgpt}
CoRover.ai. 2024.
\newblock {BharatGPT: Bridging the Multilingual Gap}.
\newblock \url{https://www.corover.ai/}.

\bibitem[{Dalai et~al.(2023)Dalai, Mishra, and Sa}]{lowres-2023-T-Dalai}
T~Dalai, TK~Mishra, and PK~Sa. 2023.
\newblock \href {https://dl.acm.org/doi/abs/10.1145/3588900?casa_token=vn7s8z15YdIAAAAA:AVqKwF7vWYsUnuHLstoLM8xgh6LlNp-CSuC3s-CDdJxvjt7lunUMqQO-6OPq5CxmLsxdEx_5IxUq} {Part-of-speech tagging of odia language using statistical and deep learning based approaches}.
\newblock \emph{ACM Transactions on Asian and Low …}.

\bibitem[{Dalai et~al.(2024)Dalai, Mishra, and Sa}]{lowres-2024-T-Dalai}
T~Dalai, TK~Mishra, and PK~Sa. 2024.
\newblock \href {https://dl.acm.org/doi/abs/10.1145/3637877?casa_token=0x9I_rTLpnAAAAAA:yglpUiM1XTUa76D0F4zgZjtybNDWpkaqAd5uCNBNwNUzYir2z1K5196bRyuxKHG8psGPXv-0VUbc} {Deep learning-based pos tagger and chunker for odia language using pre-trained transformers}.
\newblock \emph{ACM Transactions on Asian and Low …}.

\bibitem[{Das et~al.(2023{\natexlab{a}})Das, Baruah, and Roy}]{lowres-2023-A-Das-ab-nextword}
A~Das, A~Baruah, and S~Roy. 2023{\natexlab{a}}.
\newblock \href {https://search.ebscohost.com/login.aspx?direct=true&profile=ehost&scope=site&authtype=crawler&jrnl=00222755&AN=164884180&h=z5PdgKR62918rDppVZcsXvYse9oRaIO4ufxjUh6e4xlQWZ%2BCfeaqaU4vXaCUetNQPslSNdIqTjk3UGPz8UdUEg%3D%3D&crl=c} {Assamese and bodo next word detection using lstm.}
\newblock \emph{Journal of Mines, Metals and Fuels}.

\bibitem[{Das et~al.(2023{\natexlab{b}})Das, Baruah, and Roy}]{lowres-2023-A-Das-bodo-nextword}
A~Das, A~Baruah, and S~Roy. 2023{\natexlab{b}}.
\newblock \href {https://link.springer.com/chapter/10.1007/978-3-031-47224-4_20} {Bi-directional long short-term memory with gated recurrent unit approach for next word prediction in bodo language}.
\newblock \emph{International Conference on Advanced …}.

\bibitem[{Das et~al.(2024)Das, Baruah, and Roy}]{lowres-2024-A-Das-bodosum}
A~Das, A~Baruah, and S~Roy. 2024.
\newblock \href {https://www.researchsquare.com/article/rs-4841414/latest} {Bodo language text summarization using lstm encoder-decoder}.
\newblock \emph{Research Square}.

\bibitem[{Das and Mohanty(2023)}]{lowres-2023-A-Das-odia}
A~Das and MN~Mohanty. 2023.
\newblock \href {https://www.inderscienceonline.com/doi/abs/10.1504/IJGUC.2023.132619} {Handwritten odia numeral recognition using combined cnn-rnn}.
\newblock \emph{… Journal of Grid and Utility Computing}.

\bibitem[{Das et~al.(2023{\natexlab{c}})Das, Pal, Majumder, Phadikar, and Sekh}]{multimodal-2023-B-Das}
Bidyut Das, Ratnabali Pal, Mukta Majumder, Santanu Phadikar, and Arif~Ahmed Sekh. 2023{\natexlab{c}}.
\newblock A visual attention-based model for bengali image captioning.
\newblock \emph{SN Comput. Sci.}, 4(2).

\bibitem[{Das and Singh(2022)}]{multimodal-2022-R-Das}
Ringki Das and Thoudam~Doren Singh. 2022.
\newblock Assamese news image caption generation using attention mechanism.
\newblock \emph{Multimed. Tools Appl.}, 81(7):10051--10069.

\bibitem[{Das and Singh(2023)}]{Das2023-ej}
Ringki Das and Thoudam~Doren Singh. 2023.
\newblock Image--text multimodal sentiment analysis framework of assamese news articles using late fusion.
\newblock \emph{ACM Trans. Asian Low-resour. Lang. Inf. Process.}, 22(6):1--30.

\bibitem[{Dasari et~al.(2023)Dasari, Chelpuri, and Vuppala}]{lowres-2023-P-Dasari}
P~Dasari, A~Chelpuri, and N~Vuppala. 2023.
\newblock \href {https://aclanthology.org/2023.dravidianlangtech-1.4/} {Transformer-based context aware morphological analyzer for telugu}.
\newblock \emph{Proceedings of the …}.

\bibitem[{Deepajothi et~al.(2024)Deepajothi, Rao, and Ambhika}]{lowres-2024-S-Deepajothi}
S~Deepajothi, VS~Rao, and C~Ambhika. 2024.
\newblock \href {https://search.proquest.com/openview/4d932fd41c363b2d9a1fe1d6fdb5b4ad/1?pq-origsite=gscholar&cbl=4433095} {A comparative study of khasi speech recognition systems with recurrent neural network-based language model}.
\newblock \emph{Journal of Electrical …}.

\bibitem[{Desai(2022)}]{lowres-2022-N-Desai}
N~Desai. 2022.
\newblock \href {https://papers.ssrn.com/sol3/papers.cfm?abstract_id=4382585} {An extractive knowledge-based automatic summarizer system for gujarati text}.
\newblock \emph{International Journal of Innovative Research in Science …}.

\bibitem[{Dey et~al.(2023)Dey, Aishwaryasri, and Mg}]{lowres-2023-A-Dey}
A~Dey, J~Aishwaryasri, and J~Mg. 2023.
\newblock \href {https://ieeexplore.ieee.org/abstract/document/10393243/?casa_token=5wiXX_WGZhIAAAAA:tFWe-DKQeDFSn35OM0JiCl6WpLBvPbSviBDy8CwTgkzj77W9fxdS_sLlhJEwKIEEJrPkfLlk} {Exploring social media trends-a kannada dataset analysis}.
\newblock \emph{… Algorithms and Soft …}.

\bibitem[{Dey and Maringanti(2023)}]{lowres-2023-G-Dey}
G~Dey and HB~Maringanti. 2023.
\newblock \href {https://iopscience.iop.org/article/10.1088/1742-6596/2571/1/012011/meta} {Solmat: A neoteric contextual model for odia language understanding}.
\newblock \emph{Journal of Physics: Conference Series}.

\bibitem[{Dhakal et~al.(2024)Dhakal, Lamichhane, Jha, and Singh}]{lowres-2024-A-Dhakal}
A~Dhakal, A~Lamichhane, AK~Jha, and MP~Singh. 2024.
\newblock \href {https://dspace.bracu.ac.bd/xmlui/handle/10361/24343} {Tiknep: content analysis of nepali tiktok users using natural language processing}.
\newblock \emph{Brac University}.

\bibitem[{DSilva and Sharma(2022)}]{lowres-2022-J-D'Silva}
J~DSilva and U~Sharma. 2022.
\newblock \href {https://link.springer.com/chapter/10.1007/978-981-16-9012-9_4} {Explorations in graph-based ranking algorithms for automatic text summarization on konkani texts}.
\newblock \emph{… Advanced Computing: Select Proceedings of ICSAC …}.

\bibitem[{DSilva and Sharma(2023)}]{lowres-2023-J-D'Silva}
J~DSilva and U~Sharma. 2023.
\newblock \href {https://dl.acm.org/doi/abs/10.1145/3554943?casa_token=sp41b_Wuu2UAAAAA:zrq8IUxzAADgtSNytmDZauClyxKDjFEAaTgGemDFIvC44vLnbuguxDW0emHhY_wfrvlkG5D0_02D} {Impact of similarity measures in graph-based automatic text summarization of konkani texts}.
\newblock \emph{ACM Transactions on Asian and Low-Resource …}.

\bibitem[{Dua et~al.(2024)Dua, Bhagat, Dua, and Chakravarty}]{lowres-2024-M-Dua}
M~Dua, B~Bhagat, S~Dua, and N~Chakravarty. 2024.
\newblock \href {https://link.springer.com/article/10.1007/s10772-024-10087-8} {A review on gujarati language based automatic speech recognition (asr) systems}.
\newblock \emph{International Journal of Speech …}.

\bibitem[{Duggenpudi et~al.(2022)Duggenpudi, Oota, Marreddy, and Mamidi}]{duggenpudi-etal-2022-teluguner}
Suma~Reddy Duggenpudi, Subba~Reddy Oota, Mounika Marreddy, and Radhika Mamidi. 2022.
\newblock \href {https://doi.org/10.18653/v1/2022.acl-srw.20} {{T}elugu{NER}: Leveraging multi-domain named entity recognition with deep transformers}.
\newblock In \emph{Proceedings of the 60th Annual Meeting of the Association for Computational Linguistics: Student Research Workshop}, pages 262--272, Dublin, Ireland. Association for Computational Linguistics.

\bibitem[{Dutta and Choudhury(2022)}]{lowres-2022-D-Dutta}
D~Dutta and RD~Choudhury. 2022.
\newblock \href {https://journals.uob.edu.bh/handle/123456789/4680} {Assamese speech-based vocabulary identification system using convolutional neural network}.
\newblock \emph{International Journal of …}.

\bibitem[{Dutta et~al.(2022)Dutta, Rehman, Mahanta, and Sarmah}]{lowres-2022-K-Dutta}
K~Dutta, R~Rehman, P~Mahanta, and A~Sarmah. 2022.
\newblock \href {https://link.springer.com/chapter/10.1007/978-3-031-20977-2_6} {A study on feature selection for gender detection in speech processing for assamese language}.
\newblock \emph{International Conference on …}.

\bibitem[{EDGE~Project(2024)}]{bangladesh-edge}
Bangladesh Computer~Council EDGE~Project. 2024.
\newblock \href {https://ric.gov.bd/call_for_proposal/how-can-a-generative-bangla-language-model-be-developed/} {How can a generative bangla language model be developed?}

\bibitem[{Elahi et~al.(2023)Elahi, Binte~Rahman, Shahriar, Sarker, Saha~Joy, and Muhammad~Shah}]{Elahi2023-ni}
Kazi~Toufique Elahi, Tasnuva Binte~Rahman, Shakil Shahriar, Samir Sarker, Sajib~Kumar Saha~Joy, and Faisal Muhammad~Shah. 2023.
\newblock Explainable multimodal sentiment analysis on bengali memes.
\newblock In \emph{2023 26th International Conference on Computer and Information Technology ({ICCIT})}, pages 1--6. IEEE.

\bibitem[{Farooqui et~al.(2023)Farooqui, Shaikh, and Rajper}]{lowres-2023-SB-Farooqui}
Saira~Baby Farooqui, Noor~Ahmed Shaikh, and Samina Rajper. 2023.
\newblock \href {https://ojs.ilmauniversity.edu.pk/index.php/jict/article/view/53} {The role of nlp in coreference resolution in sindhi text}.
\newblock \emph{JOURNAL OF INFORMATION AND COMMUNICATION TECHNOLOGY}, 17(2).

\bibitem[{Ganatra and Domadiya(2024)}]{lowres-2024-D-Ganatra}
D~Ganatra and D~Domadiya. 2024.
\newblock \href {https://link.springer.com/chapter/10.1007/978-981-97-4533-3_14} {Toward accurate english to gujarati language translation: An artificial intelligence framework using neural machine translation}.
\newblock \emph{… Deep Learning and Visual Artificial Intelligence}.

\bibitem[{Ghimire et~al.(2024)Ghimire, Bal, and Poudyal}]{lowres-2024-RR-Ghimire}
RR~Ghimire, BK~Bal, and P~Poudyal. 2024.
\newblock \href {https://arxiv.org/abs/2402.03050} {A comprehensive study of the current state-of-the-art in nepali automatic speech recognition systems}.
\newblock \emph{arXiv preprint arXiv:2402.03050}.

\bibitem[{Ghosh et~al.(2023{\natexlab{a}})Ghosh, Senapati, and Pal}]{lowres-2023-K-Ghosh-hate2}
K~Ghosh, A~Senapati, and AS~Pal. 2023{\natexlab{a}}.
\newblock \href {https://ceur-ws.org/Vol-3681/T6-4.pdf} {Annihilate hates (task 4 hasoc 2023): Hate speech detection in assamese bengali and bodo languages.}
\newblock \emph{FIRE (Working Notes)}.

\bibitem[{Ghosh et~al.(2023{\natexlab{b}})Ghosh, Sonowal, and Basumatary}]{lowres-2023-K-Ghosh-hate}
K~Ghosh, D~Sonowal, and A~Basumatary. 2023{\natexlab{b}}.
\newblock \href {https://ieeexplore.ieee.org/abstract/document/10183497/?casa_token=JRi79DlJk-cAAAAA:XHR-zn4KF_aKXKx-E5Ai8SdP-8hu7f_gMhpHmNaC-mRs0LEscXfqXkDfVBEtEtqjOs8CztRt} {Transformer-based hate speech detection in assamese}.
\newblock \emph{2023 IEEE Guwahati …}.

\bibitem[{Ghosh and Caliskan(2023)}]{pronoun-bias}
Sourojit Ghosh and Aylin Caliskan. 2023.
\newblock \href {https://doi.org/10.1145/3600211.3604672} {Chatgpt perpetuates gender bias in machine translation and ignores non-gendered pronouns: Findings across bengali and five other low-resource languages}.
\newblock In \emph{Proceedings of the 2023 AAAI/ACM Conference on AI, Ethics, and Society}, AIES '23, page 901–912, New York, NY, USA. Association for Computing Machinery.

\bibitem[{GL and Badugu(2023)}]{lowres-2023-A-Babu-GL}
A~Babu GL and S~Badugu. 2023.
\newblock \href {https://dl.acm.org/doi/abs/10.1145/3600224?casa_token=oT_nYi6lbkgAAAAA:8zSHtA5bpgs1zzD2HmlMkQ_GTonPT1x-01dg1BQUNXEqXcqb06myWJd1bAlnUOSbdjHDWndzvE0t} {Extractive summarization of telugu text using modified text rank and maximum marginal relevance}.
\newblock \emph{ACM Transactions on Asian and Low-Resource …}.

\bibitem[{Gogoi and Nath(2023)}]{lowres-2023-J-Gogoi}
J~Gogoi and S~Nath. 2023.
\newblock \href {https://ieeexplore.ieee.org/abstract/document/10136105/} {Analysing word stress and its effects on assamese and mizo using machine learning}.
\newblock \emph{2023 2nd International Conference on …}.

\bibitem[{Gokani and Mamidi(2023)}]{lowres-2023-M-Gokani}
M~Gokani and R~Mamidi. 2023.
\newblock \href {https://aclanthology.org/2023.wassa-1.12/} {Gsac: A gujarati sentiment analysis corpus from twitter}.
\newblock \emph{Proceedings of the 13th Workshop on …}.

\bibitem[{Gorla et~al.(2022)Gorla, Tangeda, and Neti}]{lowres-2022-SK-Gorla}
SK~Gorla, SS~Tangeda, and LBM Neti. 2022.
\newblock \href {https://link.springer.com/article/10.1007/s41060-021-00305-w} {Telugu named entity recognition using bert}.
\newblock \emph{International Journal of …}.

\bibitem[{Goutom and Baruah(2023)}]{lowres-2023-PJ-Goutom}
PJ~Goutom and N~Baruah. 2023.
\newblock \href {https://sciresol.s3.us-east-2.amazonaws.com/IJST/Articles/2023/SP-2/FIRHE-2023-5429.pdf} {Text summarization in assamese language using sequence to sequence rnns}.
\newblock \emph{Indian Journal …}.

\bibitem[{Goutom et~al.(2024)Goutom, Baruah, and Sonowal}]{lowres-2024-PJ-Goutom}
PJ~Goutom, N~Baruah, and P~Sonowal. 2024.
\newblock \href {https://www.sciencedirect.com/science/article/pii/S1877050924007804} {Attention-based transformer for assamese abstractive text summarization}.
\newblock \emph{Procedia Computer Science}.

\bibitem[{GrandViewResearch(2024)}]{asia-pacific-ai-market-2024}
GrandViewResearch. 2024.
\newblock {Asia Pacific Artificial Intelligence Market Size, Share and Trends Analysis Report By Solution (Hardware, Software, Services), By Technology (Deep Learning, Machine Learning), By Function, By End-use, By Country, And Segment Forecasts, 2024 - 2030}.
\newblock {https://www.grandviewresearch.com/industry-analysis/asia-pacific-artificial-intelligence-ai-market-report}.

\bibitem[{Grootendorst(2022)}]{grootendorst2022bertopic}
Maarten Grootendorst. 2022.
\newblock Bertopic: Neural topic modeling with a class-based tf-idf procedure.
\newblock \emph{arXiv preprint arXiv:2203.05794}.

\bibitem[{Gupta et~al.(2021)Gupta, Gautam, and Mamidi}]{gupta-etal-2021-vita}
Kshitij Gupta, Devansh Gautam, and Radhika Mamidi. 2021.
\newblock \href {https://doi.org/10.18653/v1/2021.wat-1.19} {{V}i{TA}: Visual-linguistic translation by aligning object tags}.
\newblock In \emph{Proceedings of the 8th Workshop on Asian Translation (WAT2021)}, pages 166--173, Online. Association for Computational Linguistics.

\bibitem[{Gupta and Banerjee(2024)}]{Gupta-Banerjee-2024}
Smita Gupta and Atreyo Banerjee. 2024.
\newblock \href {https://socialinnovationsjournal.com/index.php/sij/article/view/7121} {Ai, justice, and the ecosystem approach – notes from the opennyai mission}.
\newblock \emph{Social Innovations Journal}, 23.

\bibitem[{Gupta et~al.(2022)Gupta, V, and Koolagudi}]{speech-2022-Gupta}
Surya~Prakash Gupta, Spoorthy. V, and Shashidhar~G Koolagudi. 2022.
\newblock \href {https://doi.org/10.1109/GlobConPT57482.2022.9938222} {Recognition of fricative phoneme based hindi words in speech-to-text system using wav2vec2.0 model}.
\newblock In \emph{2022 IEEE Global Conference on Computing, Power and Communication Technologies (GlobConPT)}, pages 1--5.

\bibitem[{Gupta et~al.(2024)Gupta, Narayanan~Venkit, Wilson, and Passonneau}]{gupta-etal-2024-sociodemographic}
Vipul Gupta, Pranav Narayanan~Venkit, Shomir Wilson, and Rebecca Passonneau. 2024.
\newblock \href {https://doi.org/10.18653/v1/2024.gebnlp-1.19} {Sociodemographic bias in language models: A survey and forward path}.
\newblock In \emph{Proceedings of the 5th Workshop on Gender Bias in Natural Language Processing (GeBNLP)}, pages 295--322, Bangkok, Thailand. Association for Computational Linguistics.

\bibitem[{Gurung(2024)}]{lowres-2024-P-Gurung}
P~Gurung. 2024.
\newblock \href {https://opus4.kobv.de/opus4-rhein-waal/frontdoor/index/index/docId/1967} {Sentiment analysis in nepali tweets: Leveraging transformerbased pre-trained models}.
\newblock \emph{Rhine-Waal University of Applied Sciences}.

\bibitem[{Gyanwali et~al.(2024)Gyanwali, Sapkota, and Koirala}]{lowres-2024-A-Gyanwali}
A~Gyanwali, B~Sapkota, and A~Koirala. 2024.
\newblock \href {http://article.publish4promo.com/id/eprint/3561/} {Modular co-attention networks in nepali visual question answering systems}.
\newblock \emph{Asian Journal of …}.

\bibitem[{Hamed et~al.(2023)Hamed, Hussein, Chellah, Chowdhury, Mubarak, Sitaram, Habash, and Ali}]{speech-2023-Hamed}
Injy Hamed, Amir Hussein, Oumnia Chellah, Shammur Chowdhury, Hamdy Mubarak, Sunayana Sitaram, Nizar Habash, and Ahmed Ali. 2023.
\newblock \href {https://doi.org/10.1109/SLT54892.2023.10023181} {Benchmarking evaluation metrics for code-switching automatic speech recognition}.
\newblock In \emph{2022 IEEE Spoken Language Technology Workshop (SLT)}, pages 999--1005.

\bibitem[{Hande et~al.(2022)Hande, Hegde, and Sangeetha}]{lowres-2022-A-Hande}
A~Hande, SU~Hegde, and S~Sangeetha. 2022.
\newblock \href {https://aclanthology.org/2022.ltedi-1.14/} {The best of both worlds: Dual channel language modeling for hope speech detection in low-resourced kannada}.
\newblock \emph{Proceedings of the …}.

\bibitem[{Haq et~al.(2023{\natexlab{a}})Haq, Qiu, Guo, and Tang}]{lowres-2023-I-Haq-toolkit}
I~Haq, W~Qiu, J~Guo, and P~Tang. 2023{\natexlab{a}}.
\newblock \href {https://search.proquest.com/openview/1f21a9e7fb328f3d541acbe9c4d5d983/1?pq-origsite=gscholar&cbl=5444811&casa_token=8i1kmRjcHiQAAAAA:yPYquZgmnFLgcvAIeEbon5v7z1xL8C_yF66w_6uTQz5DA6YSl6CLJvKlhmRO1Ro9PeVMTNIlvw} {Nlpashto: Nlp toolkit for low-resource pashto language}.
\newblock \emph{International Journal of …}.

\bibitem[{Haq et~al.(2023{\natexlab{b}})Haq, Qiu, Guo, and Tang}]{lowres-2023-I-Haq-hate}
I~Haq, W~Qiu, J~Guo, and P~Tang. 2023{\natexlab{b}}.
\newblock \href {https://peerj.com/articles/cs-1617/} {Pashto offensive language detection: a benchmark dataset and monolingual pashto bert}.
\newblock \emph{PeerJ Computer Science}.

\bibitem[{Haque et~al.(2023)Haque, Zaman, Saurav, Haque, Islam, and Amin}]{10103464}
Md.~Zahidul Haque, Sakib Zaman, Jillur~Rahman Saurav, Summit Haque, Md.~Saiful Islam, and Mohammad~Ruhul Amin. 2023.
\newblock \href {https://doi.org/10.1109/ACCESS.2023.3267746} {{B-NER}: A novel bangla named entity recognition dataset with largest entities and its baseline evaluation}.
\newblock \emph{IEEE Access}, 11:45194--45205.

\bibitem[{Harikrishnan and Bindu(2023)}]{lowres-2023-MK-Harikrishnan}
MK~Harikrishnan and KR~Bindu. 2023.
\newblock \href {https://link.springer.com/chapter/10.1007/978-981-97-4650-7_2} {Graphsage-based named entity recognition for malayalam}.
\newblock \emph{… Control, Automation and Artificial Intelligence}.

\bibitem[{Harzing(2010)}]{harzing2013}
Anne-Wil Harzing. 2010.
\newblock \href {https://www.harzing.com/publish-or-perish} {\emph{The Publish or Perish Book: Your Guide to Effective and Responsible Citation Analysis}}.
\newblock Tarma Software Research Pty Ltd.

\bibitem[{Hasan et~al.(2022)Hasan, Jannat, Hossain, Sharif, and Hoque}]{multimodal-2022-Hasan}
Md~Hasan, Nusratul Jannat, Eftekhar Hossain, Omar Sharif, and Mohammed~Moshiul Hoque. 2022.
\newblock \href {https://doi.org/10.18653/v1/2022.dravidianlangtech-1.27} {{CUET}-{NLP}@{D}ravidian{L}ang{T}ech-{ACL}2022: Investigating deep learning techniques to detect multimodal troll memes}.
\newblock In \emph{Proceedings of the Second Workshop on Speech and Language Technologies for Dravidian Languages}, pages 170--176, Dublin, Ireland. Association for Computational Linguistics.

\bibitem[{Hasan et~al.(2024{\natexlab{a}})Hasan, Das, Anjum, Alam, Anjum, Sarker, and Noori}]{hasan-zero}
Md.~Arid Hasan, Shudipta Das, Afiyat Anjum, Firoj Alam, Anika Anjum, Avijit Sarker, and Sheak Rashed~Haider Noori. 2024{\natexlab{a}}.
\newblock \href {https://aclanthology.org/2024.lrec-main.1549} {Zero- and few-shot prompting with {LLM}s: A comparative study with fine-tuned models for {B}angla sentiment analysis}.
\newblock In \emph{Proceedings of the 2024 Joint International Conference on Computational Linguistics, Language Resources and Evaluation (LREC-COLING 2024)}, pages 17808--17818, Torino, Italia. ELRA and ICCL.

\bibitem[{Hasan et~al.(2024{\natexlab{b}})Hasan, Tarannum, Dey, Razzak, and Naseem}]{hasan2024large}
Md~Arid Hasan, Prerona Tarannum, Krishno Dey, Imran Razzak, and Usman Naseem. 2024{\natexlab{b}}.
\newblock Do large language models speak all languages equally? a comparative study in low-resource settings.
\newblock \emph{arXiv preprint arXiv:2408.02237}.

\bibitem[{Hassan et~al.(2024)Hassan, Hussain, Maab, Habib, Khan, and Masood}]{llm-2024-Hassan}
Muhammad~Ehtisham Hassan, Masroor Hussain, Iffat Maab, Usman Habib, Muhammad~Attique Khan, and Anum Masood. 2024.
\newblock \href {https://doi.org/10.1109/ACCESS.2024.3393856} {Detection of sarcasm in urdu tweets using deep learning and transformer based hybrid approaches}.
\newblock \emph{IEEE Access}, 12:61542--61555.

\bibitem[{Hebbi and Mamatha(2023)}]{lowres-2023-C-Hebbi}
C~Hebbi and HR~Mamatha. 2023.
\newblock \href {http://ojs.bonviewpress.com/index.php/AIA/article/view/624} {Comprehensive dataset building and recognition of isolated handwritten kannada characters using machine learning models}.
\newblock \emph{Artificial Intelligence and …}.

\bibitem[{Hema(2022)}]{lowres-2022-A-Hema}
A~Hema. 2022.
\newblock \href {https://web2py.iiit.ac.in/research_centres/publications/download/mastersthesis.pdf.abe6b0eaede6b3f1.48656d61416c6154686573697346696e616c5375626d697373696f6e2e706466.pdf} {Towards domain adaptation for hindi-telugu machine translation}.
\newblock \emph{nan}.

\bibitem[{Hijam(2024)}]{lowres-2024-D-Hijam}
D~Hijam. 2024.
\newblock \href {http://agnee.tezu.ernet.in:8082/jspui/bitstream/1994/1707/3/03_content.pdf} {Convolutional neural network assisted off-line handwritten character recognition of meitei mayek}.
\newblock \emph{nan}.

\bibitem[{Hoque and Salma(2023)}]{Hoque-Salma-2023}
Md.~Nesarul Hoque and Umme Salma. 2023.
\newblock \href {https://doi.org/10.38032/jea.2023.02.003} {Detecting level of depression from social media posts for the low-resource bengali language}.
\newblock \emph{Journal of Engineering Advancements}, 4(02):49–56.

\bibitem[{Hoque and Seddiqui(2023)}]{llm-2023-Hoque}
Md.~Nesarul Hoque and Md.~Hanif Seddiqui. 2023.
\newblock \href {https://doi.org/10.1109/ICCIT60459.2023.10441412} {Leveraging transformer models in the cyberbullying text classification system for the low-resource bengali language}.
\newblock In \emph{2023 26th International Conference on Computer and Information Technology (ICCIT)}, pages 1--6.

\bibitem[{Hossain(2024)}]{hossain-typing-2024}
Anushah Hossain. 2024.
\newblock \href {https://doi.org/10.1109/MAHC.2024.3351948} {{ Text Standards for the “Rest of World”: The Making of the Unicode Standard and the OpenType Format }}.
\newblock \emph{IEEE Annals of the History of Computing}, 46(01):20--33.

\bibitem[{Htun et~al.(2024)Htun, Thu, and Chanlekha}]{lowres-2024-HM-Htun}
HM~Htun, YK~Thu, and H~Chanlekha. 2024.
\newblock \href {https://aclanthology.org/2024.lrec-main.1051/} {mymedicon: End-to-end burmese automatic speech recognition for medical conversations}.
\newblock \emph{Proceedings of the …}.

\bibitem[{Hujon et~al.(2024)Hujon, Singh, and Amitab}]{lowres-2024-AV-Hujon}
AV~Hujon, TD~Singh, and K~Amitab. 2024.
\newblock \href {https://www.sciencedirect.com/science/article/pii/S0957417423023151?casa_token=SgyuMDM286AAAAAA:Ae42LLNhywHMdD5OH54U57rI8oWqBkCucoF359hD0PjT8uOOjvh-arczmpRdE1F-8VxHju3c6g} {Neural machine translation systems for english to khasi: A case study of an austroasiatic language}.
\newblock \emph{Expert Systems with Applications}.

\bibitem[{Hussain et~al.(2022)Hussain, Ahmad, Muhammad, and Ullah}]{lowres-2022-I-Hussain}
I~Hussain, R~Ahmad, S~Muhammad, and K~Ullah. 2022.
\newblock \href {https://ieeexplore.ieee.org/abstract/document/9928191/} {Phti: Pashto handwritten text imagebase for deep learning applications}.
\newblock \emph{IEEE …}.

\bibitem[{Hussein et~al.(2024)Hussein, Zeinali, and Klejch}]{speech-2024-Hussein}
A~Hussein, D~Zeinali, and O~Klejch. 2024.
\newblock \href {https://ieeexplore.ieee.org/abstract/document/10446857/} {Speech collage: code-switched audio generation by collaging monolingual corpora}.
\newblock \emph{ICASSP 2024-2024 …}.

\bibitem[{Huzaifah et~al.(2024)Huzaifah, Zheng, Chanpaisit, and Wu}]{huzaifah-etal-2024-evaluating}
Muhammad Huzaifah, Weihua Zheng, Nattapol Chanpaisit, and Kui Wu. 2024.
\newblock \href {https://aclanthology.org/2024.lrec-main.565} {Evaluating code-switching translation with large language models}.
\newblock In \emph{Proceedings of the 2024 Joint International Conference on Computational Linguistics, Language Resources and Evaluation (LREC-COLING 2024)}, pages 6381--6394, Torino, Italia. ELRA and ICCL.

\bibitem[{Hyoju and Joshi(2023)}]{hyoju2023nepali}
Chanda Hyoju and Basanta~Raj Joshi. 2023.
\newblock Nepali image captioning using end to end transformer.
\newblock In \emph{Proceedings of 14th IOE Graduate Conference}, volume~14.

\bibitem[{{IISc}(2024)}]{vaani}
{IISc}. 2024.
\newblock Project {Vaani}.
\newblock \url{https://vaani.iisc.ac.in/#Data}.

\bibitem[{Inc.(2024)}]{naamche2024}
Naamche Inc. 2024.
\newblock {About Naamche}.
\newblock \url{https://www.naamche.com/about/}.

\bibitem[{{Information and Communication Technology Agency of Sri Lanka}(2024)}]{icta-local-languages}
{Information and Communication Technology Agency of Sri Lanka}. 2024.
\newblock \href {https://www.icta.lk/local-languages-initiative} {{Local Languages Initiative}}.
\newblock Accessed: 2024-10-29.

\bibitem[{International(2022)}]{minorityrights2022southasia}
Minority Rights~Group International. 2022.
\newblock \href {https://minorityrights.org/resources/south-asia-state-of-minorities-report-2021-hate-speech-against-minorities/} {South asia state of minorities report 2021 – hate speech against minorities}.
\newblock Accessed: 2024-11-07.

\bibitem[{J et~al.(2024)J, Dabre, M, Gala, Jayakumar, Puduppully, and Kunchukuttan}]{j-etal-2024-romansetu}
Jaavid J, Raj Dabre, Aswanth M, Jay Gala, Thanmay Jayakumar, Ratish Puduppully, and Anoop Kunchukuttan. 2024.
\newblock \href {https://doi.org/10.18653/v1/2024.acl-long.833} {{R}oman{S}etu: Efficiently unlocking multilingual capabilities of large language models via {R}omanization}.
\newblock In \emph{Proceedings of the 62nd Annual Meeting of the Association for Computational Linguistics (Volume 1: Long Papers)}, pages 15593--15615, Bangkok, Thailand. Association for Computational Linguistics.

\bibitem[{Jaswanth et~al.(2022)Jaswanth, Narayana, and Rahul}]{lowres-2022-M-Jaswanth}
M~Jaswanth, NKL Narayana, and S~Rahul. 2022.
\newblock \href {https://ieeexplore.ieee.org/abstract/document/10051465/?casa_token=Me68YEaO7XYAAAAA:ud3pQQyveGtHlcljOh8rWJ-QkQPR6dpHi9GCHpsnBZQRTQexcqXy-HA7K-XGBplySkxtj1ZD} {A comparative study of feature modelling methods for telugu language identification}.
\newblock \emph{… for Multi-Disciplinary …}.

\bibitem[{Javed et~al.(2023)Javed, Bhogale, Raman, Kumar, Kunchukuttan, and Khapra}]{speech-2023-Javed}
Tahir Javed, Kaushal Bhogale, Abhigyan Raman, Pratyush Kumar, Anoop Kunchukuttan, and Mitesh~M. Khapra. 2023.
\newblock \href {https://doi.org/10.1609/aaai.v37i11.26521} {Indicsuperb: A speech processing universal performance benchmark for indian languages}.
\newblock \emph{Proceedings of the AAAI Conference on Artificial Intelligence}, 37(11):12942--12950.

\bibitem[{Javed et~al.(2022)Javed, Doddapaneni, Raman, Bhogale, Ramesh, Kunchukuttan, Kumar, and Khapra}]{speech-2022-Javed}
Tahir Javed, Sumanth Doddapaneni, Abhigyan Raman, Kaushal~Santosh Bhogale, Gowtham Ramesh, Anoop Kunchukuttan, Pratyush Kumar, and Mitesh~M. Khapra. 2022.
\newblock \href {https://doi.org/10.1609/aaai.v36i10.21327} {Towards building asr systems for the next billion users}.
\newblock \emph{Proceedings of the AAAI Conference on Artificial Intelligence}, 36(10):10813--10821.

\bibitem[{Jha(2024)}]{lowres-2024-B-Jha}
B~Jha. 2024.
\newblock \href {https://www.preprints.org/manuscript/202409.1229} {Exploring nlp challenges and opportunities forlanguages with extensive character sets: A casestudy on nepali}.
\newblock \emph{Preprints}.

\bibitem[{Joshi et~al.(2023)Joshi, Bhatta, and Maharjan}]{lowres-2023-B-Joshi-end2end}
B~Joshi, B~Bhatta, and RK~Maharjan. 2023.
\newblock \href {https://nepjol.info/index.php/JIE/article/view/17-01-13} {End to end based nepali speech recognition system}.
\newblock \emph{Journal of the Institute of Engineering}.

\bibitem[{Joshi et~al.(2022)Joshi, Bhatta, Panday, and Maharjan}]{lowres-2022-B-Joshi-novel}
B~Joshi, B~Bhatta, SP~Panday, and RK~Maharjan. 2022.
\newblock \href {https://link.springer.com/chapter/10.1007/978-981-19-1677-9_39} {A novel deep learning based nepali speech recognition}.
\newblock \emph{International Conference on …}.

\bibitem[{Joshi and Shrestha(2023)}]{lowres-2023-B-Joshi-asr-attention}
B~Joshi and R~Shrestha. 2023.
\newblock \href {http://www.ijicic.org/ijicic-190606.pdf} {Nepali speech recognition using self-attention networks}.
\newblock \emph{… Journal of Innovative Computing, Information and …}.

\bibitem[{Joshi and Arolkar(2022)}]{lowres-2022-K-Joshi}
K~Joshi and H~Arolkar. 2022.
\newblock \href {https://link.springer.com/chapter/10.1007/978-981-97-0744-7_15} {Working of the tesseract ocr on different fonts of gujarati language}.
\newblock \emph{… on Information and Communication Technology for …}.

\bibitem[{Joshi et~al.(2020)Joshi, Santy, Budhiraja, Bali, and Choudhury}]{joshi2020state}
Pratik Joshi, Sowmya Santy, Amar Budhiraja, Kalika Bali, and Monojit Choudhury. 2020.
\newblock The state and fate of linguistic diversity and inclusion in the nlp world.
\newblock \emph{arXiv preprint arXiv:2004.09095}.

\bibitem[{Joshi(2022)}]{joshi:2022:WILDRE6}
Raviraj Joshi. 2022.
\newblock L3cube-mahacorpus and mahabert: Marathi monolingual corpus, marathi bert language models, and resources.
\newblock In \emph{Proceedings of The WILDRE-6 Workshop within the 13th Language Resources and Evaluation Conference}, pages 97--101, Marseille, France. European Language Resources Association.

\bibitem[{K et~al.(2023)K, Raj, and PC}]{lowres-2023-R-Rahmath-K}
R~Rahmath K, PC~Reghu Raj, and R~PC. 2023.
\newblock \href {https://www.tandfonline.com/doi/abs/10.1080/03772063.2022.2077846} {Malayalam question answering system using deep learning approaches}.
\newblock \emph{IETE Journal of Research}.

\bibitem[{Kalita et~al.(2023)Kalita, Boruah, and Kashyap}]{lowres-2023-S-Kalita}
S~Kalita, PA~Boruah, and K~Kashyap. 2023.
\newblock \href {https://ieeexplore.ieee.org/abstract/document/10127231/?casa_token=mL_M2NHPJxUAAAAA:3jdSoBtKz17gYz73tiBrfzflzul3-0ZqapLhNy2ndYjBiyKDPh-jaE9liJF6EllFp6OjC6ij} {Nmt for a low resource language bodo: Preprocessing and resource modelling}.
\newblock \emph{2023 4th International …}.

\bibitem[{Kamath et~al.(2023)Kamath, DN, and Pai}]{lowres-2023-BS-Kamath}
BS~Kamath, CK~DN, and AM~Pai. 2023.
\newblock \href {https://ieeexplore.ieee.org/abstract/document/10367083/?casa_token=fOfwV-eUgfkAAAAA:meQSV6uwg4eL-FLE2icMlmywQPphCTQN_ZZPgrR12ML9ND-gZYvb1R7TXqv9BzRoelSBQ7PL} {English to konkani translator using hindi as a pivot language}.
\newblock \emph{… on Recent Advances …}.

\bibitem[{Kannan et~al.(2023)Kannan, Ravikiran, and Rajalakshmi}]{Kannan2023-py}
R~Ramesh Kannan, Manikandan Ravikiran, and Ratnavel Rajalakshmi. 2023.
\newblock {MMOD-MEME}: A dataset for multimodal face emotion recognition on code-mixed tamil memes.
\newblock In \emph{Communications in Computer and Information Science}, Communications in computer and information science, pages 335--345. Springer International Publishing, Cham.

\bibitem[{Kapoor et~al.(2022)Kapoor, Dhawan, Goel, T~H, Bhatnagar, Agrawal, Agrawal, Bhattacharya, Kumaraguru, and Modi}]{kapoor-etal-2022-hldc}
Arnav Kapoor, Mudit Dhawan, Anmol Goel, Arjun T~H, Akshala Bhatnagar, Vibhu Agrawal, Amul Agrawal, Arnab Bhattacharya, Ponnurangam Kumaraguru, and Ashutosh Modi. 2022.
\newblock \href {https://doi.org/10.18653/v1/2022.findings-acl.278} {{HLDC}: {H}indi legal documents corpus}.
\newblock In \emph{Findings of the Association for Computational Linguistics: ACL 2022}, pages 3521--3536, Dublin, Ireland. Association for Computational Linguistics.

\bibitem[{Kapri(2022)}]{lowres-2022-U-Kapri}
U~Kapri. 2022.
\newblock \href {https://journals.e-palli.com/home/index.php/ajirb/article/view/1021} {Ai, the global 4th industrial system and nepali labour future?}
\newblock \emph{American Journal of IR 4.0 and Beyond}.

\bibitem[{Kashi and Vineeth(2023)}]{lowres-2023-SR-Kashi}
SR~Kashi and HR~Vineeth. 2023.
\newblock \href {https://ieeexplore.ieee.org/abstract/document/10307428/?casa_token=IZF16OiczTkAAAAA:ckMCCQAzLA5WhB5t3_jv-c7uyysv2VxhxUXJx1vDiP8BvYDbFi7Wc8skCWNmcal6houqylt7} {egrumet: Enhanced gated recurrent unit machine for english to kannada lingual translation}.
\newblock \emph{2023 14th International …}.

\bibitem[{Kashyap et~al.(2024)Kashyap, Sarma, and Ahmed}]{lowres-2024-K-Kashyap}
K~Kashyap, SK~Sarma, and MA~Ahmed. 2024.
\newblock \href {https://link.springer.com/article/10.1007/s41870-023-01714-9} {Improving translation between english, assamese bilingual pair with monolingual data, length penalty and model averaging}.
\newblock \emph{International Journal of Information …}.

\bibitem[{Kaur et~al.(2024)Kaur, Bush, and Shi}]{speech-2024-Kaur}
Prabhjot Kaur, L.~Andrew~M. Bush, and Weisong Shi. 2024.
\newblock Direct punjabi to english speech translation using discrete units.
\newblock \emph{International Journal on Cybernetics \& Informatics (IJCI)}, 13(2).

\bibitem[{Kevat and Degadwala(2024)}]{lowres-2024-R-Kevat-pagerank}
R~Kevat and S~Degadwala. 2024.
\newblock \href {https://www.researchgate.net/profile/Sheshang-Degadwala/publication/379413465_Developing_Gujarati_Article_Summarization_Utilizing_Improved_Page-Rank_System/links/660bbd71b839e05a20b6d888/Developing-Gujarati-Article-Summarization-Utilizing-Improved-Page-Rank-System.pdf} {Developing gujarati article summarization utilizing improved page-rank system}.
\newblock \emph{International Journal of Scientific Research in Computer Science Engineering and Information Technology}.

\bibitem[{Kevat and Degadwala(2023)}]{lowres-2023-R-Kevat-review}
Riddhi Kevat and Sheshang Degadwala. 2023.
\newblock A comprehensive review on gujarati-text summarization through different features.
\newblock \emph{International Journal of Scientific Research in Computer Science, Engineering and Information Technology}, pages 301--306.

\bibitem[{Khadka et~al.(2023)Khadka, Ranju, Paudel, and Shah}]{lowres-2023-S-Khadka}
S~Khadka, GC~Ranju, P~Paudel, and R~Shah. 2023.
\newblock \href {https://www.isca-archive.org/sigul_2023/khadka23_sigul.pdf} {Nepali text-to-speech synthesis using tacotron2 for melspectrogram generation}.
\newblock \emph{Proc. 2nd Annual …}.

\bibitem[{Khaliq et~al.(2023)Khaliq, Shabir, Khan, Ahmad, and Usman}]{lowres-2023-F-Khaliq}
F~Khaliq, M~Shabir, I~Khan, S~Ahmad, and M~Usman. 2023.
\newblock \href {https://www.mdpi.com/1424-8220/23/13/6060} {Pashto handwritten invariant character trajectory prediction using a customized deep learning technique}.
\newblock \emph{Sensors}.

\bibitem[{Khan et~al.(2023)Khan, Kamal, Chowdhury, Ahmed, Laskar, and Ahmed}]{khan-etal-2023-banglachq}
Alvi Khan, Fida Kamal, Mohammad~Abrar Chowdhury, Tasnim Ahmed, Md~Tahmid~Rahman Laskar, and Sabbir Ahmed. 2023.
\newblock \href {https://doi.org/10.18653/v1/2023.banglalp-1.10} {{B}angla{CHQ}-summ: An abstractive summarization dataset for medical queries in {B}angla conversational speech}.
\newblock In \emph{Proceedings of the First Workshop on Bangla Language Processing (BLP-2023)}, pages 85--93, Singapore. Association for Computational Linguistics.

\bibitem[{Khandelwal et~al.(2024)Khandelwal, Tonneau, Bean, Kirk, and Hale}]{indian-bhed}
Khyati Khandelwal, Manuel Tonneau, Andrew~M. Bean, Hannah~Rose Kirk, and Scott~A. Hale. 2024.
\newblock \href {https://doi.org/10.1145/3677525.3678666} {Indian-bhed: A dataset for measuring india-centric biases in large language models}.
\newblock In \emph{Proceedings of the 2024 International Conference on Information Technology for Social Good}, GoodIT '24, page 231–239, New York, NY, USA. Association for Computing Machinery.

\bibitem[{Khanuja et~al.(2023)Khanuja, Ruder, and Talukdar}]{khanuja-etal-2023-evaluating}
Simran Khanuja, Sebastian Ruder, and Partha Talukdar. 2023.
\newblock \href {https://doi.org/10.18653/v1/2023.findings-eacl.131} {Evaluating the diversity, equity, and inclusion of {NLP} technology: A case study for {I}ndian languages}.
\newblock In \emph{Findings of the Association for Computational Linguistics: EACL 2023}, pages 1763--1777, Dubrovnik, Croatia. Association for Computational Linguistics.

\bibitem[{Khenglawt et~al.(2022)Khenglawt, Laskar, Manna, Pakray, and Khan}]{miso-visual-genome}
Vanlalmuansangi Khenglawt, Sahinur~Rahman Laskar, Riyanka Manna, Partha Pakray, and Ajoy~Kumar Khan. 2022.
\newblock \href {https://doi.org/10.1109/SILCON55242.2022.10028882} {Mizo visual genome 1.0 : A dataset for english-mizo multimodal neural machine translation}.
\newblock In \emph{2022 IEEE Silchar Subsection Conference (SILCON)}, pages 1--6.

\bibitem[{Kim(2024)}]{ncit-maldives}
Eunsong Kim. 2024.
\newblock \href {https://www.unesco.org/en/articles/maldives-takes-first-step-towards-ai-readiness-unesco-led-steering-committee-meeting} {Maldives takes first step towards ai readiness with unesco-led steering committee meeting}.
\newblock \emph{UNESCO website}.

\bibitem[{Kishore and Shaik(2024)}]{lowres-2024-KS-Kishore}
KS~Kishore and R~Shaik. 2024.
\newblock \href {https://arxiv.org/abs/2404.19369} {Evaluating telugu proficiency in large language models- a comparative analysis of chatgpt and gemini}.
\newblock \emph{arXiv preprint arXiv:2404.19369}.

\bibitem[{Kohli et~al.(2023)Kohli, Parida, Sekhar, Saha, and Nair}]{lowres-2023-GS-Kohli}
GS~Kohli, S~Parida, S~Sekhar, S~Saha, and NB~Nair. 2023.
\newblock \href {https://dl.acm.org/doi/abs/10.1145/3639856.3639890?casa_token=JtqGH6EtI-wAAAAA:MUc4JTx9hdJBNlVkuLGa1RdzEfJVFimEqZZ0Qa1wP79GkmfNRNqEAb8ebtagqlLSm5xzs6nIDRMr} {Building a llama2-finetuned llm for odia language utilizing domain knowledge instruction set}.
\newblock \emph{… Conference on AI-ML …}.

\bibitem[{Kolar and Kumar(2023)}]{lowres-2023-S-Kolar}
S~Kolar and R~Kumar. 2023.
\newblock \href {https://arxiv.org/abs/2307.15376} {Multilingual tourist assistance using chatgpt: Comparing capabilities in hindi, telugu, and kannada}.
\newblock \emph{arXiv preprint arXiv:2307.15376}.

\bibitem[{Koppula et~al.(2022)Koppula, Kumar, and Rao}]{lowres-2022-N-Koppula}
N~Koppula, J~Pradeep Kumar, and K~Srinivas Rao. 2022.
\newblock \href {https://link.springer.com/chapter/10.1007/978-981-16-4435-1_23} {Word sense disambiguation system for information retrieval in telugu language}.
\newblock \emph{… Techniques for IoT …}.

\bibitem[{Kothadiya et~al.(2023)Kothadiya, Bhatt, and Chaudhari}]{lowres-2023-DR-Kothadiya}
DR~Kothadiya, C~Bhatt, and A~Chaudhari. 2023.
\newblock \href {https://link.springer.com/chapter/10.1007/978-981-99-9524-0_8} {Gujformer: A vision transformer-based architecture for gujarati handwritten character recognition}.
\newblock \emph{… Conference on Advances …}.

\bibitem[{Krishna et~al.(2017)Krishna, Zhu, Groth, Johnson, Hata, Kravitz, Chen, Kalantidis, Li, Shamma, Bernstein, and Fei-Fei}]{Krishna2017-qy}
Ranjay Krishna, Yuke Zhu, Oliver Groth, Justin Johnson, Kenji Hata, Joshua Kravitz, Stephanie Chen, Yannis Kalantidis, Li-Jia Li, David~A Shamma, Michael~S Bernstein, and Li~Fei-Fei. 2017.
\newblock Visual genome: Connecting language and vision using crowdsourced dense image annotations.
\newblock \emph{Int. J. Comput. Vis.}, 123(1):32--73.

\bibitem[{Krishnan and Anastasopoulos(2022)}]{lowres-2022-J-Krishnan}
J~Krishnan and A~Anastasopoulos. 2022.
\newblock \href {https://ieeexplore.ieee.org/abstract/document/10021079/?casa_token=H0y0EniEWb8AAAAA:ESdxiPUqGZoti74IDOq_VQ2PYg33bmvYfqErJnbzvF-_w2MLWK8JzSoUab9VnXu6yeIo28PQ} {Cross-lingual text classification of transliterated hindi and malayalam}.
\newblock \emph{… Conference on Big …}.

\bibitem[{Kukana et~al.(2024)Kukana, Sharma, and Bhardwaj}]{lowres-2024-P-Kukana}
P~Kukana, P~Sharma, and N~Bhardwaj. 2024.
\newblock \href {https://link.springer.com/article/10.1007/s41870-024-02189-y} {Optimized featured swarm convolutional neural network (ofscnn) model based dialect recognition system for bagri rajasthani language}.
\newblock \emph{International Journal of Information …}.

\bibitem[{Kulkarni et~al.(2023)Kulkarni, Kulkarni, and Couceiro}]{speech-2023-Kulkarni}
A~Kulkarni, A~Kulkarni, and M~Couceiro. 2023.
\newblock \href {https://arxiv.org/abs/2310.07423} {Adapting the adapters for code-switching in multilingual asr}.
\newblock \emph{arXiv preprint}.

\bibitem[{Kulkarni et~al.(2022)Kulkarni, Joshi, Kamble, and Apte}]{speech-2022-Kulkarni}
Rushikesh Kulkarni, Aditi Joshi, Milind Kamble, and Shaila Apte. 2022.
\newblock Spoken language identification for native indian languages using deep learning techniques.
\newblock In \emph{Machine Learning and Autonomous Systems}, pages 75--97, Singapore. Springer Nature Singapore.

\bibitem[{Kumar et~al.(2024)Kumar, Pathak, and Jaiswal}]{lowres-2024-A-Kumar-safety}
A~Kumar, A~Pathak, and S~Jaiswal. 2024.
\newblock \href {https://africanjournalofbiomedicalresearch.com/index.php/AJBR/article/view/2993} {Artificial intelligence (ai) for women's and children's safety at world heritage sites: Comparative approaches in indian and nepali cities}.
\newblock \emph{African …}.

\bibitem[{Kumar et~al.(2023)Kumar, Rao, and Senapati}]{lowres-2023-CA-Kumar}
CA~Kumar, YC~Rao, and RK~Senapati. 2023.
\newblock \href {https://ieeexplore.ieee.org/abstract/document/10212348/?casa_token=phHNvkDtbkgAAAAA:i-2fliGUVvRUWuanEBKpN-H6Zd-WCh39kZBjdc--hBQj3ZJG8K9Igv44YhjNAzPcVARnXiu1} {Building a light weight intelligible text-to-speech voice model for indian accent telugu}.
\newblock \emph{2023 2nd International …}.

\bibitem[{Kumar et~al.(2022{\natexlab{a}})Kumar, Singh, Ratan, Raj, and Sinha}]{lowres-2022-R-Kumar}
R~Kumar, S~Singh, S~Ratan, M~Raj, and S~Sinha. 2022{\natexlab{a}}.
\newblock \href {https://arxiv.org/abs/2206.12931} {Annotated speech corpus for low resource indian languages: Awadhi, bhojpuri, braj and magahi}.
\newblock \emph{arXiv preprint arXiv …}.

\bibitem[{Kumar et~al.(2022{\natexlab{b}})Kumar, Adiga, Ranjan, Krishna, Ramakrishnan, Goyal, and Jyothi}]{speech-2022-Kumar}
Rishabh Kumar, Devaraja Adiga, Rishav Ranjan, Amrith Krishna, Ganesh Ramakrishnan, Pawan Goyal, and Preethi Jyothi. 2022{\natexlab{b}}.
\newblock Linguistically informed post-processing for {ASR} error correction in sanskrit.
\newblock In \emph{Interspeech 2022}, pages 2293--2297, ISCA. ISCA.

\bibitem[{Kumar(2024)}]{lowres-2024-VR-KUMAR}
VR~Kumar. 2024.
\newblock \href {https://cdn.iiit.ac.in/cdn/web2py.iiit.ac.in/research_centres/publications/download/mastersthesis.pdf.b2f4944e8d751092.323031393730313032375f5468657369732e706466.pdf} {Towards building question answering resources for telugu}.
\newblock \emph{IIIT Hyderabad}.

\bibitem[{Lahkar and Gogoi(2023)}]{lowres-2023-MP-Lahkar}
MP~Lahkar and A~Gogoi. 2023.
\newblock \href {https://ieeexplore.ieee.org/abstract/document/10084269/} {Ascul: Annotated dataset and a deep learning based framework for assamese cultural object detection}.
\newblock \emph{… Conference on Intelligent …}.

\bibitem[{Lakshmi and Latha(2022)}]{lowres-2022-A-Lakshmi}
A~Lakshmi and D~Latha. 2022.
\newblock \href {https://ieeexplore.ieee.org/abstract/document/9781921/?casa_token=EBey-CdtXl8AAAAA:O9sI89AtZCtcAgNnaiidospMPy1iDUS8crVXyrgyFS7LfUuzbUcJNYQcmVVMhhjs65ZlUQLz} {Automatic text summarization for telugu language}.
\newblock \emph{… on Recent Trends in Computer Science …}.

\bibitem[{Lalrempuii and Soni(2023{\natexlab{a}})}]{lowres-2023-C-Lalrempuii}
C~Lalrempuii and B~Soni. 2023{\natexlab{a}}.
\newblock \href {https://ieeexplore.ieee.org/abstract/document/10354679/?casa_token=pxMHJzy4jikAAAAA:NpnaCzAwGvufldsrZ87v1itTqqBdj7hm4h8KgFMWqK8_wMj9HB4WhgMRpSEP5KL361j1RKWb} {Investigation of data augmentation techniques for assamese-english language pair machine translation}.
\newblock \emph{… Symposium on Artificial Intelligence and …}.

\bibitem[{Lalrempuii and Soni(2023{\natexlab{b}})}]{lexicalenhancing}
Candy Lalrempuii and Badal Soni. 2023{\natexlab{b}}.
\newblock \href {https://doi.org/10.1145/3609222} {Investigating unsupervised neural machine translation for low-resource language pair english-mizo via lexically enhanced pre-trained language models}.
\newblock \emph{ACM Trans. Asian Low-Resour. Lang. Inf. Process.}, 22(8).

\bibitem[{{Lanka Law}(2024)}]{lankalawchatbot2024}
{Lanka Law}. 2024.
\newblock {Lanka Law Assistant Chatbot}.
\newblock \url{https://lankalaw.net/}.

\bibitem[{Laskar et~al.(2022{\natexlab{a}})Laskar, Dadure, Manna, Pakray, and Bandyopadhyay}]{multimodal-2022-SR-Laskar}
Sahinur~Rahman Laskar, Pankaj Dadure, Riyanka Manna, Partha Pakray, and Sivaji Bandyopadhyay. 2022{\natexlab{a}}.
\newblock \href {https://aclanthology.org/2022.wat-1.14} {{E}nglish to {B}engali multimodal neural machine translation using transliteration-based phrase pairs augmentation}.
\newblock In \emph{Proceedings of the 9th Workshop on Asian Translation}, pages 111--116, Gyeongju, Republic of Korea. International Conference on Computational Linguistics.

\bibitem[{Laskar et~al.(2023{\natexlab{a}})Laskar, Paul, Pakray, and Bandyopadhyay}]{multimodal-2023-SR-Laskar}
Sahinur~Rahman Laskar, Bishwaraj Paul, Partha Pakray, and Sivaji Bandyopadhyay. 2023{\natexlab{a}}.
\newblock \href {https://doi.org/https://doi.org/10.1016/j.procs.2023.01.078} {English-assamese multimodal neural machine translation using transliteration-based phrase augmentation approach}.
\newblock \emph{Procedia Computer Science}, 218:979--988.
\newblock International Conference on Machine Learning and Data Engineering.

\bibitem[{Laskar et~al.(2022{\natexlab{b}})Laskar, Manna, and Pakray}]{lowres-2022-SR-Laskar}
SR~Laskar, R~Manna, and P~Pakray. 2022{\natexlab{b}}.
\newblock \href {https://www.scielo.org.mx/scielo.php?pid=S1405-55462022000401669&script=sci_arttext} {A domain specific parallel corpus and enhanced english-assamese neural machine translation}.
\newblock \emph{Computación y …}.

\bibitem[{Laskar et~al.(2023{\natexlab{b}})Laskar, Paul, and Pakray}]{lowres-2023-SR-Laskar}
SR~Laskar, B~Paul, and P~Pakray. 2023{\natexlab{b}}.
\newblock \href {https://www.sciencedirect.com/science/article/pii/S1877050923000789} {English-assamese multimodal neural machine translation using transliteration-based phrase augmentation approach}.
\newblock \emph{Procedia Computer}.

\bibitem[{Lavanya and Swamy(2024)}]{lowres-2024-CB-Lavanya}
CB~Lavanya and HSN Swamy. 2024.
\newblock \href {https://ieeexplore.ieee.org/abstract/document/10582201/?casa_token=6Rods0dnBWcAAAAA:aIItnHof_tzfwj93buW0ApjWbuqZ0i6VuWFGLXGEvlrBHmO1mVR5wV7l_VCFC9pdE1MrintO} {Data pre-processing framework for kannada vachana sahitya}.
\newblock \emph{… Conference on Advances …}.

\bibitem[{Li et~al.(2022)Li, Ai, and Xu}]{lowres-2022-J-Li}
Jia Li, Bin Ai, and Cora~Lingling Xu. 2022.
\newblock \href {https://doi.org/10.1177/14687968211018881} {Examining burmese students’ multilingual practices and identity positionings at a border high school in china}.
\newblock \emph{Ethnicities}, 22(2):233--252.

\bibitem[{Liebowitz et~al.(2021)Liebowitz, Macdonald, Shivaram, and Vignaraja}]{liebowitz2021digitalization}
Jeremy Liebowitz, Geoffrey Macdonald, Vivek Shivaram, and Sanjendra Vignaraja. 2021.
\newblock \href {https://gjia.georgetown.edu/2021/05/05/the-digitalization-of-hate-speech-in-south-and-southeast-asia-conflict-mitigation-approaches/} {The digitalization of hate speech in south and southeast asia: Conflict mitigation approaches}.
\newblock Accessed: 2024-11-07.

\bibitem[{Lone et~al.(2022)Lone, Giri, and Bashir}]{lowres-2022-NA-Lone}
NA~Lone, KJ~Giri, and R~Bashir. 2022.
\newblock \href {https://doi.org/10.17485/IJST/v15i43.1964} {Natural language processing resources for the kashmiri language}.
\newblock \emph{Indian Journal of Science and Technology}, 15(43):2275--2281.

\bibitem[{Luitel et~al.(2024)Luitel, Bekoju, Sah, and Shakya}]{lowres-2024-N-Luitel}
N~Luitel, N~Bekoju, AK~Sah, and S~Shakya. 2024.
\newblock \href {https://arxiv.org/abs/2404.18071} {Can perplexity predict fine-tuning performance? an investigation of tokenization effects on sequential language models for nepali}.
\newblock \emph{arXiv preprint arXiv:2404.18071}.

\bibitem[{Maddu and Sanapala(2024)}]{lowres-2024-S-Maddu}
S~Maddu and VR~Sanapala. 2024.
\newblock \href {https://dl.acm.org/doi/abs/10.1145/3695766} {A survey on nlp tasks, resources and techniques for low-resource telugu-english code-mixed text}.
\newblock \emph{ACM Transactions on Asian and Low …}.

\bibitem[{Madhavaraj and Ganesan(2022)}]{speech-2022-Madhavaraj}
A.~Madhavaraj and Ramakrishnan~Angarai Ganesan. 2022.
\newblock \href {http://arxiv.org/abs/2201.09494} {Data and knowledge-driven approaches for multilingual training to improve the performance of speech recognition systems of indian languages}.

\bibitem[{Majhi and Saha(2024{\natexlab{a}})}]{speech-2024-Majhi}
MK~Majhi and SK~Saha. 2024{\natexlab{a}}.
\newblock \href {https://link.springer.com/article/10.1007/s10772-024-10132-6} {An automatic speech recognition system in odia language using attention mechanism and data augmentation}.
\newblock \emph{International Journal of Speech Technology}.

\bibitem[{Majhi and Saha(2024{\natexlab{b}})}]{lowres-2024-MK-Majhi}
MK~Majhi and SK~Saha. 2024{\natexlab{b}}.
\newblock \href {https://link.springer.com/article/10.1007/s10772-024-10132-6} {An automatic speech recognition system in odia language using attention mechanism and data augmentation}.
\newblock \emph{International Journal of Speech Technology}.

\bibitem[{Majhi et~al.(2024)Majhi, Kaphle, and Dahal}]{lowres-2024-S-Majhi}
S~Majhi, S~Kaphle, and S~Dahal. 2024.
\newblock \href {https://nepjol.info/index.php/kjse/article/view/69277} {Nepali virtual ai assistant}.
\newblock \emph{KEC Journal of Science and Engineering}.

\bibitem[{Malage et~al.(2023{\natexlab{a}})Malage, Ashish, Hukkeri, Kavya, and Jayashree}]{speech-2023-Malage}
Ritu~V. Malage, H.~Ashish, Sanjana Hukkeri, E.~Kavya, and R.~Jayashree. 2023{\natexlab{a}}.
\newblock \href {https://doi.org/10.1109/ICICCS56967.2023.10142578} {Low resource speech-to-speech translation of english videos to kannada with lip-synchronization}.
\newblock In \emph{2023 7th International Conference on Intelligent Computing and Control Systems (ICICCS)}, pages 1680--1687.

\bibitem[{Malage et~al.(2023{\natexlab{b}})Malage, Ashish, and Hukkeri}]{lowres-2023-RV-Malage}
RV~Malage, H~Ashish, and S~Hukkeri. 2023{\natexlab{b}}.
\newblock \href {https://ieeexplore.ieee.org/abstract/document/10142578/?casa_token=GG_pw7djtu8AAAAA:HGut7g1JR9CvgqaVI1VWFyYnYJAXaaA1MmbbRCVdroFJRT59Yd8AcF1SB3NRHYeMcl9PQU_D} {Low resource speech-to-speech translation of english videos to kannada with lip-synchronization}.
\newblock \emph{… and Control Systems …}.

\bibitem[{Malik et~al.(2023)Malik, Ghous, and Perveen}]{lowres-2023-MH-Malik}
MH~Malik, H~Ghous, and S~Perveen. 2023.
\newblock \href {https://sjah.isp.edu.pk/index.php/sjah123/article/view/10} {A hierarchical part of speech tag set for saraiki language}.
\newblock \emph{Southern Journal of Arts and Humanities}.

\bibitem[{Mamatha(2023)}]{lowres-2023-HR-Mamatha}
HR~Mamatha. 2023.
\newblock \href {https://ieeexplore.ieee.org/abstract/document/10201030/?casa_token=T8P4Sxy8GtEAAAAA:xZdGQGFF8HF494nC0arQ4eJSwaCqv5ahDYhqZ2D34VY_rUh9e6bDGEWmgYIrw0QYfUQzsNA2} {An empirical analysis of pos tagging for kannada machine translation}.
\newblock \emph{2023 International Conference on Applied …}.

\bibitem[{Manghat et~al.(2022)Manghat, Manghat, and Schultz}]{speech-2022-Manghat}
S~Manghat, S~Manghat, and T~Schultz. 2022.
\newblock \href {https://www.isca-archive.org/interspeech_2022/manghat22_interspeech.pdf} {Normalization of code-switched text for speech synthesis.}
\newblock \emph{INTERSPEECH}.

\bibitem[{Manohar(2023)}]{lowres-2023-K-Manohar-challenges}
K~Manohar. 2023.
\newblock \href {https://kavyamanohar.com/documents/Kavya-phd-2023.pdf} {Linguistic challenges in malayalam speech recognition: Analysis and solutions}.
\newblock \emph{nan}.

\bibitem[{Manohar et~al.(2023)Manohar, Menon, and Abraham}]{lowres-2023-K-Manohar-multilingual}
K~Manohar, GG~Menon, and A~Abraham. 2023.
\newblock \href {https://ieeexplore.ieee.org/abstract/document/10100598/?casa_token=rbYJ8WJkes8AAAAA:fiVF09TN8-RpDqM-JaZtHsKxp9m6bnhiR59UgiKHaE8Se1WNw3o-Xf6Q59QVyA_9seFVQZWc} {Automatic recognition of continuous malayalam speech using pretrained multilingual transformers}.
\newblock \emph{… , IoT and Security …}.

\bibitem[{Manohar and Rajan(2023)}]{lowres-2023-K-Manohar-subword}
K~Manohar and R~Rajan. 2023.
\newblock \href {https://link.springer.com/article/10.1186/s13636-023-00313-7} {Improving speech recognition systems for the morphologically complex malayalam language using subword tokens for language modeling}.
\newblock \emph{EURASIP Journal on Audio, Speech, and Music …}.

\bibitem[{Maqsood(2023)}]{llm-2023-maqsood}
Zoya Maqsood. 2023.
\newblock \href {https://aclanthology.org/2023.ranlp-stud.9} {Weakly supervised learning for aspect based sentiment analysis of {U}rdu tweets}.
\newblock In \emph{Proceedings of the 8th Student Research Workshop associated with the International Conference Recent Advances in Natural Language Processing}, pages 78--86, Varna, Bulgaria. INCOMA Ltd., Shoumen, Bulgaria.

\bibitem[{Marreddy(2023)}]{lowres-2023-M-Marreddy}
M~Marreddy. 2023.
\newblock \href {https://cdn.iiit.ac.in/cdn/web2py.iiit.ac.in/research_centres/publications/download/phdthesis.pdf.bf03d48c390b54d2.32303137323135322d46696e616c5468657369732e706466.pdf} {Text classification for telugu: Datasets, embeddings and models for downstream nlp tasks}.
\newblock \emph{nan}.

\bibitem[{Medhi and Sarma(2023)}]{lowres-2023-SP-Medhi}
SP~Medhi and SK~Sarma. 2023.
\newblock \href {https://link.springer.com/chapter/10.1007/978-3-031-47224-4_21} {Authorship attribution for assamese language documents: Initial results}.
\newblock \emph{International Conference on Advanced Computing …}.

\bibitem[{MEENA et~al.(2022)MEENA, RAO, and CHITTINENI}]{lowres-2022-B-MEENA}
B~MEENA, KV~RAO, and S~CHITTINENI. 2022.
\newblock \href {http://www.jatit.org/volumes/Vol100No18/3Vol100No18.pdf} {… convolution neural network model using soft computing technique to recognize telugu hand-written character for …}.
\newblock \emph{Journal of Theoretical and Applied …}.

\bibitem[{Meetei et~al.(2023)Meetei, Singh, Singh, Das, Singh, and Bandyopadhyay}]{multimodal-2023-LS-Meetei}
Loitongbam~Sanayai Meetei, Salam~Michael Singh, Alok Singh, Ringki Das, Thoudam~Doren Singh, and Sivaji Bandyopadhyay. 2023.
\newblock \href {https://doi.org/https://doi.org/10.1016/j.procs.2023.01.186} {Hindi to english multimodal machine translation on news dataset in low resource setting}.
\newblock \emph{Procedia Computer Science}, 218:2102--2109.
\newblock International Conference on Machine Learning and Data Engineering.

\bibitem[{Megha et~al.(2024)Megha, Abhinav, and Kumar}]{lowres-2024-S-Megha}
S~Megha, GJ~Abhinav, and G~Kumar. 2024.
\newblock \href {https://ieeexplore.ieee.org/abstract/document/10690289/?casa_token=mzh7O_IvuWUAAAAA:kNNHc1WDU8rueDKgQ9Vanl7dWyjhle-oQ-z2mANZ3g5-Fw9ogQqkIXo7y2cMvfSIo1Viyo9Z} {Kannada legal document summarizer: A survey}.
\newblock \emph{… on Advances in …}.

\bibitem[{Mehta et~al.(2022)Mehta, Bharti, and Doshi}]{lowres-2022-H-Mehta}
H~Mehta, SK~Bharti, and N~Doshi. 2022.
\newblock \href {https://ieeexplore.ieee.org/abstract/document/10051338/?casa_token=_7O-Ra9pc4cAAAAA:jyET88kNKRCG7jUcZVrSeqjUFurKWo2HFwsRx-cynCpXSRg52jJUvybvSJBFtYWfGDlbt7XR} {Automatic text summarization in gujarati language}.
\newblock \emph{2022 IEEE 2nd International …}.

\bibitem[{Mehta and Thaker(2023)}]{lowres-2023-N-Mehta}
Nirav Mehta and Hetal Thaker. 2023.
\newblock Data collection for a machine learning model to suggest gujarati recipes to cardiac patients using gujarati food and fruit with nutritive values.
\newblock In \emph{ICT for Intelligent Systems}, pages 271--281, Singapore. Springer Nature Singapore.

\bibitem[{{MeitY}(2023)}]{meity2023}
{MeitY}. 2023.
\newblock \href {https://www.meity.gov.in/writereaddata/files/Bhashini%20Whitepaper%20ver%2.6.0.pdf} {Bhashini whitepaper ver 6.0}.

\bibitem[{Memon et~al.(2024)Memon, Hina, and Kazi}]{lowres-2024-AA-Memon}
AA~Memon, S~Hina, and AK~Kazi. 2024.
\newblock \href {https://search.informit.org/doi/abs/10.3316/informit.T2024072300013400326304109} {Parts-of-speech tagger for sindhi language using deep neural network architecture}.
\newblock \emph{… Research Journal Of …}.

\bibitem[{Meta(2024)}]{no-language-left-behind}
Meta. 2024.
\newblock {No Language Left Behind}.
\newblock {https://ai.facebook.com/blog/no-language- left-behind-democratizing-access- to-ai-through-translation/}.

\bibitem[{Microsoft(2024)}]{microsoft-indian-languages}
Microsoft. 2024.
\newblock {Microsoft Azure Cognitive Services}.
\newblock \url{https://azure.microsoft.com/en-us/services/cognitive-services/}.

\bibitem[{Mim et~al.(2023)Mim, Oussalah, and Singhal}]{lowres-2023-JK-Mim}
JK~Mim, M~Oussalah, and A~Singhal. 2023.
\newblock \href {https://arxiv.org/abs/2312.10528} {Cross-linguistic offensive language detection: Bert-based analysis of bengali, assamese, and bodo conversational hateful content from social media}.
\newblock \emph{arXiv preprint arXiv:2312.10528}.

\bibitem[{Mirishkar et~al.(2023{\natexlab{a}})Mirishkar, Raju~V, Naroju, Maity, Yalla, and Vuppala}]{speech-2023-Mirishkar}
Ganesh~S. Mirishkar, Vishnu~Vidyadhara Raju~V, Meher~Dinesh Naroju, Sudhamay Maity, Prakash Yalla, and Anil~Kumar Vuppala. 2023{\natexlab{a}}.
\newblock \href {https://doi.org/10.1145/3600228} {Iiith-cstd corpus: Crowdsourced strategies for the collection of a large-scale telugu speech corpus}.
\newblock \emph{ACM Trans. Asian Low-Resour. Lang. Inf. Process.}, 22(7).

\bibitem[{Mirishkar et~al.(2023{\natexlab{b}})Mirishkar, V, Naroju, and Maity}]{lowres-2023-GS-Mirishkar}
GS~Mirishkar, VV~Raju V, MD~Naroju, and S~Maity. 2023{\natexlab{b}}.
\newblock \href {https://dl.acm.org/doi/abs/10.1145/3600228?casa_token=Knvf0zZOWloAAAAA:LacafY1fPgZwlv8AyiOPRp-qseVyfgkdwWMQwDJ1OMT0pSU_wez8bgwgfnkdJBBsY1Af2HVgyhnh} {Iiith-cstd corpus: Crowdsourced strategies for the collection of a large-scale telugu speech corpus}.
\newblock \emph{ACM Transactions on …}.

\bibitem[{Mishra et~al.(2023{\natexlab{a}})Mishra, Dash, Parida, and Prasad}]{lowres-2023-N-Mishra}
N~Mishra, SR~Dash, S~Parida, and RS~Prasad. 2023{\natexlab{a}}.
\newblock \href {https://link.springer.com/chapter/10.1007/978-981-99-6706-3_9} {A systematic review on automatic speech recognition for odia language}.
\newblock \emph{International Conference on …}.

\bibitem[{Mishra et~al.(2022)Mishra, Harshit, Saha, and Bhattacharyya}]{multimodal-2022-SK-Mishra}
Santosh~Kumar Mishra, Harshit, Sriparna Saha, and Pushpak Bhattacharyya. 2022.
\newblock \href {https://doi.org/10.1145/3558391} {An object localization-based dense image captioning framework in hindi}.
\newblock \emph{ACM Trans. Asian Low-Resour. Lang. Inf. Process.}, 22(2).

\bibitem[{Mishra et~al.(2023{\natexlab{b}})Mishra, Sinha, Saha, and Bhattacharyya}]{multimodal-2023-SK-Mishra}
Santosh~Kumar Mishra, Sushant Sinha, Sriparna Saha, and Pushpak Bhattacharyya. 2023{\natexlab{b}}.
\newblock \href {https://doi.org/10.1145/3573891} {Dynamic convolution-based encoder-decoder framework for image captioning in hindi}.
\newblock \emph{ACM Trans. Asian Low-Resour. Lang. Inf. Process.}, 22(4).

\bibitem[{Mohanty et~al.(2022{\natexlab{a}})Mohanty, Sahoo, and Nayak}]{lowres-2022-P-Mohanty-dl}
Prithviraj Mohanty, Jyoti~Prakash Sahoo, and Ajit~Kumar Nayak. 2022{\natexlab{a}}.
\newblock Application of deep learning approach for recognition of voiced odia digits.
\newblock \emph{International Journal of Computational Science and Engineering (IJCSE)}, 25(5).

\bibitem[{Mohanty et~al.(2022{\natexlab{b}})Mohanty, Sahoo, and Nayak}]{lowres-2022-P-Mohanty-cnn}
Prithviraj Mohanty, Jyoti~Prakash Sahoo, and Ajit~Kumar Nayak. 2022{\natexlab{b}}.
\newblock Voiced odia digit recognition using convolutional neural network.
\newblock In \emph{Advances in Distributed Computing and Machine Learning}, pages 161--173, Singapore. Springer Singapore.

\bibitem[{More and DSilva(2023)}]{lowres-2023-C-More}
C~More and J~DSilva. 2023.
\newblock \href {https://dl.acm.org/doi/abs/10.1145/3632754.3632758?casa_token=Pu16eAbHw9QAAAAA:m8Q3BC5EaP0HNA0nl4jhL-G0LhrHCOGyYsWNWOpVD3R3_3_82-I7uSQHvD4yOLgL1rNCtOgVFnMC} {Keyword driven language-independent low-resource graph-based automatic text summarization of konkani texts}.
\newblock \emph{Proceedings of the 15th Annual Meeting of the Forum …}.

\bibitem[{Mundotiya et~al.(2023)Mundotiya, Kumar, Kumar, and Chaudhary}]{lowres-2023-R-Mundotiya}
R~Mundotiya, S~Kumar, A~Kumar, and U~Chaudhary. 2023.
\newblock \href {https://dl.acm.org/doi/abs/10.1145/3533428?casa_token=xJLIqjgCLaUAAAAA:jUbCD7gDPJPIHIRFp12ApyyAqDGlxXivo18sjxl0OjMEuyiZHTEvAMAphOwsFgoqSrV32dN-32Jm} {Development of a dataset and a deep learning baseline named entity recognizer for three low resource languages: Bhojpuri, maithili, and magahi}.
\newblock \emph{ACM Transactions on …}.

\bibitem[{Mundotiya et~al.(2022)Mundotiya, Mishra, and Singh}]{lowres-2022-RK-Mundotiya}
Rajesh~Kumar Mundotiya, Swasti Mishra, and Anil~Kumar Singh. 2022.
\newblock \href {https://doi.org/https://doi.org/10.1016/j.jksuci.2021.09.022} {Hierarchical self attention based sequential labelling model for bhojpuri, maithili and magahi languages}.
\newblock \emph{Journal of King Saud University - Computer and Information Sciences}, 34(10, Part A):8739--8749.

\bibitem[{Murthy et~al.(2022)Murthy, Bhattacharjee, Sharnagat, Khatri, Kanojia, and Bhattacharyya}]{murthy-etal-2022-hiner}
Rudra Murthy, Pallab Bhattacharjee, Rahul Sharnagat, Jyotsana Khatri, Diptesh Kanojia, and Pushpak Bhattacharyya. 2022.
\newblock \href {https://aclanthology.org/2022.lrec-1.475} {{H}i{NER}: A large {H}indi named entity recognition dataset}.
\newblock In \emph{Proceedings of the Thirteenth Language Resources and Evaluation Conference}, pages 4467--4476, Marseille, France. European Language Resources Association.

\bibitem[{Muttaraju et~al.(2022)Muttaraju, Singh, Kabber, and R}]{humor}
Chakita Muttaraju, Aakansha Singh, Anusha Kabber, and Mamatha~H. R. 2022.
\newblock \href {https://aclanthology.org/2022.nsurl-1.2} {Semi-supervised and unsupervised detection of humour in code-mixed {H}indi-{E}nglish tweets}.
\newblock In \emph{Proceedings of the Third International Workshop on NLP Solutions for Under Resourced Languages (NSURL 2022) co-located with ICNLSP 2022}, pages 8--13, Trento, Italy. Association for Computational Linguistics.

\bibitem[{Myint~Oo et~al.(2022)Myint~Oo, Tanprasert, Kyaw~Thu, and Supnithi}]{lowres-2022-TM-Oo}
Thazin Myint~Oo, Thitipong Tanprasert, Ye~Kyaw~Thu, and Thepchai Supnithi. 2022.
\newblock \href {https://doi.org/10.1109/iSAI-NLP56921.2022.9960259} {Syllable-to-syllable and word-to-word transducers for burmese dialect translation}.
\newblock In \emph{2022 17th International Joint Symposium on Artificial Intelligence and Natural Language Processing (iSAI-NLP)}, pages 1--6.

\bibitem[{Naidu and Seshashayee(2024)}]{lowres-2024-GJ-Naidu}
GJ~Naidu and M~Seshashayee. 2024.
\newblock \href {https://doi.org/10.15622/ia.23.1.2} {Sentiment analysis framework for telugu text based on novel contrived passive aggressive with fuzzy weighting classifier (cpsc-fwc)}.
\newblock \emph{Informatics and Automation}, 23(1):39--64.

\bibitem[{Nambiar et~al.(2023)Nambiar, S, and Idicula}]{lowres-2023-S-K.-Nambiar}
S~K. Nambiar, D~Peter S, and S~Mary Idicula. 2023.
\newblock \href {https://dl.acm.org/doi/abs/10.1145/3561819?casa_token=3LM3LPSFUXEAAAAA:L2A0WyWsMBgz1vmEB3IU0NjujyqeFZuBTLoFW1vZkKBexAPRsgbgaE0GvFkUY3Qd5G3eayExzowN} {Abstractive summarization of text document in malayalam language: Enhancing attention model using pos tagging feature}.
\newblock \emph{ACM Transactions on Asian …}.

\bibitem[{Namburu et~al.(2024)Namburu, Soman, and Kumar}]{lowres-2024-SSG-Namburu}
SSG Namburu, KP~Soman, and SS~Kumar. 2024.
\newblock \href {https://ieeexplore.ieee.org/abstract/document/10626622/?casa_token=vBEKUSLYpWUAAAAA:zKVhx4-aF4iRnmUcMxo37mF_IENRM06kqViBCQjcULV5nd9Yx96xo5MwtHM19YmWI4CIo1jc} {Effectiveness of gnn based approach for topic classification of telugu text}.
\newblock \emph{2023 4th International …}.

\bibitem[{Narzary et~al.(2022{\natexlab{a}})Narzary, Brahma, Narzary, and Senapati}]{lowres-2022-M-Narzary}
M~Narzary, M~Brahma, S~Narzary, and A~Senapati. 2022{\natexlab{a}}.
\newblock \href {https://link.springer.com/chapter/10.1007/978-981-99-2609-1_3} {A computational approach for the tonal identification in bodo language}.
\newblock \emph{North-East Research …}.

\bibitem[{Narzary et~al.(2024{\natexlab{a}})Narzary, Brahma, Nandi, and Som}]{lowres-2024-S-Narzary-ner}
S~Narzary, A~Brahma, S~Nandi, and B~Som. 2024{\natexlab{a}}.
\newblock \href {https://www.sciencedirect.com/science/article/pii/S1877050924009049} {Deep learning based named entity recognition for the bodo language}.
\newblock \emph{Procedia Computer Science}.

\bibitem[{Narzary et~al.(2024{\natexlab{b}})Narzary, Brahma, Nandi, and Som}]{lowres-2024-S-Narzary-smt}
S~Narzary, M~Brahma, S~Nandi, and B~Som. 2024{\natexlab{b}}.
\newblock \href {https://link.springer.com/article/10.1007/s11042-024-20277-w} {Bpe and morphologically segmented phrase based statistical machine translation system for indian languages to resource constrained language bodo}.
\newblock \emph{Multimedia Tools and …}.

\bibitem[{Narzary et~al.(2022{\natexlab{b}})Narzary, Brahma, and Narzary}]{lowres-2022-S-Narzary}
S~Narzary, M~Brahma, and M~Narzary. 2022{\natexlab{b}}.
\newblock \href {https://aclanthology.org/2022.lrec-1.705/} {Generating monolingual dataset for low resource language bodo from old books using google keep}.
\newblock \emph{Proceedings of the …}.

\bibitem[{Nasrat(2023)}]{lowres-2023-PN-Nasrat}
PN~Nasrat. 2023.
\newblock \href {https://randwickresearch.com/index.php/rielsj/article/view/646} {Graph metical problems of the pashto language alphabet}.
\newblock \emph{… of Education and Linguistics Science Journal}.

\bibitem[{Nath et~al.(2023{\natexlab{a}})Nath, Mannan, and Krishnaswamy}]{lowres-2023-A-Nath}
A~Nath, S~Mannan, and N~Krishnaswamy. 2023{\natexlab{a}}.
\newblock \href {https://arxiv.org/abs/2305.13641} {Axomiyaberta: A phonologically-aware transformer model for assamese}.
\newblock \emph{arXiv preprint arXiv:2305.13641}.

\bibitem[{Nath et~al.(2022)Nath, Adhikary, Dadure, and Pakray}]{lowres-2022-P-Nath}
P~Nath, PK~Adhikary, P~Dadure, and P~Pakray. 2022.
\newblock \href {https://aclanthology.org/2022.rocling-1.33/} {Image caption generation for low-resource assamese language}.
\newblock \emph{… Speech Processing …}.

\bibitem[{Nath et~al.(2023{\natexlab{b}})Nath, Singh, and Gupta}]{llm-2023-Nath}
T~Nath, VK~Singh, and V~Gupta. 2023{\natexlab{b}}.
\newblock \href {https://www.researchsquare.com/article/rs-2819284/latest} {Bonghope: An annotated corpus for bengali hope speech detection}.
\newblock \emph{nan}.

\bibitem[{Nathani et~al.(2023)Nathani, Arora, Joshi, and Katyayan}]{lowres-2023-B-Nathani}
B~Nathani, P~Arora, N~Joshi, and P~Katyayan. 2023.
\newblock \href {https://link.springer.com/chapter/10.1007/978-3-031-51167-7_4} {Sindhi pos tagger using lstm and pre-trained word embeddings}.
\newblock \emph{… Conference on Data …}.

\bibitem[{Nayak and Das(2022)}]{lowres-2022-M-Nayak}
M~Nayak and N~Das. 2022.
\newblock \href {https://link.springer.com/chapter/10.1007/978-981-16-9873-6_25} {Graph based automatic keyword extraction from odia text document}.
\newblock \emph{Intelligent and Cloud Computing: Proceedings of ICICC …}.

\bibitem[{Nielsen and McConville(2022)}]{nielsen2022mumin}
Dan~S Nielsen and Ryan McConville. 2022.
\newblock Mumin: A large-scale multilingual multimodal fact-checked misinformation social network dataset.
\newblock In \emph{Proceedings of the 45th international ACM SIGIR conference on research and development in information retrieval}, pages 3141--3153.

\bibitem[{Niraula and Chapagain(2023)}]{lowres-2023-N-Niraula}
N~Niraula and J~Chapagain. 2023.
\newblock \href {https://journals.flvc.org/FLAIRS/article/view/133384} {Danfener-named entity recognition in nepali tweets}.
\newblock \emph{The International FLAIRS Conference …}.

\bibitem[{Niyogi and Bhattacharya(2024)}]{niyogi2024}
Mitodru Niyogi and Arnab Bhattacharya. 2024.
\newblock \href {http://arxiv.org/abs/2401.18034} {Paramanu: A family of novel efficient generative foundation language models for indian languages}.

\bibitem[{of~Communications \& Information Technology~(MoCIT)(2024)}]{mocit2024}
Ministry of~Communications \& Information Technology~(MoCIT). 2024.
\newblock \href {https://mocit.gov.np/content/12663/12663-nepal/} {Government of nepal ministry of communications and information technology}.

\bibitem[{Ola(2024)}]{ola-krutrim}
Ola. 2024.
\newblock {Ola Krutrim}.
\newblock \url{https://olacabs.com/krutrim/}.

\bibitem[{Oo et~al.(2023)Oo, Tanprasert, Thu, and Supnithi}]{lowres-2023-TM-Oo}
Thazin~Myint Oo, Thitipong Tanprasert, Ye~Kyaw Thu, and Thepchai Supnithi. 2023.
\newblock \href {https://doi.org/10.1109/ACCESS.2023.3236804} {Transfer and triangulation pivot translation approaches for burmese dialects}.
\newblock \emph{IEEE Access}, 11:6150--6168.

\bibitem[{OpenAI(2024)}]{openai2024gpt4}
OpenAI. 2024.
\newblock \href {https://platform.openai.com/docs/guides/gpt} {Gpt-4}.

\bibitem[{Pahari and Shimada(2023)}]{lowres-2023-N-Pahari}
N~Pahari and K~Shimada. 2023.
\newblock \href {https://aclanthology.org/2023.calcs-1.3/} {Language preference for expression of sentiment for nepali-english bilingual speakers on social media}.
\newblock \emph{Proceedings of the 6th Workshop on …}.

\bibitem[{Panchal and Shah(2024)}]{lowres-2024-BY-Panchal}
BY~Panchal and A~Shah. 2024.
\newblock \href {https://link.springer.com/chapter/10.1007/978-981-97-3191-6_21} {Spell checker using norvig algorithm for gujarati language}.
\newblock \emph{International Conference on Smart Data Intelligence}.

\bibitem[{Panda et~al.(2022)Panda, Dash, Padhy, and Palo}]{lowres-2022-R-Panda}
R~Panda, S~Dash, S~Padhy, and S~Palo. 2022.
\newblock \href {https://ieeexplore.ieee.org/abstract/document/10032934/} {Complex odia handwritten character recognition using deep learning model}.
\newblock \emph{2022 IEEE International …}.

\bibitem[{Pandit and Bhatt(2023)}]{lowres-2023-P-Pandit}
P~Pandit and S~Bhatt. 2023.
\newblock \href {https://www.inderscienceonline.com/doi/abs/10.1504/IJICA.2023.134184} {Automatic speech recognition of gujarati digits using wavelet coefficients in machine learning algorithms}.
\newblock \emph{International Journal of Innovative …}.

\bibitem[{Paneru et~al.(2024)Paneru, Paneru, and Poudyal}]{lowres-2024-B-Paneru}
B~Paneru, B~Paneru, and KN~Poudyal. 2024.
\newblock \href {https://www.sciencedirect.com/science/article/pii/S2772941924000942} {Advancing human-computer interaction: Ai-driven translation of american sign language to nepali using convolutional neural networks and text-to …}.
\newblock \emph{Systems and Soft Computing}.

\bibitem[{Pangsatabam et~al.(2023)Pangsatabam, Chanu, and Singh}]{lowres-2023-H-Pangsatabam}
H~Pangsatabam, YJ~Chanu, and NK~Singh. 2023.
\newblock \href {https://link.springer.com/chapter/10.1007/978-981-99-4713-3_33} {Design and evaluation of speech processing systems for meetei/meitei mayek}.
\newblock \emph{International Conference on …}.

\bibitem[{Pankaj(2023)}]{lowres-2023-DS-Pankaj}
DS~Pankaj. 2023.
\newblock \href {https://ieeexplore.ieee.org/abstract/document/10165363/} {Challenges in creating text summarization models in malayalam: A study}.
\newblock \emph{2023 International Conference on Control …}.

\bibitem[{Parajuli and Joshi(2023)}]{lowres-2023-K-Parajuli}
K~Parajuli and SR~Joshi. 2023.
\newblock \href {https://arxiv.org/abs/2312.07418} {Attention based encoder decoder model for video captioning in nepali (2023)}.
\newblock \emph{arXiv preprint arXiv:2312.07418}.

\bibitem[{Parida et~al.(2023)Parida, Sekhar, Panda, and Jena}]{lowres-2023-S-Parida}
S~Parida, S~Sekhar, S~Panda, and S~Jena. 2023.
\newblock \href {https://ieeexplore.ieee.org/abstract/document/10404195/?casa_token=oHdbDsGL-GMAAAAA:qHjeE_gQ99YlVIGDLx_gUVj_1SlrtBoB5fITipoGtSbTGHmL6PPJWG2fb2hYEhzQf0t7jTn2} {Olive: An instruction following llama model for odia language}.
\newblock \emph{2023 IEEE Silchar …}.

\bibitem[{Parida et~al.(2024)Parida, Panwar, Lata, Mishra, and Sekhar}]{parida2024building}
Shantipriya Parida, Shakshi Panwar, Kusum Lata, Sanskruti Mishra, and Sambit Sekhar. 2024.
\newblock Building pre-train llm dataset for the indic languages: a case study on hindi.
\newblock \emph{arXiv preprint arXiv:2407.09855}.

\bibitem[{Patel and Joshi(2022)}]{lowres-2022-M-Patel-dsetgens}
Margi Patel and Brijendra~Kumar Joshi. 2022.
\newblock \href {https://doi.org/10.1109/CSNT54456.2022.9787677} {Dsetgens: An automated technique for building dataset from speech with respect to gujarati-english}.
\newblock In \emph{2022 IEEE 11th International Conference on Communication Systems and Network Technologies (CSNT)}, pages 314--317.

\bibitem[{Patel and Joshi(2023)}]{lowres-2022-M-Patel-gujagra}
Margi Patel and Brijendra~Kumar Joshi. 2023.
\newblock Gujagra: An acyclic graph to unify semantic knowledge, antonyms, and gujarati--english translation of input text.
\newblock In \emph{Machine Learning and Computational Intelligence Techniques for Data Engineering}, pages 151--160, Singapore. Springer Nature Singapore.

\bibitem[{Pathak et~al.(2022)Pathak, Nandi, and Sarmah}]{lowres-2022-D-Pathak}
D~Pathak, S~Nandi, and P~Sarmah. 2022.
\newblock \href {https://arxiv.org/abs/2207.03422} {Asner--annotated dataset and baseline for assamese named entity recognition}.
\newblock \emph{arXiv preprint arXiv:2207.03422}.

\bibitem[{Pathak et~al.(2023)Pathak, Nandi, and Sarmah}]{lowres-2023-D-Pathak-ensemble}
D~Pathak, S~Nandi, and P~Sarmah. 2023.
\newblock \href {https://dl.acm.org/doi/abs/10.1145/3617653?casa_token=sRvCYc4ZwS0AAAAA:rHw8Qrrkm4b4yPJQkSfbHflj827812DzLP52J3hBx9qLmqTI5npeopgk5GCgDzq1V4U4bpyhUADr} {Part-of-speech tagger for assamese using ensembling approach}.
\newblock \emph{ACM Transactions on Asian and Low …}.

\bibitem[{Pathak et~al.(2024)Pathak, Narzary, Nandi, and Som}]{lowres-2024-D-Pathak-bodo}
D~Pathak, S~Narzary, S~Nandi, and B~Som. 2024.
\newblock \href {https://www.cambridge.org/core/journals/natural-language-processing/article/partofspeech-tagger-for-bodo-language-using-deep-learning-approach/8E223194F0DCB9837640577190631CC3} {Part-of-speech tagger for bodo language using deep learning approach}.
\newblock \emph{Natural Language …}.

\bibitem[{Patil et~al.(2024)Patil, Ikshu, and Patil}]{lowres-2024-HS-Patil}
HS~Patil, KV~Ikshu, and SS~Patil. 2024.
\newblock \href {https://ieeexplore.ieee.org/abstract/document/10561318/?casa_token=EKJHjyk8rVgAAAAA:ZQ3oZDbHrYThsHzp1NTo2fKZfmY2bMmX9IYtbU59qBGbJA_pImmQOx6A_z_8kgCRzepMvnGA} {Translation of different dialects of kannada to english using machine learning}.
\newblock \emph{… on Smart Systems …}.

\bibitem[{Pattnaik and Nayak(2022)}]{lowres-2022-S-Pattnaik}
S~Pattnaik and AK~Nayak. 2022.
\newblock \href {https://link.springer.com/chapter/10.1007/978-981-16-4807-6_22} {Extractive document summarization of text in odia language}.
\newblock \emph{Advances in Distributed Computing and Machine …}.

\bibitem[{Paudel et~al.(2024)Paudel, Khadka, and Shah}]{lowres-2024-P-Paudel}
P~Paudel, S~Khadka, and R~Shah. 2024.
\newblock \href {https://arxiv.org/abs/2407.04577} {Optimizing nepali pdf extraction: A comparative study of parser and ocr technologies}.
\newblock \emph{arXiv preprint arXiv:2407.04577}.

\bibitem[{Phukan et~al.(2024)Phukan, Baruah, Sarma, and Konwar}]{lowres-2024-R-Phukan-pos}
R~Phukan, N~Baruah, SK~Sarma, and D~Konwar. 2024.
\newblock \href {https://www.sciencedirect.com/science/article/pii/S1877050924008147} {Exploring character-level deep learning models for pos tagging in assamese language}.
\newblock \emph{Procedia Computer Science}.

\bibitem[{Pokhrel and Adhikari(2023)}]{lowres-2023-C-Pokhrel}
C~Pokhrel and R~Adhikari. 2023.
\newblock \href {https://www.researchgate.net/profile/Chitran-Pokhrel/publication/372221925_Automatic_Extractive_Text_Summarization_for_Text_in_Nepali_Language_with_Bidirectional_Encoder_Representation_Transformers_and_K-Mean_Clustering/links/64aa7f1dc41fb852dd60d232/Automatic-Extractive-Text-Summarization-for-Text-in-Nepali-Language-with-Bidirectional-Encoder-Representation-Transformers-and-K-Mean-Clustering.pdf} {Automatic extractive text summarization for text in nepali language with bidirectional encoder representation transformers and k-mean clustering}.
\newblock \emph{nan}.

\bibitem[{Poudel et~al.(2023)Poudel, Ghimire, Subedi, and Singh}]{lowres-2023-S-Poudel}
S~Poudel, N~Ghimire, B~Subedi, and S~Singh. 2023.
\newblock \href {https://arxiv.org/abs/2311.06898} {Retrieval and generative approaches for a pregnancy chatbot in nepali with stemmed and non-stemmed data: A comparative study}.
\newblock \emph{arXiv preprint arXiv:2311.06898}.

\bibitem[{Pradhan and Yajnik(2024)}]{lowres-2024-A-Pradhan}
A~Pradhan and A~Yajnik. 2024.
\newblock \href {https://link.springer.com/article/10.1007/s11042-023-15679-1} {Parts-of-speech tagging of nepali texts with bidirectional lstm, conditional random fields and hmm}.
\newblock \emph{Multimedia Tools and Applications}.

\bibitem[{Priyadharshini and Chakravarthi(2023)}]{lowres-2023-R-Priyadharshini}
R~Priyadharshini and BR~Chakravarthi. 2023.
\newblock \href {https://aclanthology.org/2023.dravidianlangtech-1.11/} {Overview of shared-task on abusive comment detection in tamil and telugu}.
\newblock \emph{Proceedings of the …}.

\bibitem[{Priyamvada et~al.(2022)Priyamvada, Govind, Menon, and Premjith}]{lowres-2022-R-Priyamvada}
R~Priyamvada, D~Govind, VK~Menon, and B~Premjith. 2022.
\newblock \href {https://link.springer.com/chapter/10.1007/978-981-16-6624-7_5} {Grapheme to phoneme conversion for malayalam speech using encoder-decoder architecture}.
\newblock \emph{… Data Engineering and …}.

\bibitem[{Pudasaini et~al.(2024)Pudasaini, Ghimire, Ale, and Shakya}]{lowres-2024-S-Pudasaini}
S~Pudasaini, S~Ghimire, P~Ale, and A~Shakya. 2024.
\newblock \href {https://dl.acm.org/doi/abs/10.1145/3647782.3647804} {Application of nepali large language models to improve sentiment analysis}.
\newblock \emph{Proceedings of the …}.

\bibitem[{Pudasaini et~al.(2023)Pudasaini, Shakya, and Tamang}]{lowres-2023-S-Pudasaini}
S~Pudasaini, S~Shakya, and A~Tamang. 2023.
\newblock \href {https://ieeexplore.ieee.org/abstract/document/10290690/?casa_token=3FWqrNu76ZAAAAAA:a8qXRysY6LaPgQ2a9cpqvIE3ACMr42uOwKGk1Qaj_l5-O9oFocuPfTzwT6GSnhROpnjEfEl4} {Nepalibert: Pre-training of masked language model in nepali corpus}.
\newblock \emph{… Conference on I …}.

\bibitem[{Qumar et~al.(2024{\natexlab{a}})Qumar, Azim, and Quadri}]{lowres-2024-SMU-Qumar-emerging}
SMU Qumar, M~Azim, and SMK Quadri. 2024{\natexlab{a}}.
\newblock \href {https://link.springer.com/article/10.1007/s00146-024-01981-5} {Emerging resources, enduring challenges: a comprehensive study of kashmiri parallel corpus}.
\newblock \emph{AI and Society}.

\bibitem[{Qumar et~al.(2024{\natexlab{b}})Qumar, Azim, and Quadri}]{lowres-2024-SMU-Qumar-parallel}
Syed Matla~Ul Qumar, Muzaffar Azim, and S.~M.~K. Quadri. 2024{\natexlab{b}}.
\newblock \href {https://doi.org/10.1007/s41870-024-01979-8} {Addressing the data gap: building a parallel corpus for kashmiri language}.
\newblock \emph{International Journal of Information Technology}, 16:4363--4379.

\bibitem[{Rahmath and Raj(2023)}]{lowres-2023-K-Reji-Rahmath}
K~Reji Rahmath and PC~Reghu Raj. 2023.
\newblock \href {https://link.springer.com/chapter/10.1007/978-981-97-0892-5_54} {Mbabi: A benchmark dataset for malayalam text understanding and reasoning}.
\newblock \emph{… Conference on Computing …}.

\bibitem[{Rahul et~al.(2023)Rahul, Arathi, and Panicker}]{lowres-2023-C-Rahul}
C~Rahul, T~Arathi, and LS~Panicker. 2023.
\newblock \href {https://www.sciencedirect.com/science/article/pii/S1746809423004846} {Morphology and word sense disambiguation embedded multimodal neural machine translation system between sanskrit and malayalam}.
\newblock \emph{… Signal Processing and …}.

\bibitem[{Rai et~al.(2024)Rai, Shiwakoti, Basukala, and Dahal}]{lowres-2024-A-Rai}
A~Rai, S~Shiwakoti, S~Basukala, and SS~Dahal. 2024.
\newblock \href {https://nepjol.info/index.php/kjse/article/view/69276} {Nepali text-to-speech synthesis using tacotron2 and waveglow}.
\newblock \emph{KEC Journal of Science and …}.

\bibitem[{Raj et~al.(2023{\natexlab{a}})Raj, Krishna, and Balaram}]{lowres-2023-GR-Raj-aware}
GR~Raj, BB~Krishna, and P~Balaram. 2023{\natexlab{a}}.
\newblock \href {https://aclanthology.org/2023.icon-1.4/} {Pronunciation-aware syllable tokenizer for nepali automatic speech recognition system}.
\newblock \emph{Proceedings of the 20th …}.

\bibitem[{Raj et~al.(2023{\natexlab{b}})Raj, Krishna, and Prakash}]{lowres-2023-GR-Raj-active}
GR~Raj, BB~Krishna, and P~Prakash. 2023{\natexlab{b}}.
\newblock \href {https://aclanthology.org/2023.icon-1.9/} {Active learning approach for fine-tuning pre-trained asr model for a low-resourced language: A case study of nepali}.
\newblock \emph{Proceedings of the 20th …}.

\bibitem[{Rajan and Salgaonkar(2022)}]{lowres-2022-A-Rajan}
A~Rajan and A~Salgaonkar. 2022.
\newblock \href {https://link.springer.com/chapter/10.1007/978-981-16-0739-4_12} {Survey of nlp resources in low-resource languages nepali, sindhi and konkani}.
\newblock \emph{… for Competitive Strategies (ICTCS 2020) ICT …}.

\bibitem[{Ramesh et~al.(2022)Ramesh, Doddapaneni, Bheemaraj, Jobanputra, AK, Sharma, Sahoo, Diddee, J, Kakwani, Kumar, Pradeep, Nagaraj, Deepak, Raghavan, Kunchukuttan, Kumar, and Khapra}]{ramesh-etal-2022-samanantar}
Gowtham Ramesh, Sumanth Doddapaneni, Aravinth Bheemaraj, Mayank Jobanputra, Raghavan AK, Ajitesh Sharma, Sujit Sahoo, Harshita Diddee, Mahalakshmi J, Divyanshu Kakwani, Navneet Kumar, Aswin Pradeep, Srihari Nagaraj, Kumar Deepak, Vivek Raghavan, Anoop Kunchukuttan, Pratyush Kumar, and Mitesh~Shantadevi Khapra. 2022.
\newblock \href {https://doi.org/10.1162/tacl_a_00452} {Samanantar: The largest publicly available parallel corpora collection for 11 indic languages}.
\newblock \emph{Transactions of the Association for Computational Linguistics}, 10:145--162.

\bibitem[{Ranasinghe et~al.(2023)Ranasinghe, Ghosh, Pal, and Senapati}]{lowres-2023-T-Ranasinghe}
T~Ranasinghe, K~Ghosh, AS~Pal, and A~Senapati. 2023.
\newblock \href {https://dl.acm.org/doi/abs/10.1145/3632754.3633278?casa_token=hS-3-e2ucs8AAAAA:GqH80eCwKqhg3FvZTa0WeWREtghmuoP_R-eYFbx1u6mFyhAMFWYj1GFPB_8LZbEx2FnMRYz4sD0h} {Overview of the hasoc subtracks at fire 2023: Hate speech and offensive content identification in assamese, bengali, bodo, gujarati and sinhala}.
\newblock \emph{Proceedings of the 15th …}.

\bibitem[{Rani et~al.(2024)Rani, Negi, Jha, Suryawanshi, Ojha, Buitelaar, and McCrae}]{llm-2024-rani-wildre}
Priya Rani, Gaurav Negi, Saroj Jha, Shardul Suryawanshi, Atul~Kr. Ojha, Paul Buitelaar, and John~P. McCrae. 2024.
\newblock \href {https://aclanthology.org/2024.wildre-1.3} {Findings of the wildre shared task on code-mixed less-resourced sentiment analysis for indo-aryan languages}.
\newblock In \emph{Proceedings of the 7th Workshop on Indian Language Data: Resources and Evaluation}, pages 17--23, Torino, Italy. ELRA and ICCL.

\bibitem[{Ratnam et~al.(2024)Ratnam, Karthika, Praveena, and Taniya}]{lowres-2024-DJ-Ratnam}
DJ~Ratnam, AN~Karthika, K~Praveena, and R~Taniya. 2024.
\newblock \href {https://www.sciencedirect.com/science/article/pii/S187705092400588X} {Phonogram-based automatic typo correction in malayalam social media comments}.
\newblock \emph{Procedia Computer …}.

\bibitem[{Rauf et~al.(2022)Rauf, Irfan, Mushtaq, and Ashraf}]{llm-2022-Rauf}
F~Rauf, R~Irfan, L~Mushtaq, and M~Ashraf. 2022.
\newblock \href {https://vfast.org/journals/index.php/VTSE/article/view/1290} {Fake news detection in urdu using deep learning}.
\newblock \emph{VFAST Transactions on Software Engineering}.

\bibitem[{Ravva(2023)}]{lowres-2023-P-Ravva}
P~Ravva. 2023.
\newblock \href {https://web2py.iiit.ac.in/research_centres/publications/download/mastersthesis.pdf.b93b1f715416394f.46696e616c5468657369735375626d697373696f6e5f50726979616e6b61202831292e706466.pdf} {Systems and resources for telugu: Question answering and summarization}.
\newblock \emph{nan}.

\bibitem[{Rayala et~al.(2023)Rayala, Seshadri, and Sristy}]{lowres-2023-UR-Rayala}
UR~Rayala, K~Seshadri, and NB~Sristy. 2023.
\newblock \href {https://dl.acm.org/doi/abs/10.1145/3620670?casa_token=RoYFtxXGP7kAAAAA:69Z34_J8x68dwFIM6uqqlly2LTjp9bgzd9LacnBrf8eIm0C6sauPGmpFgzcHVrGEvLY8TqjA_eXx} {Sentiment analysis of code-mixed telugu-english data leveraging syllable and word embeddings}.
\newblock \emph{ACM Transactions on Asian and …}.

\bibitem[{Reddy et~al.(2023)Reddy, Gunti, Kumar, and {Sridevi}}]{multimodal-2023-R-Reddy}
Rohan Reddy, Swathi Gunti, Prasanna Kumar, and {Sridevi}. 2023.
\newblock Multilingual image captioning: Multimodal framework for bridging visual and linguistic realms in tamil and telugu through transformers.
\newblock (preprint).

\bibitem[{Reedy et~al.(2024)Reedy, Meghana, and Subhadra}]{lowres-2024-BG-Reedy}
BG~Reedy, G~Meghana, and T~Subhadra. 2024.
\newblock \href {https://ieeexplore.ieee.org/abstract/document/10649114/?casa_token=nqZeaHYLqikAAAAA:Lr1S9wGUDkUMwYO5Bv-MwJZJZ-d-dstfOV9d1MjSXUR3wkKLB7lZX-5n-PHeOaAN7gsXwwkd} {Enhancing the recognition of hand written telugu characters: Natural language processing and machine learning approach}.
\newblock \emph{2024 IEEE …}.

\bibitem[{{Reliance-NVIDIA}(2024)}]{reliance-nvidia-llm}
{Reliance-NVIDIA}. 2024.
\newblock {Reliance and {NVIDIA} Collaborate on India-Focused Large Language Model}.
\newblock {https://nvidianews.nvidia.com/news/reliance -and-nvidia-partner-to-advance-ai-in-india-for-india}.

\bibitem[{Revathi et~al.(2024)Revathi, Prasad, and Gattim}]{lowres-2024-B-Revathi}
B~Revathi, MVD Prasad, and NK~Gattim. 2024.
\newblock \href {https://beei.org/index.php/EEI/article/view/8170} {Computationally efficient resnet based telugu handwritten text detection}.
\newblock \emph{Bulletin of Electrical Engineering and …}.

\bibitem[{Roshan et~al.(2023)Roshan, Bhacho, and Zai}]{lowres-2023-R-Roshan}
R~Roshan, IA~Bhacho, and S~Zai. 2023.
\newblock \href {https://www.mdpi.com/2673-4591/46/1/5} {Comparative analysis of tf–idf and hashing vectorizer for fake news detection in sindhi: A machine learning and deep learning approach}.
\newblock \emph{Engineering Proceedings}.

\bibitem[{Roy et~al.(2022)Roy, Bhawal, and Subalalitha}]{llm-2022-Roy}
PK~Roy, S~Bhawal, and CN~Subalalitha. 2022.
\newblock \href {https://www.sciencedirect.com/science/article/pii/S0885230822000250} {Hate speech and offensive language detection in dravidian languages using deep ensemble framework}.
\newblock \emph{Computer Speech and Language}.

\bibitem[{Rudregowda et~al.(2023)Rudregowda, Kulkarni, HL, and Ravi}]{lowres-2023-S-Rudregowda}
S~Rudregowda, S~Patil Kulkarni, G~HL, and V~Ravi. 2023.
\newblock \href {https://www.mdpi.com/2624-599X/5/1/20} {Visual speech recognition for kannada language using vgg16 convolutional neural network}.
\newblock \emph{Acoustics}.

\bibitem[{Rynjah et~al.(2022)Rynjah, Syiem, and Singh}]{lowres-2022-F-Rynjah}
F~Rynjah, B~Syiem, and LJ~Singh. 2022.
\newblock \href {https://www.researchgate.net/profile/Joyprakash-Lairenlakpam-2/publication/362385557_Investigating_Khasi_Speech_Recognition_Systems_using_a_Recurrent_Neural_Network-Based_Language_Model/links/6646daf522a7f16b4f2ff2a6/Investigating-Khasi-Speech-Recognition-Systems-using-a-Recurrent-Neural-Network-Based-Language-Model.pdf} {Investigating khasi speech recognition systems using a recurrent neural network-based language model}.
\newblock \emph{Int J Eng Trends Technol}.

\bibitem[{Sadhu et~al.(2024)Sadhu, Saha, and Shahriyar}]{sadhu2024}
Jayanta Sadhu, Maneesha~Rani Saha, and Rifat Shahriyar. 2024.
\newblock \href {http://arxiv.org/abs/2407.03536} {Social bias in large language models for bangla: An empirical study on gender and religious bias}.

\bibitem[{Sai et~al.(2024)Sai, Kumar, and Saumya}]{lowres-2024-C-Sai}
C~Sai, R~Kumar, and S~Saumya. 2024.
\newblock \href {https://aclanthology.org/2024.dravidianlangtech-1.19/} {Iiitdwd-svc@ dravidianlangtech-2024: Breaking language barriers; hate speech detection in telugu-english code-mixed text}.
\newblock \emph{Proceedings of the Fourth …}.

\bibitem[{Sai~B et~al.(2023)Sai~B, Dixit, Nagarajan, Kunchukuttan, Kumar, Khapra, and Dabre}]{sai-b-etal-2023-indicmt}
Ananya Sai~B, Tanay Dixit, Vignesh Nagarajan, Anoop Kunchukuttan, Pratyush Kumar, Mitesh~M. Khapra, and Raj Dabre. 2023.
\newblock \href {https://doi.org/10.18653/v1/2023.acl-long.795} {{I}ndic{MT} eval: A dataset to meta-evaluate machine translation metrics for {I}ndian languages}.
\newblock In \emph{Proceedings of the 61st Annual Meeting of the Association for Computational Linguistics (Volume 1: Long Papers)}, pages 14210--14228, Toronto, Canada. Association for Computational Linguistics.

\bibitem[{Sankalp et~al.(2024)Sankalp, Jain, Bhaduri, Roy, and Chadha}]{kj2024decoding}
KJ~Sankalp, Vinija Jain, Sreyoshi Bhaduri, Tamoghna Roy, and Aman Chadha. 2024.
\newblock Decoding the diversity: A review of the indic ai research landscape.
\newblock \emph{arXiv preprint arXiv:2406.09559}.

\bibitem[{Santhoshi and Badugu(2022)}]{lowres-2022-G-Santhoshi}
G~Santhoshi and S~Badugu. 2022.
\newblock \href {https://link.springer.com/chapter/10.1007/978-981-16-9669-5_3} {Development of different word vectors and testing using text classification algorithms for telugu}.
\newblock \emph{… and Applications, Volume 1: Proceedings of …}.

\bibitem[{Santosh and Livingston(2023)}]{lowres-2023-Santosh}
Santosh and LM~Jenila Livingston. 2023.
\newblock \href {https://www.worldscientific.com/doi/abs/10.1142/S0218488523400068} {An approach for automated kannada subtitle generation from kannada video}.
\newblock \emph{International Journal of Uncertainty …}.

\bibitem[{Sarma and Pathak(2024)}]{lowres-2024-S-Sarma-blstm}
S~Sarma and N~Pathak. 2024.
\newblock \href {https://link.springer.com/chapter/10.1007/978-981-97-3604-1_37} {An emerging area of research: Building an assamese ai chatbot for educational institutions using bi-lstm model}.
\newblock \emph{NIELIT's International Conference on Communication}.

\bibitem[{Sarmah et~al.(2023)Sarmah, Rehman, and Mahanta}]{lowres-2023-A-Sarmah}
A~Sarmah, R~Rehman, and P~Mahanta. 2023.
\newblock \href {https://ieeexplore.ieee.org/abstract/document/10127265/?casa_token=zsg4FQWFvFIAAAAA:QUPP0bQWfKdYnIZwCrzTvD3jz1kROEzi4eBwsEImE-D6e7bns0sNUu_svtaHyrwD7NM243sJ} {Identification and analysis of assamese vowel speech signal using formant feature-fusion and feed-forward neural network model}.
\newblock \emph{2023 4th International …}.

\bibitem[{Sathya et~al.(2023)Sathya, Gopalakrishnan, and Manickam}]{lowres-2023-MK-Sathya}
MK~Sathya, KH~Gopalakrishnan, and PA~Manickam. 2023.
\newblock \href {https://ceur-ws.org/Vol-3681/T6-18.pdf} {Sinhala and gujarati hate speech detection.}
\newblock \emph{FIRE (Working …}.

\bibitem[{Sen et~al.(2022)Sen, Parida, Kotwal, Panda, Bojar, and Dash}]{Sen2022-tg}
Arghyadeep Sen, Shantipriya Parida, Ketan Kotwal, Subhadarshi Panda, Ond{\v r}ej Bojar, and Satya~Ranjan Dash. 2022.
\newblock Bengali visual genome: A multimodal dataset for machine translation and image captioning.
\newblock In \emph{Intelligent Data Engineering and Analytics}, Smart innovation, systems and technologies, pages 63--70. Springer Nature Singapore, Singapore.

\bibitem[{Sethi et~al.(2022)Sethi, Dev, and Bansal}]{multimodal-2022-N-Sethi}
Nandini Sethi, Amita Dev, and Poonam Bansal. 2022.
\newblock \href {https://doi.org/10.1109/AIST55798.2022.10064790} {Multimodal machine translation for sanskrit-hindi: An empirical analysis}.
\newblock In \emph{2022 4th International Conference on Artificial Intelligence and Speech Technology (AIST)}, pages 1--4.

\bibitem[{Shah and Swaminarayan(2022)}]{lowres-2022-P-Shah-reviews}
P~Shah and P~Swaminarayan. 2022.
\newblock \href {https://www.inderscienceonline.com/doi/abs/10.1504/IJDATS.2022.124763} {Machine learning-based sentiment analysis of gujarati reviews}.
\newblock \emph{International Journal of Data …}.

\bibitem[{Shah et~al.(2022{\natexlab{a}})Shah, Swaminarayan, and Patel}]{lowres-2022-P-Shah-film}
P~Shah, P~Swaminarayan, and M~Patel. 2022{\natexlab{a}}.
\newblock \href {https://www.researchgate.net/profile/Maitri-Patel-2/publication/358267134_Sentiment_analysis_on_film_review_in_Gujarati_language_using_machine_learning/links/62467f638068956f3c5e96f5/Sentiment-analysis-on-film-review-in-Gujarati-language-using-machine-learning.pdf} {Sentiment analysis on film review in gujarati language using machine learning}.
\newblock \emph{International Journal of …}.

\bibitem[{Shah et~al.(2022{\natexlab{b}})Shah, Chadha, Gupta, Dhuriya, Chhimwal, Gaur, and Raghavan}]{shah2022worderrorrategood}
Priyanshi Shah, Harveen~Singh Chadha, Anirudh Gupta, Ankur Dhuriya, Neeraj Chhimwal, Rishabh Gaur, and Vivek Raghavan. 2022{\natexlab{b}}.
\newblock \href {http://arxiv.org/abs/2203.16601} {Is word error rate a good evaluation metric for speech recognition in indic languages?}

\bibitem[{Shahariar et~al.(2024)Shahariar, Shawon, Shah, Alam, and Mahbub}]{fakereview}
G~M Shahariar, Md. Tanvir~Rouf Shawon, Faisal~Muhammad Shah, Mohammad~Shafiul Alam, and Md.~Shahriar Mahbub. 2024.
\newblock \href {https://doi.org/https://doi.org/10.1016/j.neucom.2024.127732} {Bengali fake reviews: A benchmark dataset and detection system}.
\newblock \emph{Neurocomputing}, 592:127732.

\bibitem[{Shashidhar et~al.(2024)Shashidhar, Shashank, and Jagadamba}]{lowres-2024-R-Shashidhar}
R~Shashidhar, MP~Shashank, and G~Jagadamba. 2024.
\newblock \href {https://link.springer.com/chapter/10.1007/978-3-031-68602-3_20} {A fusion approach for kannada speech recognition using audio and visual cue}.
\newblock \emph{IoT Sensors, ML, AI and …}.

\bibitem[{Shashirekha et~al.(2022)Shashirekha, Balouchzahi, and Anusha}]{lowres-2022-HL-Shashirekha}
HL~Shashirekha, F~Balouchzahi, and MD~Anusha. 2022.
\newblock \href {https://arxiv.org/abs/2211.09847} {Coli-machine learning approaches for code-mixed language identification at the word level in kannada-english texts}.
\newblock \emph{arXiv preprint arXiv …}.

\bibitem[{Sheshadri and Bharath(2022)}]{lowres-2022-SK-Sheshadri}
SK~Sheshadri and BS~Bharath. 2022.
\newblock \href {https://ieeexplore.ieee.org/abstract/document/9984521/} {Unsupervised neural machine translation for english to kannada using pre-trained language model}.
\newblock \emph{2022 13th …}.

\bibitem[{Shetty et~al.(2022)Shetty, Hegde, and Shetty}]{lowres-2022-S-Shetty}
S~Shetty, S~Hegde, and S~Shetty. 2022.
\newblock \href {https://search.proquest.com/openview/c0c46eaeba7bf0e061c62de06c344345/1?pq-origsite=gscholar&cbl=2035897} {Implementation on parts of speech tagging of kannada language text}.
\newblock \emph{NeuroQuantology}.

\bibitem[{Shetty et~al.(2023)Shetty, Mady, and Bhustali}]{lowres-2023-S-Shetty}
S~Shetty, P~Mady, and VK~Bhustali. 2023.
\newblock \href {https://ieeexplore.ieee.org/abstract/document/10276314/?casa_token=kCxf7ce_nX4AAAAA:pE5Sz26wEMq1uaeEZ-hX3OMNH8Q2g7gPgcZnK2SJnhw9dd93hZlvFzCA7IhR7SOVLbT7N3G3} {English to kannada translation using bert model}.
\newblock \emph{… Conference on Network …}.

\bibitem[{Shokoori et~al.(2022)Shokoori, Shinwari, and Popal}]{lowres-2022-AF-Shokoori}
AF~Shokoori, M~Shinwari, and JA~Popal. 2022.
\newblock \href {https://ieeexplore.ieee.org/abstract/document/9753959/?casa_token=bKaHq8X1B1IAAAAA:-J0gVwjabbd8_fAcupHqpBkbb0CAPO6rMankbfr18ec3owWcA5fLku9XdMSC1ljI9vh9aFgZ} {Sign language recognition and translation into pashto language alphabets}.
\newblock \emph{2022 6th International …}.

\bibitem[{Shree and Shambhavi(2022)}]{lowres-2022-M-Rajani-Shree}
M~Rajani Shree and BR~Shambhavi. 2022.
\newblock \href {https://link.springer.com/chapter/10.1007/978-981-16-7330-6_5} {Syntactical parsing of kannada text based on cyk algorithm}.
\newblock \emph{Proceedings of Third International …}.

\bibitem[{Shrestha et~al.(2024)Shrestha, Pokhrel, and Adhikari}]{lowres-2024-S-Shrestha}
S~Shrestha, S~Pokhrel, and S~Adhikari. 2024.
\newblock \href {http://conference.ioe.edu.np/publications/ioegc15/IOEGC-15-002-A1-2-27.pdf} {Topic-wide keyword generation from nepali documents through graph-based extraction and lda topic models}.
\newblock \emph{nan}.

\bibitem[{Si~Thu(2024)}]{lowres-2024-M-Si-Thu}
Min Si~Thu. 2024.
\newblock \href {https://papers.ssrn.com/sol3/papers.cfm?abstract_id=4904320} {{Burmese-microbiology-1K}}.
\newblock \emph{SSRN Electron. J.}

\bibitem[{Singh et~al.(2023{\natexlab{a}})Singh, Jayakumar, Deekshitha, and Kumar}]{lowres-2023-A-Singh-chhattis}
A~Singh, A~Jayakumar, G~Deekshitha, and H~Kumar. 2023{\natexlab{a}}.
\newblock \href {https://link.springer.com/chapter/10.1007/978-3-031-48312-7_13} {An end-to-end tts model in chhattisgarhi, a low-resource indian language}.
\newblock \emph{… Conference on Speech …}.

\bibitem[{Singh et~al.(2024{\natexlab{a}})Singh, Sharma, and Singh}]{mimic-misogyny}
Aakash Singh, Deepawali Sharma, and Vivek~Kumar Singh. 2024{\natexlab{a}}.
\newblock \href {https://doi.org/10.1145/3656169} {Mimic: Misogyny identification in multimodal internet content in hindi-english code-mixed language}.
\newblock \emph{ACM Trans. Asian Low-Resour. Lang. Inf. Process.}
\newblock Just Accepted.

\bibitem[{Singh et~al.(2024{\natexlab{b}})Singh, Sharma, and Singh}]{multimodal-2024-A-Singh}
Aakash Singh, Deepawali Sharma, and Vivek~Kumar Singh. 2024{\natexlab{b}}.
\newblock \href {https://doi.org/10.1145/3656169} {Mimic: Misogyny identification in multimodal internet content in hindi-english code-mixed language}.
\newblock \emph{ACM Trans. Asian Low-Resour. Lang. Inf. Process.}
\newblock Just Accepted.

\bibitem[{Singh et~al.(2024{\natexlab{c}})Singh, Murthy, kumar, Sen, and Ramakrishnan}]{singh2024}
Abhishek~Kumar Singh, Rudra Murthy, Vishwajeet kumar, Jaydeep Sen, and Ganesh Ramakrishnan. 2024{\natexlab{c}}.
\newblock \href {http://arxiv.org/abs/2407.13522} {Indic qa benchmark: A multilingual benchmark to evaluate question answering capability of llms for indic languages}.

\bibitem[{Singh et~al.(2024{\natexlab{d}})Singh, Gupta, Bharadwaj, Tewari, and Talukdar}]{indicgenbench}
Harman Singh, Nitish Gupta, Shikhar Bharadwaj, Dinesh Tewari, and Partha Talukdar. 2024{\natexlab{d}}.
\newblock \href {https://doi.org/10.18653/v1/2024.acl-long.595} {{I}ndic{G}en{B}ench: A multilingual benchmark to evaluate generation capabilities of {LLM}s on {I}ndic languages}.
\newblock In \emph{Proceedings of the 62nd Annual Meeting of the Association for Computational Linguistics (Volume 1: Long Papers)}, pages 11047--11073, Bangkok, Thailand. Association for Computational Linguistics.

\bibitem[{Singh et~al.(2023{\natexlab{b}})Singh, Aggarwal, and Singh}]{Singh2023-sy}
Lovejit Singh, Naveen Aggarwal, and Sarbjeet Singh. 2023{\natexlab{b}}.
\newblock {PUMAVE-D}: panjab university multilingual audio and video facial expression dataset.
\newblock \emph{Multimed. Tools Appl.}, 82(7):10117--10144.

\bibitem[{Sodhar and Sulaiman(2023{\natexlab{a}})}]{lowres-2023-IN-Sodhar}
IN~Sodhar and S~Sulaiman. 2023{\natexlab{a}}.
\newblock \href {https://sciresol.s3.dualstack.us-east-2.amazonaws.com/IJST/Articles/2023/Issue-12/IJST-2023-236.pdf} {Exploration of sindhi corpus through statistical analysis on the basis of reality}.
\newblock \emph{Indian Journal of Science}.

\bibitem[{Sodhar and Sulaiman(2023{\natexlab{b}})}]{lowres-2023-IN-Sodhar-morpho}
IN~Sodhar and S~Sulaiman. 2023{\natexlab{b}}.
\newblock \href {https://sciresol.s3.us-east-2.amazonaws.com/IJST/Articles/2023/Issue-35/IJST-2023-1719.pdf} {Morphology-assisted sindhi text analysis for natural language processing applications}.
\newblock \emph{Indian Journal …}.

\bibitem[{Sodhar et~al.(2023)Sodhar, Sulaiman, and Buller}]{lowres-2023-IN-Sodhar-senti}
IN~Sodhar, S~Sulaiman, and AH~Buller. 2023.
\newblock \href {https://search.proquest.com/openview/f35dcf45fa779188e9712c7430334d9f/1?pq-origsite=gscholar&cbl=5444811&casa_token=8SyXEZQ2BlIAAAAA:KRnxB1nsz7HCgkAROwnbRWmxzzgHW_91x1OSitm_7CNWUz-HV3Son2eGGtAfqKnMwdA1Jrtmww} {Hybrid approach used to analyze the sentiments of romanized text (sindhi)}.
\newblock \emph{International Journal of …}.

\bibitem[{Som et~al.(2024)Som, Mishra, Das, and Singh}]{lowres-2024-P-Som}
P~Som, R~Mishra, S~Das, and RK~Singh. 2024.
\newblock \href {https://ieeexplore.ieee.org/abstract/document/10530821/?casa_token=416TUYmHZycAAAAA:4gI70V88qZ1H-pqNw40CjnGie_8uwC0p6oDLBM54j2c9alH9BurIrTBwRbSi6WonjeX70hBU} {Evaluating machine learning models for hate speech detection in odia language}.
\newblock \emph{… on Cognitive, Green …}.

\bibitem[{Spiesberger et~al.(2023)Spiesberger, Triantafyllopoulos, Tsangko, and Schuller}]{speech-2024-Spiesberger}
Anika~A Spiesberger, Andreas Triantafyllopoulos, Iosif Tsangko, and Bj{\"o}rn~W Schuller. 2023.
\newblock Abusive speech detection in indic languages using acoustic features.
\newblock In \emph{{INTERSPEECH} 2023}, pages 2683--2687, ISCA. ISCA.

\bibitem[{Srinivas et~al.(2023)Srinivas, Prashanth, and Malapaka}]{lowres-2023-J-Srinivas}
J~Srinivas, US~Prashanth, and A~Malapaka. 2023.
\newblock \href {https://link.springer.com/chapter/10.1007/978-981-99-7820-5_32} {Raithubot: An rlhf-fine-tuned telugu chatbot for farmers}.
\newblock \emph{… Conference on Data …}.

\bibitem[{Srivastava et~al.(2023)Srivastava, Gupta, Prakash, Kuriakose, and Murthy}]{speech-2023-Srivastava}
Sudhanshu Srivastava, Ishika Gupta, Anusha Prakash, Jom Kuriakose, and Hema~A. Murthy. 2023.
\newblock \href {http://arxiv.org/abs/2302.06227} {Fast and small footprint hybrid hmm-hifigan based system for speech synthesis in indian languages}.

\bibitem[{Su et~al.(2020)Su, Jin, and Finkelstein}]{su2020hifigan}
Jiaqi Su, Zeyu Jin, and Adam Finkelstein. 2020.
\newblock \href {http://arxiv.org/abs/2006.05694} {Hifi-gan: High-fidelity denoising and dereverberation based on speech deep features in adversarial networks}.

\bibitem[{Subbarao(2008)}]{subbarao2008typological}
Karumuri~V Subbarao. 2008.
\newblock Typological characteristics of south asian languages.
\newblock \emph{Language in South Asia}, pages 49--78.

\bibitem[{Subedi et~al.(2024)Subedi, Regmi, and Bal}]{lowres-2024-B-Subedi}
B~Subedi, S~Regmi, and BK~Bal. 2024.
\newblock \href {https://aclanthology.org/2024.lrec-main.611/} {… potential of large language models (llms) for low-resource languages: A study on named-entity recognition (ner) and part-of-speech (pos) tagging for nepali …}.
\newblock \emph{Proceedings of the 2024 …}.

\bibitem[{Sudarsan and Sankar(2024)}]{lowres-2024-D-Sudarsan}
D~Sudarsan and D~Sankar. 2024.
\newblock \href {https://dl.acm.org/doi/abs/10.1145/3686311} {An ensemble neural network model for malayalam character recognition from palm leaf manuscripts}.
\newblock \emph{ACM Transactions on Asian and Low-Resource …}.

\bibitem[{Suresh and Damotharan(2024)}]{lowres-2024-SK-Suresh}
SK~Suresh and U~Damotharan. 2024.
\newblock \href {https://ieeexplore.ieee.org/abstract/document/10631613/?casa_token=cL08uaKI4j0AAAAA:hojxsOFh0M0PeZZDn8ddA_sT_IT2hnJ6z-mTTAS6cjNs59rsnJtyD-IT1jG9eOBYT69mKMd4} {Kannada-english code-mixed speech synthesis}.
\newblock \emph{2024 International Conference on …}.

\bibitem[{Swain et~al.(2022)Swain, Maji, Kabisatpathy, and Routray}]{lowres-2022-M-Swain}
M~Swain, B~Maji, P~Kabisatpathy, and A~Routray. 2022.
\newblock \href {https://link.springer.com/article/10.1007/s40747-022-00713-w} {A dcrnn-based ensemble classifier for speech emotion recognition in odia language}.
\newblock \emph{Complex and Intelligent …}.

\bibitem[{Taheri et~al.(2023)Taheri, Roy, and Kabir}]{Taheri2023-mt}
Zaima~Sartaj Taheri, Animesh~Chandra Roy, and Ahasan Kabir. 2023.
\newblock {BEmoFusionNet}: A deep learning approach for multimodal emotion classification in bangla social media posts.
\newblock In \emph{2023 26th International Conference on Computer and Information Technology ({ICCIT})}, pages 1--6. IEEE.

\bibitem[{Tallu and Battula(2022)}]{lowres-2022-DAV-Padmaja-Tallu}
DAV~Padmaja Tallu and V~Battula. 2022.
\newblock \href {https://www.academia.edu/download/84529053/154432.pdf} {Sentiment analysis for social media telugu language}.
\newblock \emph{nan}.

\bibitem[{Talpur et~al.(2023)Talpur, Talpur, and Samar}]{lowres-2023-N-Talpur}
N~Talpur, MJ~Talpur, and T~Samar. 2023.
\newblock \href {https://rjllp.muet.edu.pk/index.php/repertus/article/view/24} {Researching on analysis and creating corpus from primary level sindhi language book for sindhi}.
\newblock \emph{Repertus: Journal of Linguistics …}.

\bibitem[{Talukdar and Sarma(2023{\natexlab{a}})}]{lowres-2023-C-Talukdar}
C~Talukdar and SK~Sarma. 2023{\natexlab{a}}.
\newblock \href {https://journals.uob.edu.bh/handle/123456789/5179} {Hybrid model for efficient assamese text classification using cnn-lstm}.
\newblock \emph{International Journal of Computing and …}.

\bibitem[{Talukdar and Sarma(2023{\natexlab{b}})}]{lowres-2023-K-Talukdar}
K~Talukdar and SK~Sarma. 2023{\natexlab{b}}.
\newblock \href {https://ieeexplore.ieee.org/abstract/document/10372865/?casa_token=AfK0zEddLFkAAAAA:Ue-XcAGicDkG1noNHmeNHgpSUAiYelOfHt2HFkua7wosh2kpEW9H2kBV8Y6jaArJrd2go5bV} {Upos tagger for low resource assamese language: Lstm and bilstm based modelling}.
\newblock \emph{2023 IEEE International Conference on …}.

\bibitem[{Talukdar and Sarma(2024{\natexlab{a}})}]{lowres-2024-K-Talukdar}
K~Talukdar and SK~Sarma. 2024{\natexlab{a}}.
\newblock \href {https://search.ebscohost.com/login.aspx?direct=true&profile=ehost&scope=site&authtype=crawler&jrnl=2158107X&AN=178397245&h=AvvqjrzUliHxMHgijOrz0%2BGGU7fGEShYYprUEyi5PkMCvndJGz7xhBoR5pFwPhNCGZQCex1URh1HtGgHXCxtQQ%3D%3D&crl=c} {Incremental learning for gru and rnn-based assamese upos tagger.}
\newblock \emph{International Journal of Advanced …}.

\bibitem[{Talukdar and Sarma(2024{\natexlab{b}})}]{lowres-2024-S-Talukdar}
S~Talukdar and SK~Sarma. 2024{\natexlab{b}}.
\newblock \href {https://search.proquest.com/openview/375aaca5a64d56de4f018846b22b63bd/1?pq-origsite=gscholar&cbl=4433095} {Enabling natural language processing and ai research in low-resource languages: Development and description of an assamese upos tagged dataset}.
\newblock \emph{Journal of Electrical Systems}.

\bibitem[{Tamang and Bora(2024)}]{lowres-2024-S-Tamang-tokenizer}
S~Tamang and DJ~Bora. 2024.
\newblock \href {https://arxiv.org/abs/2410.03718} {Performance evaluation of tokenizers in large language models for the assamese language}.
\newblock \emph{arXiv preprint arXiv:2410.03718}.

\bibitem[{{Tata-NVIDIA}(2024)}]{tata-nvidia-llm}
{Tata-NVIDIA}. 2024.
\newblock {Tata partners with {NVIDIA} to Build Large-Scale AI Infrastructure}.
\newblock {https://nvidianews.nvidia.com/news/ tata-partners-with-nvidia-to-build-large- scale-ai-infrastructure}.

\bibitem[{Tech-Mahindra(2024)}]{techmahindra-indus}
Tech-Mahindra. 2024.
\newblock {The Indus Project: Large Language Model for Hindi and Indian Languages}.
\newblock \url{https://www.techmahindra.com/makers-lab/indus-project/}.

\bibitem[{Tejas and Dutta(2022)}]{lowres-2022-S-Tejas}
S~Tejas and KK~Dutta. 2022.
\newblock \href {https://ieeexplore.ieee.org/abstract/document/9971971/} {Siamese neural networks for kannada handwritten dataset}.
\newblock \emph{2022 IEEE 3rd Global Conference for …}.

\bibitem[{Thabah et~al.(2022)Thabah, Mitri, Saha, Maji, and Purkayastha}]{lowres-2022-NDJ-Thabah}
N~Donald~Jefferson Thabah, Aiom~Minnette Mitri, Goutam Saha, Arnab~Kumar Maji, and Bipul~Shyam Purkayastha. 2022.
\newblock \href {https://doi.org/10.1007/s11334-022-00497-9} {A deep connection to khasi language through pre-trained embedding}.
\newblock \emph{Innovations in Systems and Software Engineering}.

\bibitem[{Thandil and Basheer(2023)}]{lowres-2023-RK-Thandil}
RK~Thandil and KP~Mohamed Basheer. 2023.
\newblock \href {https://link.springer.com/chapter/10.1007/978-981-99-6553-3_1} {Deep spectral feature representations via attention-based neural network architectures for accented malayalam speech—a low-resourced language}.
\newblock \emph{… Conference on Data …}.

\bibitem[{Thapliyal et~al.(2022)Thapliyal, Pont-Tuset, Chen, and Soricut}]{thapliyal2022crossmodal3600massivelymultilingualmultimodal}
Ashish~V. Thapliyal, Jordi Pont-Tuset, Xi~Chen, and Radu Soricut. 2022.
\newblock \href {http://arxiv.org/abs/2205.12522} {Crossmodal-3600: A massively multilingual multimodal evaluation dataset}.

\bibitem[{Thimmaraja et~al.(2023)Thimmaraja, Nagaraja, and Jayanna}]{lowres-2023-YG-Thimmaraja}
YG~Thimmaraja, BG~Nagaraja, and HS~Jayanna. 2023.
\newblock \href {https://www.sciencedirect.com/science/article/pii/S2667305323001138} {Advancements in end-to-end isolated kannada asr system by combining robust noise elimination technique and tdnn}.
\newblock \emph{Intelligent Systems with …}.

\bibitem[{Tonmoy(2023)}]{llm-2023-Tonmoy}
S.m Towhidul~Islam Tonmoy. 2023.
\newblock \href {https://doi.org/10.18653/v1/2023.banglalp-1.46} {Embeddings at {BLP}-2023 task 2: Optimizing fine-tuned transformers with cost-sensitive learning for multiclass sentiment analysis}.
\newblock In \emph{Proceedings of the First Workshop on Bangla Language Processing (BLP-2023)}, pages 340--346, Singapore. Association for Computational Linguistics.

\bibitem[{Traversaal.ai(2024)}]{traversaal2024}
Traversaal.ai. 2024.
\newblock {Pakistani AI Startup Traversaal.ai Wins \$100,000 Grant at Meta AI Accelerator Competition 2024}.
\newblock {https://www.islamabadscene.com/ pakistani-ai-startup-traversaal-ai-wins- 100000-grant-at-meta-ai-accelerator-competition-2024/}.

\bibitem[{Trumpp(2023)}]{lowres-2023-E-Trumpp}
E~Trumpp. 2023.
\newblock \href {https://books.google.com/books?hl=en&lr=&id=7OK0EAAAQBAJ&oi=fnd&pg=PR1&dq=%22language+model%22%7C%22speech+processing%22%7C%22multimodal+model%22%7C%22retrieval+augmented+generation%22%7C%22graph%22%7C%22ai%22%7C%22artificial+intelligence%22+%22rohingya%22%7C%22santali%22%7C%22sindhi%22%7C%22kurukh%22%7C%22telugu%22&ots=TvweJAJtV8&sig=CC9XZX2xv7YaG80TVJma1EPdJb0} {Grammar of the sindhi language}.
\newblock \emph{nan}.

\bibitem[{Ullah et~al.(2024{\natexlab{a}})Ullah, Ullah, Khan, and Aurangzeb}]{lowres-2024-I-Ullah}
I~Ullah, K~Ullah, H~Khan, and K~Aurangzeb. 2024{\natexlab{a}}.
\newblock \href {https://peerj.com/articles/cs-2163/} {Pashto poetry generation: deep learning with pre-trained transformers for low-resource languages}.
\newblock \emph{PeerJ Computer …}.

\bibitem[{Ullah et~al.(2024{\natexlab{b}})Ullah, Ali, Chandio, Brohi, and Laghari}]{lowres-2024-S-Ullah}
Sami Ullah, Najma~Imtiaz Ali, Shah~Murad Chandio, Imtiaz~Ali Brohi, and Barkat~Ali Laghari. 2024{\natexlab{b}}.
\newblock \href {https://doi.org/10.62019/abbdm.v4i1.111} {Part-of-speech tagging for balochi language: A data driven application of conditional random fields}.
\newblock \emph{The Asian Bulletin of Big Data Management}, 4(1):Data Science 4(1),229–236.

\bibitem[{Upadhayay and Behzadan(2024)}]{taco}
Bibek Upadhayay and Vahid Behzadan. 2024.
\newblock \href {http://arxiv.org/abs/2311.10797} {Taco: Enhancing cross-lingual transfer for low-resource languages in llms through translation-assisted chain-of-thought processes}.

\bibitem[{{UrduPoint News}(2024)}]{ignite-urdu}
{UrduPoint News}. 2024.
\newblock {Ignite – NICAT Pakistan Hosts Asia-Pacific’s First Meta Llama Pitchathon}.
\newblock \url{https://www.urdupoint.com/en/technology/ignite-nicat-pakistan-hosts-asia-pacifics-1542925.html}.

\bibitem[{Vajrobol(2022)}]{lowres-2022-V-Vajrobol}
V~Vajrobol. 2022.
\newblock \href {https://aclanthology.org/2022.icon-wlli.2/} {Coli-kanglish: Word-level language identification in code-mixed kannada-english texts shared task using the distilka model}.
\newblock \emph{… Language Identification in Code-mixed Kannada …}.

\bibitem[{Vamsi and Bataineh(2023)}]{lowres-2023-B-Vamsi}
B~Vamsi and A~Al Bataineh. 2023.
\newblock \href {https://search.ebscohost.com/login.aspx?direct=true&profile=ehost&scope=site&authtype=crawler&jrnl=0992499X&AN=173573084&h=%2BhGf%2FPae%2BYSFSLUz5ckqAoFvHkiVGGWV%2B7BfZaWeqtFxbh7GKgikf8A%2FB%2BRTS8EoeO9zWxeBiK1xEdzVn62bZQ%3D%3D&crl=c} {Lexical based reordering models for english to telugu machine translation.}
\newblock \emph{Revue d'Intelligence …}.

\bibitem[{Vedula et~al.(2023)Vedula, Kodali, Shrivastava, and Kumaraguru}]{llm-2023-Vedula}
Bhaskara~Hanuma Vedula, Prashant Kodali, Manish Shrivastava, and Ponnurangam Kumaraguru. 2023.
\newblock \href {https://doi.org/10.18653/v1/2023.wassa-1.58} {{P}recog{IIITH}@{WASSA}2023: Emotion detection for {U}rdu-{E}nglish code-mixed text}.
\newblock In \emph{Proceedings of the 13th Workshop on Computational Approaches to Subjectivity, Sentiment, {\&} Social Media Analysis}, pages 601--605, Toronto, Canada. Association for Computational Linguistics.

\bibitem[{Venkateswarlu et~al.(2022)Venkateswarlu, Kumar, and Veeraswamy}]{lowres-2022-SC-Venkateswarlu}
SC~Venkateswarlu, NU~Kumar, and D~Veeraswamy. 2022.
\newblock \href {https://link.springer.com/chapter/10.1007/978-981-19-0011-2_40} {Speech intelligibility quality in telugu speech patterns using a wavelet-based hybrid threshold transform method}.
\newblock \emph{Intelligent systems and …}.

\bibitem[{Verma et~al.(2023)Verma, Shree, and Modi}]{speech-2023-Verma}
Tushar Verma, Atul Shree, and Ashutosh Modi. 2023.
\newblock \href {https://api.semanticscholar.org/CorpusID:260911679} {Asr for low resource and multilingual noisy code-mixed speech}.
\newblock \emph{Interspeech}.

\bibitem[{Vishnu~Kumar and Lalithamani(2022)}]{multimodal-2022-VHV-Kumar}
V~H Vishnu~Kumar and N~Lalithamani. 2022.
\newblock \href {https://doi.org/10.1109/AIC55036.2022.9848810} {English to tamil multi-modal image captioning translation}.
\newblock In \emph{2022 IEEE World Conference on Applied Intelligence and Computing (AIC)}, pages 332--338.

\bibitem[{Viswanadh(2024)}]{lowres-2024-VB-Viswanadh}
VB~Viswanadh. 2024.
\newblock \href {https://ieeexplore.ieee.org/abstract/document/10493230/} {Sentiment analysis of telugu news articles decoding textual nuances}.
\newblock \emph{2024 Second International Conference on …}.

\bibitem[{Weston et~al.(2015)Weston, Bordes, Chopra, Rush, van Merriënboer, Joulin, and Mikolov}]{weston2015aicompletequestionansweringset}
Jason Weston, Antoine Bordes, Sumit Chopra, Alexander~M. Rush, Bart van Merriënboer, Armand Joulin, and Tomas Mikolov. 2015.
\newblock \href {http://arxiv.org/abs/1502.05698} {Towards ai-complete question answering: A set of prerequisite toy tasks}.

\bibitem[{Yadava et~al.(2024)Yadava, Nagaraja, and Jayanna}]{lowres-2024-GT-Yadava}
GT~Yadava, BG~Nagaraja, and HS~Jayanna. 2024.
\newblock \href {https://link.springer.com/article/10.1007/s11042-023-15854-4} {An end-to-end continuous kannada asr system under uncontrolled environment}.
\newblock \emph{Multimedia Tools and …}.

\bibitem[{Yadavalli and Mirishkar(2022)}]{lowres-2022-A-Yadavalli}
A~Yadavalli and GS~Mirishkar. 2022.
\newblock \href {https://aclanthology.org/2022.naacl-srw.36/} {Exploring the effect of dialect mismatched language models in telugu automatic speech recognition}.
\newblock \emph{Proceedings of the 2022 …}.

\bibitem[{Yang et~al.(2021)Yang, Chi, Chuang, Lai, Lakhotia, Lin, Liu, Shi, Chang, Lin, Huang, Tseng, Lee, Liu, Huang, Dong, Li, Watanabe, Mohamed, and Lee}]{Yang2021-sd}
Shu-Wen Yang, Po-Han Chi, Yung-Sung Chuang, Cheng-I~Jeff Lai, Kushal Lakhotia, Yist~Y Lin, Andy~T Liu, Jiatong Shi, Xuankai Chang, Guan-Ting Lin, Tzu-Hsien Huang, Wei-Cheng Tseng, Ko-Tik Lee, Da-Rong Liu, Zili Huang, Shuyan Dong, Shang-Wen Li, Shinji Watanabe, Abdelrahman Mohamed, and Hung-Yi Lee. 2021.
\newblock {SUPERB}: Speech processing universal {PERformance} benchmark.
\newblock In \emph{Interspeech 2021}, ISCA. ISCA.

\bibitem[{Yune and Soe(2023)}]{lowres-2023-F-Yune}
F~Yune and KM~Soe. 2023.
\newblock \href {https://ieeexplore.ieee.org/abstract/document/10482974/?casa_token=fIevN7Q1haYAAAAA:dq9BYNtNJ-z4XN2fNsF8O2VDWsVf0tEtZp3OL1AipB2K01KXezzie6odbwk3BShzCDVqddJ7} {Advancing transfer learning paradigms for myanmar (burmese) to wa (austroasiatic language family) language translation}.
\newblock \emph{2023 26th Conference of the Oriental …}.

\bibitem[{Zhong et~al.(2024)Zhong, Liu, Pan, Zhang, Zhou, Liang, Wu, Lyu, Shu, Yu, Cao, Jiang, Chen, Li, Chen, Hu, Liu, Zhao, Xu, Dai, Zhao, Zhang, Zhao, Yang, Chen, Wang, Ruan, Wang, Zhao, Zhang, Ren, Qin, Chen, Li, Zidan, Jahin, Chen, Xia, Holmes, Zhuang, Wang, Xu, Xia, Yu, Tang, Yang, Sun, Yang, Lu, Wang, Chai, Li, Lu, Sun, Zhang, Ge, Hu, Zhang, Zhou, Zhang, Zhang, Liu, Jiang, Kong, Xiang, Ren, Liu, Jiang, Bao, Zhang, Li, Li, Liu, Shen, Sikora, Zhai, Zhu, and Liu}]{zhong2024evaluationopenaio1opportunities}
Tianyang Zhong, Zhengliang Liu, Yi~Pan, Yutong Zhang, Yifan Zhou, Shizhe Liang, Zihao Wu, Yanjun Lyu, Peng Shu, Xiaowei Yu, Chao Cao, Hanqi Jiang, Hanxu Chen, Yiwei Li, Junhao Chen, Huawen Hu, Yihen Liu, Huaqin Zhao, Shaochen Xu, Haixing Dai, Lin Zhao, Ruidong Zhang, Wei Zhao, Zhenyuan Yang, Jingyuan Chen, Peilong Wang, Wei Ruan, Hui Wang, Huan Zhao, Jing Zhang, Yiming Ren, Shihuan Qin, Tong Chen, Jiaxi Li, Arif~Hassan Zidan, Afrar Jahin, Minheng Chen, Sichen Xia, Jason Holmes, Yan Zhuang, Jiaqi Wang, Bochen Xu, Weiran Xia, Jichao Yu, Kaibo Tang, Yaxuan Yang, Bolun Sun, Tao Yang, Guoyu Lu, Xianqiao Wang, Lilong Chai, He~Li, Jin Lu, Lichao Sun, Xin Zhang, Bao Ge, Xintao Hu, Lian Zhang, Hua Zhou, Lu~Zhang, Shu Zhang, Ninghao Liu, Bei Jiang, Linglong Kong, Zhen Xiang, Yudan Ren, Jun Liu, Xi~Jiang, Yu~Bao, Wei Zhang, Xiang Li, Gang Li, Wei Liu, Dinggang Shen, Andrea Sikora, Xiaoming Zhai, Dajiang Zhu, and Tianming Liu. 2024.
\newblock \href {http://arxiv.org/abs/2409.18486} {Evaluation of openai o1: Opportunities and challenges of agi}.

\bibitem[{Zimmer and Tripathee(2024)}]{lowres-2024-F-Zimmer}
F~Zimmer and MG~Tripathee. 2024.
\newblock \href {https://opus4.kobv.de/opus4-rhein-waal/frontdoor/deliver/index/docId/1967/file/Purnima_Gurung_27993.pdf} {Sentiment analysis in nepali tweets: Leveraging transformer-based pre-trained models}.
\newblock \emph{nan}.

\bibitem[{Zoho(2024)}]{zoho-llm}
Zoho. 2024.
\newblock {Zoho's Ascent in the LLM Race}.
\newblock {https://www.constellationr.com/blog-news/insights/ zoho-focuses-llm-efforts-nvidia-architecture}.

\end{thebibliography}
\bibliographystyle{acl_natbib}

\end{document}